\documentclass[pdflatex,sn-basic]{sn-jnl}

\usepackage{graphicx}%
\usepackage{multirow}%
\usepackage{amsmath,amssymb,amsfonts}%
\usepackage{amsthm}%
\usepackage{mathrsfs}%
\usepackage[title]{appendix}%
\usepackage{xcolor}%
\usepackage{textcomp}%
\usepackage{manyfoot}%
\usepackage{booktabs}%
\usepackage{algorithm}%
\usepackage{algorithmicx}%
\usepackage{algpseudocode}%
\usepackage{listings}%

\usepackage{subfigure}
\usepackage{scalefnt}
\usepackage{longtable}
\usepackage{enumitem}

\usepackage{colortbl}

\theoremstyle{thmstyleone}%
\theoremstyle{thmstyletwo}%

\theoremstyle{thmstylethree}%

\begin{document}

\title[Imbalanced Regression Pipeline Recommendation]{Imbalanced Regression Pipeline Recommendation}


\author*[1]{\fnm{Juscimara G.} \sur{Avelino}}\email{jga2@cin.ufpe.br}

\author[1]{\fnm{George D. C.} \sur{Cavalcanti}}\email{gdcc@cin.ufpe.br}

\author[2]{\fnm{Rafael M. O.} \sur{Cruz}}\email{rafael.menelau-cruz@etsmtl.ca}

\affil[1]{\orgdiv{Centro de Informática}, \orgname{Universidade Federal de Pernambuco}, \orgaddress{\street{Cidade Universitária}, \city{Recife}, \postcode{50740-560}, \state{PE}, \country{Brazil}}}

\affil[2]{\orgdiv{École de Technologie Supérieure}, \orgname{Université du Québec}, \orgaddress{\street{1100 Notre Dame St. W.}, \city{Montréal}, \postcode{QC H3C 1K3}, \state{Québec}, \country{Canada}}}


\abstract{Imbalanced problems are prevalent in various real-world scenarios and are extensively explored in classification tasks. However, they also present challenges for regression tasks due to the rarity of certain target values. A common alternative is to employ balancing algorithms in preprocessing to address dataset imbalance. However, due to the variety of resampling methods and learning models, determining the optimal solution requires testing many combinations. Furthermore, the learning model, dataset, and evaluation metric affect the best strategies. This work proposes the Meta-learning for Imbalanced Regression (Meta-IR) framework, which diverges from existing literature by training meta-classifiers to recommend the best pipeline composed of the resampling strategy and learning model per task in a zero-shot fashion. The meta-classifiers are trained using a set of meta-features to learn how to map the meta-features to the classes indicating the best pipeline. We propose two formulations: Independent and Chained. Independent trains the meta-classifiers to separately indicate the best learning algorithm and resampling strategy. Chained involves a sequential procedure where the output of one meta-classifier is used as input for another to model intrinsic relationship factors. The Chained scenario showed superior performance, suggesting a relationship between the learning algorithm and the resampling strategy per task. Compared with AutoML frameworks, Meta-IR obtained better results. Moreover, compared with baselines of six learning algorithms and six resampling algorithms plus no resampling, totaling 42 (6 $\times$ 7) configurations, Meta-IR outperformed all of them. The code, data, and further information of the experiments can be found on GitHub: \url{https://github.com/JusciAvelino/Meta-IR}.}

\keywords{Imbalance domain learning, Regression, Resampling Strategies, Meta-Learning.}



\maketitle

\section{Introduction}\label{Introduction}

\begin{figure} [!h]
    \centering
    \includegraphics[width=9cm]{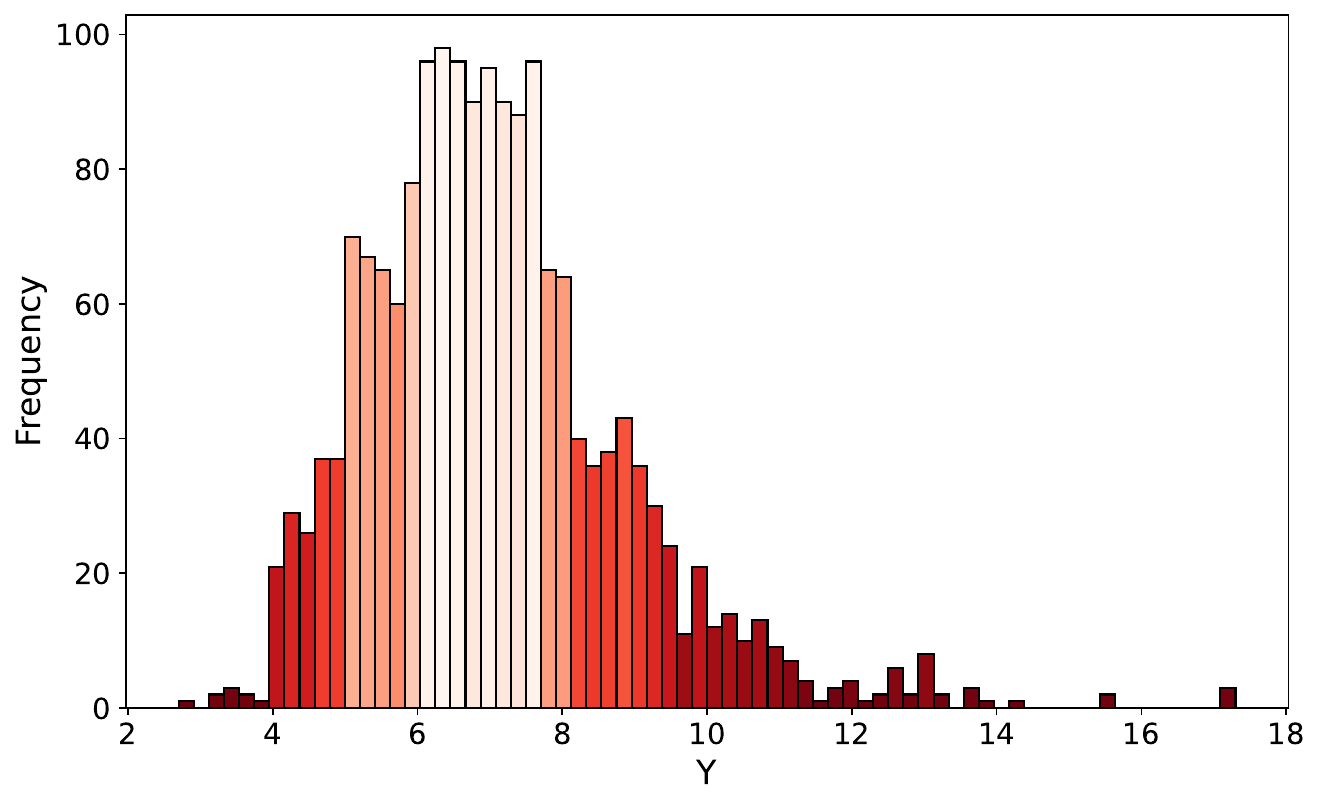}
    \caption{Distribution and frequency of the target value Y from the FuelCons dataset. Stronger colors indicate less represented examples (rare examples), while lighter colors indicate more represented examples (normal examples).}

    \label{imb}
 
\end{figure}

Imbalanced problems are characterized by the low representation of some target values. In classification tasks, an imbalanced dataset is determined by the presence of a class with low representation (minority class) compared to another (majority class)~\citep{ganganwar2012overview,haixiang2017learning}. However, in regression problems, the target value is continuous, representing a complex definition. In this context, \cite{ribeiro2011utility} proposes the concept of relevance function that determines the relevance of continuous target values to define which examples are rare and normal. Such classification allows verifying the imbalance between the instances called rare and normal. Figure \ref{imb} represents the distribution and frequency of examples from an imbalanced dataset (FuelCons). The values on the edges of the graph have low frequency and are considered rare examples.


 

In many cases, low-represented values hold significant importance, both to users and prediction processes. For instance, in software engineering, prediction mistakes in large projects are associated with higher development costs~\citep{rathore2017linear, rathore2017towards}. Similarly, forecasting errors become significantly more costly in meteorological applications when dealing with extreme conditions such as extremely high temperatures~\citep{ribeiro2020imbalanced}. Learning algorithms often struggle with these scenarios, prioritizing frequent target ranges while neglecting rare cases. This limitation can lead to suboptimal predictions for these specific instances, which are sometimes misinterpreted as outliers. Outliers, while typically rare and often considered unwanted, can actually hold significant value in some situations. In imbalanced regression problems, these cases can provide critical insights that are crucial for accurate predictions. In such contexts, ignoring these rare occurrences could lead to severe prediction errors, demonstrating that, far from being irrelevant, outliers can offer essential information for achieving more accurate and valuable outcomes.

Few studies explore solutions for handling imbalanced regression problems. Commonly, such solutions employ resampling strategies such as the Synthetic Minority Oversampling TEchnique for Regression (SmoteR)~\citep{torgo2013smote} and WEighted Relevance-based Combination Strategy (WERCS)~\citep{branco2019pre} to balance the training data distributions prior to the training process. Although there is limited research exploring solutions for handling imbalanced regression problems, there are various resampling strategies available to choose from~\citep{wu2022imbalancedlearningregression}. Recent findings in \cite{avelino2024resampling} indicate that no single solution — a learning model and a resampling strategy — performs consistently well in all cases. Furthermore, there is a significant dependency between the blocks in the pipeline. The application of a resampling algorithm alters the optimal learning model in nearly 50\% of the datasets studied.  This interdependence means that the choice of the resampling strategy can influence the optimal choice for the learning model and vice versa.

These findings underscore that approaches focusing solely on selecting and optimizing the best model per dataset, without accounting for preprocessing algorithms like Naive AutoML~\citep{mohr2023naive}, are likely to underperform. It may overlook the significant impact of resampling techniques on model performance, missing potentially more effective models during the initial selection. Thus, identifying the optimal pipeline is crucial for accurate imbalanced regression results. In a landscape with numerous resampling strategies and regression algorithms, identifying the most fitting pipeline requires potentially evaluating all or a significant subset, thereby leading to a resource-intensive endeavor.

To enhance optimal solution selection, meta-learning (MtL)~\citep{vanschoren2019meta, khan2020literature} models prove effective. These models are based on meta-features, i.e., problem characteristics extracted from data, to learn new tasks more quickly. These meta-features — such as number of examples, number of attributes, number of rare cases, percentage of rare cases and data complexity measures~\citep{lorena2019complex} — enable a meta-classifier to correlate them with algorithm performance (target attribute). Once trained, the meta-learning model works in a zero-shot fashion, suggesting algorithms and techniques that best fit the new task's meta-features.

\begin{figure} [!h]
    \centering
    \includegraphics[width=13cm]{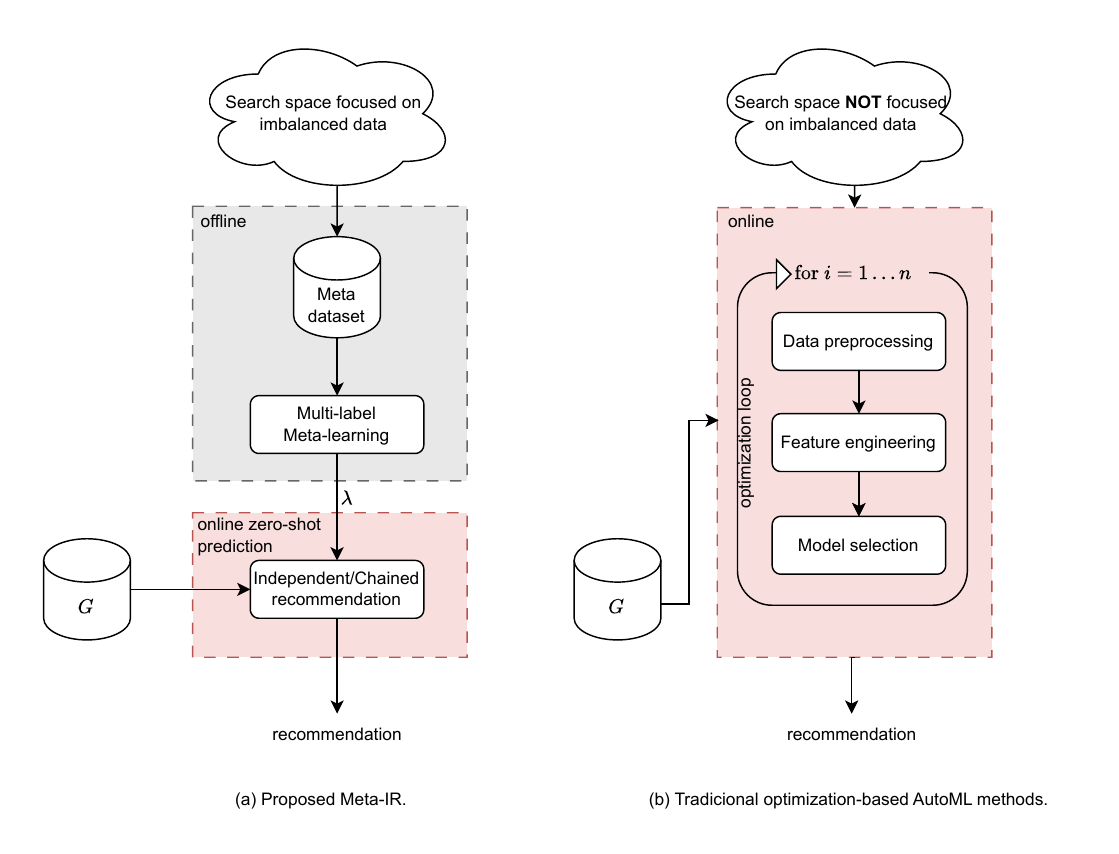}
    \caption{Comparison between our proposed meta-learning-based model Meta-IR and traditional AutoML.}
    \label{fig:comparison}
\end{figure}

In light of this, we propose a meta-learning-based model — Meta-learning for Imbalanced Regression (Meta-IR)— as an alternative solution for imbalanced regression problems. Meta-IR, illustrated in Figure \ref{fig:comparison}(a), is designed to suggest optimal learning models and resampling strategies based on problem meta-features, leveraging metrics tailored for optimization in imbalanced regression problems, such as F1-score for Regression (F1-scoreR) and Squared error-relevance area (SERA). The meta-classifier recommends resampling strategy and learning algorithm — a \textbf{pipeline} — customized to the new problem's meta-features. For example, it could recommend the Random Over-sampling strategy with the Decision Tree learning model for a given dataset. Alternatively, for a different dataset, it might be more appropriate to use the WEighted Relevance-based Combination strategy with the Random Forest regression model, demonstrating the need to adjust the strategy and model according to the specific characteristics of the dataset. We propose two ways to train the meta-models: Independent and Chained. The Independent approach involves training the meta-classifiers to indicate the best learning model and resample strategy separately. In contrast, the Chained model is proposed based on the hypothesis that a dependency exists between the resampling algorithm and the regressor. In this way, Chained performs a sequential procedure where the output of one meta-classifier is used as input for another. In other words, the information of the best resampling algorithm is fed into the meta-classifier that predicts the best regressor, and vice versa.

On the other hand, traditional AutoML methods, illustrated in Figure \ref{fig:comparison}(b), lack specific approaches to handle imbalanced regression problems in the search space, as is the case, for example, with frameworks such as Auto-Sklearn, H2O, and TPOT. In contrast, our method addresses imbalanced regression problems within the search space by recommending resampling strategies tailored to the dataset at hand.

Furthermore, current AutoML frameworks impose a high computational burden, requiring exhaustive searches through iterative optimization loops, such as those employed by Auto-Sklearn and H2O framework. These loops span multiple stages, including data preprocessing, feature engineering, model selection, and hyperparameter tuning, evaluating numerous combinations to identify the best-performing pipeline. For example, Auto-Sklearn handles preprocessing by imputing missing values and normalizing features, applying feature engineering techniques such as principal component analysis (PCA), and evaluating models such as Random Forests and Logistic Regression during the model selection phase. Similarly, H2O AutoML automates tasks such as encoding categorical variables, exploring feature selection to determine key predictors, and testing models such as Gradient Boosting Machines (GBMs) and Deep Neural Networks. Our method, however, can recommend the optimal pipeline in a zero-shot prediction, offering a significantly faster and more efficient solution. Thus, our method is not only suitable for imbalanced regression problems but also incurs a substantially lower computational cost compared to traditional AutoML frameworks.

The experimental analysis, including 218 datasets, six learning models, six resampling strategies, and no resampling, resulting in a total of 42 possible configurations, showed that Meta-IR outperforms the Random recommendation and Majority (always recommends the technique that appears most frequently in the meta-dataset) at the meta-level. At the base-level, comparing the performance achieved by the recommended learning models and resampling strategies, Meta-IR recommendations were better than Random, Majority, and AutoML frameworks — such as Auto-sklearn~\citep{feurer2015efficient}, H2O~\citep{H2OAutoML20}, and TPOT~\citep{OlsonGECCO2016} —  
for F1-scoreR metric. Furthermore, statistical analysis revealed that the proposed method produces statistically better significant results at the base-level. In conclusion, the Meta-IR approach demonstrated an advantage in terms of time efficiency, as it was approximately 50 times faster than the AutoML frameworks. In the analysis of meta-feature importance, it was observed that certain characteristics significantly influenced the recommendation of resampling strategies and learning models. We conducted an interpretation and identified the most important features for these meta-learning tasks, providing insights into which factors are most influential in the recommendation process.



The main contributions of this work are:







\begin{itemize}
    \item We propose a meta-learning method for recommending pipelines for imbalanced regression problems.
    \item We introduce two meta-learning recommendation approaches: Independent and Chained. The Independent approach trains meta-classifiers separately to recommend the resampling strategy and regression model, while the Chained approach uses the output of one meta-classifier as input for the next, aiming to enhance overall performance.
    \item We conducted extensive experiments on 218 datasets, extracting valuable analyses and conclusions that contribute significantly to the literature. To our knowledge, this is the first work to perform such an extensive experiment across so many datasets. Additionally, these experiments generated comprehensive metadata, providing new insights and valuable resources for the research community.
    \item An analysis of meta-feature importance reveals key meta-features crucial for the meta-model's predictive ability in recommending learning models and resampling strategies for imbalanced regression problems.
\end{itemize}

This work is organized as follows: Section~\ref{Background} presents an overview of the areas of imbalanced regression, automated machine learning and meta-learning. The proposed model is presented in detail in Section~\ref{Meta-IR}. Section~\ref{Methodology} presents the experimental methodology, describing the metadata, meta-features, meta-targets, meta-classifier, and evaluation methodology. In Sections~\ref{Results and Discussions}~and~\ref{Conclusions}, we discuss the results and conclude this research.

\section{Background}\label{Background}

Imbalanced Regression and AutoML are the two areas of research that have been combined in the proposed method in this work. Therefore, this section provides a global overview of these two areas of research.

\subsection{Imbalanced Regression}\label{IR}

\subsubsection{Relevance Function}
\label{sec:relevancefunction}


The simultaneous occurrence of two factors characterizes the imbalance problem: i) the disproportionate user preference in the target variable domain; and ii) the insufficient representation, in the available data, of the most relevant cases for the user~\citep{branco2017smogn}. There is a greater difficulty when the problem occurs in regression tasks, compared to classification tasks, since the continuous target value can have an infinite number of values and the determination of rare cases is not trivial~\citep{branco2017smogn}. To address this,~\cite{ribeiro2011utility} proposed the notion of a relevance function ($\phi : Y \rightarrow [0,1]$). For each dataset, the relevance function maps the target value ($Y$) into a [0,1] scale of relevance, where 0 and 1 represent the minimum and maximum relevance, respectively.

The relevance function automatically set the significance of data points using the Piecewise Cubic Hermite Interpolating Polynomials (pchip)~\citep{dougherty1989nonnegativity} over a set of control points. These control points can be defined either through domain knowledge or provided by an automated method. Defining control points using an automatic function is preferable since knowledge can often be unavailable~\citep{ribeiro2020imbalanced}. In~\cite{ribeiro2011utility} the control points are based on Tukey's bloxpot~\citep{tukey1970exploratory}. The interval proposed by Tukey is based on the adjacent limits [$adj_L = Q1 - 1.5 \cdot IQR$ and  $adj_H = Q3 + 1.5 \cdot IQR$] where $Q1$ and $Q3$ are the first and third quartile, respectively, and $IQR = Q3 - Q1$. In turn, the control points are defined by the adjacent limits ($adj_L$ and $adj_H$) and the median value ($\Tilde{Y}$).

Given a dataset $D$, two sets containing rare ($D_R$) and normal ($D_N$) instances are defined considering the relevance threshold ($t_R$) defined by the user as follows: $D_R = \{\langle \textbf{x},y \rangle \in D: \phi(y) \geq t_R\}$ and $D_N = \{\langle \textbf{x},y \rangle \in D:\phi(y) < t_R\}$. The described method is used to determine the rare cases in each dataset, such as the ones shown in Table \ref{tab:dadosTempo}. Full details about the relevance function can be found in~\cite{ribeiro2011utility, avelino2024resampling}.

\subsubsection{Resampling Strategies}
\label{sec:strategies}

Strategies for Imbalanced Regression can be categorized into three main groups~\citep{avelino2024resampling}: i)~Regression models, ii)~Learning Process Modification, and iii)~Evaluate Metrics. The first group consists of regression models, including both single models and ensembles. The second group highlights additional strategies aimed at refining the learning process to better handle imbalanced target, such as resampling strategies. Finally, the third group focuses on evaluation metrics proposed specific for imbalanced regression. Among these strategies, the most common way to deal with imbalanced data sets is to use resampling strategies, changing the data distribution to balance the targets~\citep{moniz2017resampling}. 

Resampling strategies or balancing algorithms, precede the learning process by altering the distribution of examples through resampling the training data. This method removes samples of normal cases (i.e., under-sampling) or generates synthetic samples for rare cases (i.e., over-sampling). Different resampling strategies have been proposed to deal with imbalanced regression problems. Most of these techniques are based on existing resampling strategies proposed for classification problems~\citep{avelino2024resampling}.

The SmoteR algorithm, which is a variant of Smote~\citep{chawla2002smote}, in which adaptations were made to fit the regression problem, the main ones being: i) definition of rare cases; ii) creation of synthetic examples; and iii) definition of target values for new examples. Also on the basis of the Smote algorithm,~\cite{camacho2022geometric} proposed Geometric SMOTE, which generates synthetic data points along the line connecting two existing data points. Other strategies adapted from the imbalanced classification are: Random Under-sampling~\citep{torgo2013smote} which was based on the idea in~\cite{kubat1997addressing}, Random Over-sampling~\citep{branco2019pre} proposed for classification in~\cite{batista2004study}, and Introduction of Gaussian Noise~\citep{branco2016ubl} adapted from~\cite{lee1999regularization, lee2000noisy}. In turn, the SmoteR with Gaussian Noise (SMOGN)~\citep{branco2017smogn} and WEighted Relevance-based Combination Strategy (WERCS)~\citep{branco2019pre} strategies were originally proposed for the imbalanced regression problem. Furthermore,~\cite{song2022distsmogn} introduced a distributed version of the SMOGN called DistSMOGN. The method uses a weighted sampling technique to generate synthetic samples for rare cases, in addition to considering the data distribution in each node of the distributed system. For the imbalanced data streams in regression models context,~\cite{aminian2021chebyshev} introduced two sampling strategies (ChebyUS, ChebyOS) based on the Chebyshev inequality to improve the performance of existing regression methods on imbalanced data streams. The approaches use a weighted learning strategy that assigns higher weights to rare cases in order to balance the training process.

\begin{figure}[]
\centering
\subfigure[SmoteR\label{subfig:SmoteR}]{
\includegraphics[width=12cm,height=2cm]{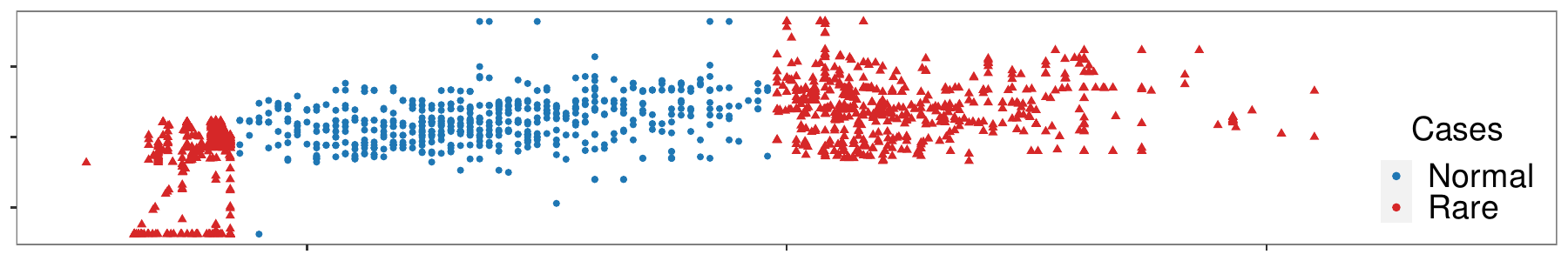}}

\subfigure[Random Over-sampling\label{subfig:RO}]{
\includegraphics[width=12cm,height=2cm]{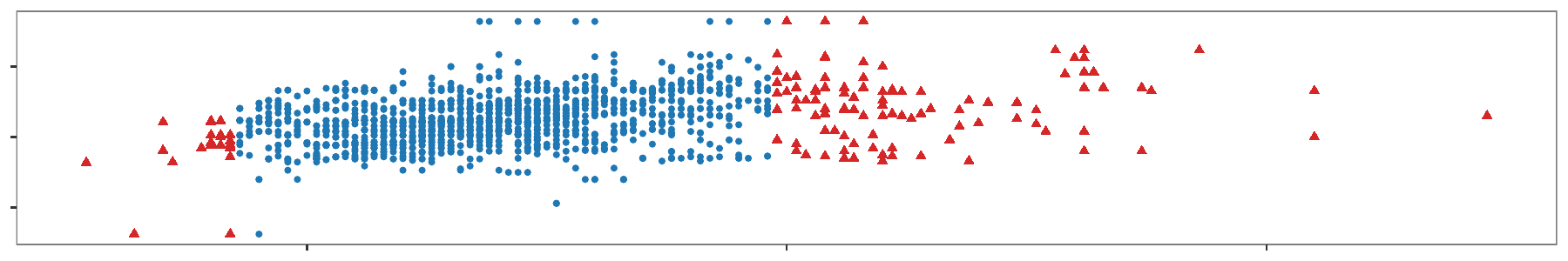}}

\subfigure[Introduction of Gaussian Noise\label{subfig:GN}]{
\includegraphics[width=12cm,height=2cm]{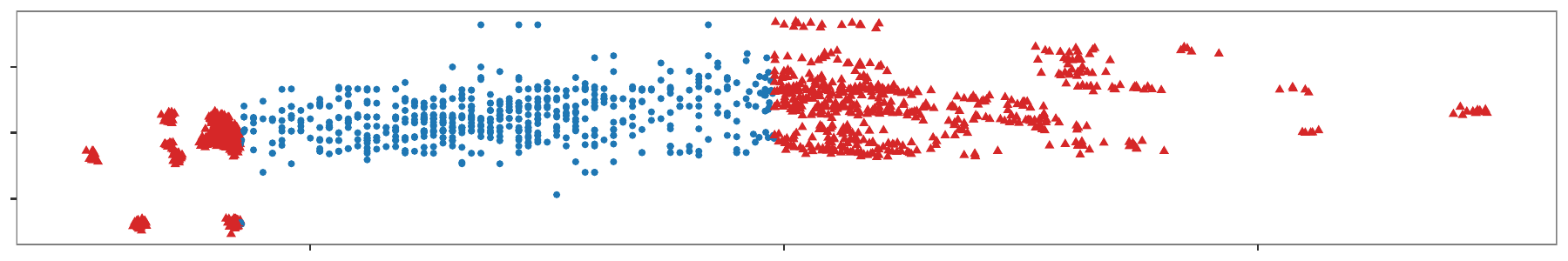}}

\subfigure[SMOGN\label{subfig:SG}]{
\includegraphics[width=12cm,height=2cm]{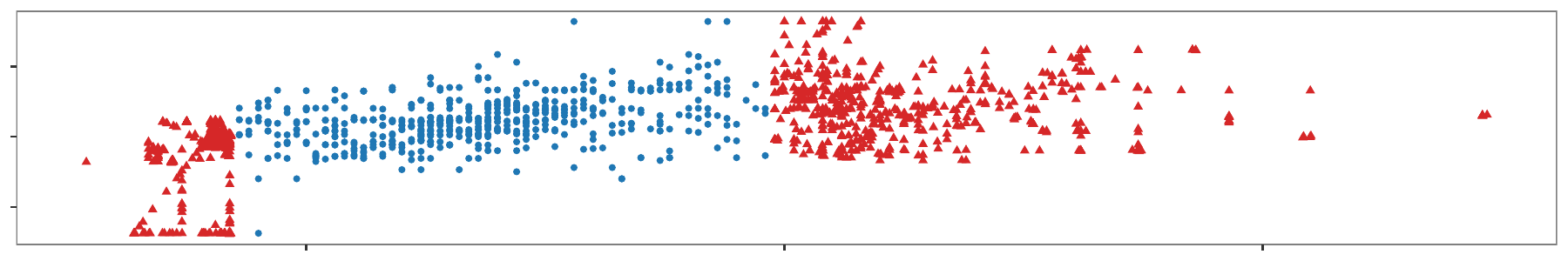}}

\subfigure[Random Under-sampling\label{subfig:RU}]{
\includegraphics[width=12cm,height=2cm]{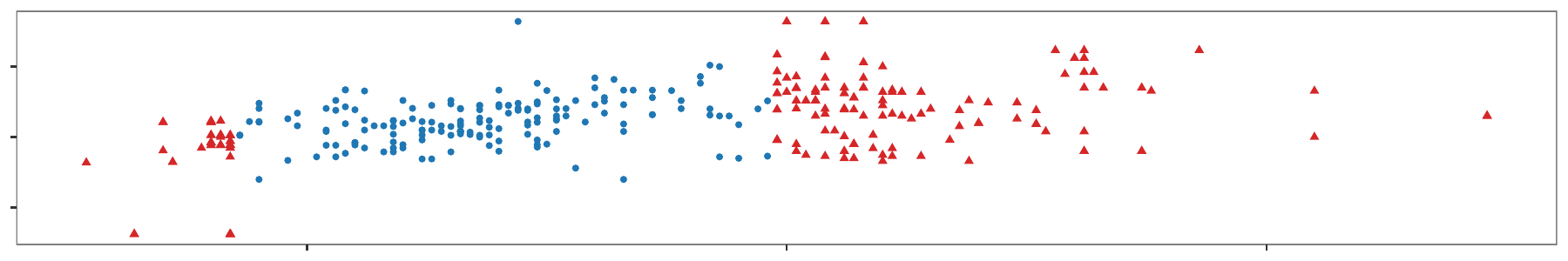}}

\subfigure[WERCS\label{subfig:WC}]{
\includegraphics[width=12cm,height=2cm]{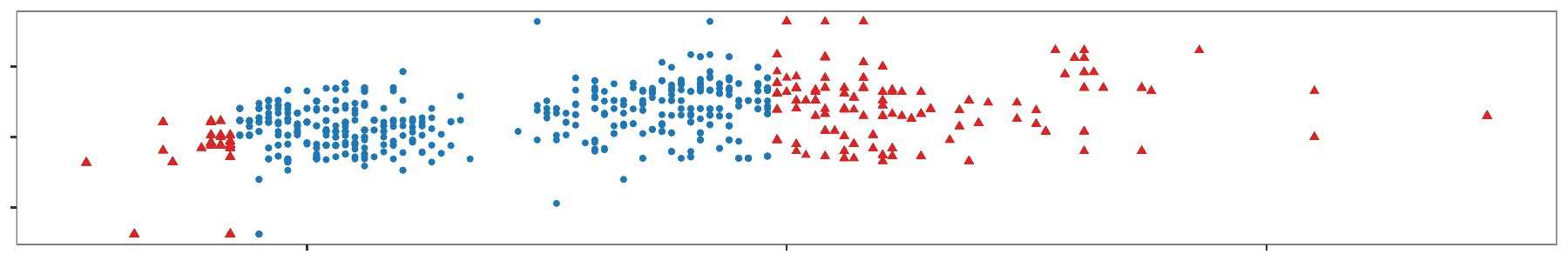}}

\caption{Distribution of the examples of the FuelCons dataset after applying the resampling strategies, considering $t_R$=0.8.}
\label{fig:strategies}
\end{figure}

Each strategy resamples the data differently. Figure~\ref{fig:strategies} shows this difference, applying the resampling strategies SmoteR, Random Over-sampling, Random Under-sampling, Gaussian Noise Introduction, SMOGN and WERCS to the \textit{FuelCons} dataset. The $t_R$ parameter value was set to 0.8, except for the WERCS, as it operates without a threshold. Standard values were utilized for the other parameters. In the visualization, the Y-axis represented the target values, and the X-axis represented the attribute. More details on how the points in this figure were generated can be found in \cite{avelino2024resampling}.

In contrast to SmoteR~\ref{subfig:SmoteR}, which generates synthetic data through oversampling, the Random Over-sampling strategy~\ref{subfig:RO} increases data by simply replicating rare cases (data points in darker shade). Another way of over-sampling is through the Introduction of Gaussian Noise, as shown in Figure~\ref{subfig:GN}, where the generated data differs from the originals, creating more diversity than Random Over-sampling. Similarly, the SMOGN strategy~\ref{subfig:SG} combines the Introduction of Gaussian Noise and SmoteR, maintaining a similarity with the SMOTER distribution. However, compared to the Introduction of Gaussian Noise, it becomes evident that the diversity of generated examples is higher in SMOGN. On the other hand, Random Under-sampling~\ref{subfig:RU} removes normal data randomly. Furthermore, the WERCS strategy~\ref{subfig:WC} combines Random Over-sampling and Random Under-sampling and is the only strategy that does not require a threshold.

Despite the difference, resampling strategies are based on the same principles: decrease normal examples and/or increase rare examples. Under-sampling, which reduces normal examples, is the basis of the Random Under-sampling strategy. On the contrary, over-sampling, which increases rare examples, can be performed simply as in Random Over-sampling or by generating synthetic cases as in the SmoteR Algorithm and Introduction of Gaussian Noise. Other strategies are based on the models already described, such as the SmoteR with Gaussian Noise (SMOGN), which combines the Random Under-sampling strategy and the over-sampling strategies SmoteR and Introduction of Gaussian Noise. And the WEighted Relevance-based Combination Strategy (WERCS) which combines the Random Under-sampling and Random Over-sampling strategies, using weights for resampling. More details about these strategies can be found in~\cite{branco2019pre, torgo2013smote, avelino2024resampling}.


Resampling strategies have the advantage of allowing the use of any learning algorithm without affecting the model's explainability~\citep{branco2019pre}. However, applying and validating these strategies require effort due to the vast space of solutions. Furthermore, selecting the optimal strategy depends on the dataset, the regression model employed, and the metrics used to assess system performance~\citep{avelino2024resampling}. Therefore, given the variety of problems and the increasing interest in this field, there is a need to advance the studies and address the problem of imbalanced regression from another perspective. One way to address the problem is through a recommendation model based on meta-learning. Thus, we propose a model to automate the choice of solutions, i.e., to recommend a pipeline for the problem, with decisions based on data meta-features. It is important to highlight that we are the first to approach this problem from this perspective.

\subsection{Meta-Learning}\label{ML}

The concept of meta-learning (MtL) can be defined as the process of acquiring knowledge through experiences~\citep{giraud2004introduction}. Unlike traditional machine learning tasks, in MtL, multiple datasets are used to accumulate experience. In meta-learning, there are three main components~\citep{khan2020literature}: meta-features, which represent data characteristics; the meta-learner, which is tasked with inducing knowledge; and the meta-target, which is the learning target.


\begin{itemize}

    \item[] \textbf{Meta-features} are characteristics extracted from data to describe its properties~\citep{rivolli2022meta}. For each dataset, meta-features are extracted, becoming a meta-example of the meta-dataset. The meta-features are used as input for meta-learner algorithms to select the most appropriate model. There are different ways to extract features from the data~\citep{brazdil2008metalearning}. Simple and statistical, complexity-based, model-based and landmarkers are commonly used meta-features.

        \begin{itemize}
        
        \item \textit{Simple and Statistical:} These characteristics are directly extracted from datasets~\citep{reif2014automatic} and are similar to those used in classification problems, except for those involving the target variable, which needs to be adapted to consider the continuous target value. Simple meta-features include measures such as the number of examples, the number of attributes, the proportion of discrete attributes, the proportion of missing values, and the proportion of outliers. Statistical measures may include, for example, calculations of kurtosis, correlation, and covariance of instances.
        
        \item \textit{Data Complexity:} Various studies have used data complexity measures as meta-features. For classification problems, this category is investigated in~\cite{cavalcanti2012data, leyva2014set, garcia2016noise, moran2017can, garcia2018classifier}, and for regression in~\cite{lorena2018data}. Complexity measures have also been adapted for imbalanced classification problems~\citep{barella2018data, barella2020simulating, barella2021assessing}. In~\cite{lorena2018data}, the complexity measures are divided into Feature correlation measures, Linearity measures, Smoothness measures and Geometry, topology and density measures.

        \item \textit{Model-based:} This measure provides information obtained from learning models. Measures such as mean absolute value and residual variation of a linear regressor were described in~\cite{lorena2018data} as complexity-based measures, but because they are derived from models, they can also be considered model-based measures.

         \item \textit{Landmarking:} These measures are derived from the performance of baseline models on specific datasets~\citep{pfahringer2000meta}. These measures offer valuable insights into problem difficulty and data characteristics. Their simplicity and computational efficiency make them a quick and informative tool for assessing dataset challenges before engaging in more complex tasks like model selection or hyperparameter optimization. However, while landmarking measures provide a rapid overview, they may not fully capture dataset complexity or accurately evaluate the performance of more advanced models.
         
         \end{itemize}
    
    \item[] \textbf{Meta-learner} is the main component of the meta-learning framework. Meta-learner is the algorithm that models the relationship between dataset characteristics (meta-features) and candidate algorithms (meta-target). The meta-learner receives meta-features as input and recommends appropriate algorithms~\citep{khan2020literature}. The most common classifiers used as meta-models are instance-based~\citep{brazdil2003ranking, brazdil2000comparison, garouani2023autoencoder} and decision tree-based models~\citep{brazdil2000comparison, aguiar2022using, de2024meta, sousa2016active}. Instance-based models offer flexibility in adapting and are extensible for new data without need for re-learning~\citep{brazdil2003ranking}. On the other hand, decision tree-based models provide interpretability, implicitly select important features.

    \item[] \textbf{Meta-target} there are three meta-target types: Best Algorithm, Ranked List, and Multiple Algorithms~\citep{khan2020literature}. 1) Best Algorithm: aims to determine the most appropriate algorithm that achieves the highest performance on a specific task or dataset based on a given metric (e.g., F1-measure); 2) Ranked List: involves evaluating and ranking multiple algorithms based on their performance, permitting practitioners to choose from a list of top-performing options. This method presents a more comprehensive view by considering multiple candidates rather than a singular winner; 3) Multiple Algorithms: provide a set of algorithms that indicate equivalent predicted performance in the given task, indicating no significant difference in their performance. In such a scenario, the user can choose any recommended algorithm.
\end{itemize}

As far as we know, no meta-learning-based models have been applied to pipeline recommendations for imbalanced regression problems. Therefore, we explored related works that use meta-learning for pipeline recommendation to set our method apart regarding the meta-learning level. Table \ref{tab:metaworks} summarizes these methods. In~\cite{moniz2021automated} meta-learning was used for imbalanced classification problems, where a single meta-model is induced to predict the performance of workflow configurations. Their method deals with a particular case of the full model selection formulation~\citep{escalante2009particle} and a variant of the popular CASH problem~\citep{thornton2013auto}, which involves both workflow selection and hyperparameter optimization combined.

Meta-learning has also been used to suggest algorithms for data that change over time~\citep{rossi2021micro}, using one meta-model for each set of examples. Additionally, meta-learning is employed in~\cite{de2024meta} to build meta-models to automatically select the best Scaling Techniques for a given dataset and classification algorithm. In~\cite{aguiar2022using}, meta-learning is used in multi-target regression problems to recommend transformation methods and regression models, addressing three types of recommendations: i)~both algorithms independently, ii)~first the learning model then the transformation method, and iii)~first the transformation method then the regression model.


\begin{table}[!h]
\caption{Meta-learning related works}
\label{tab:metaworks}
\centering
\scalefont{0.8}
\begin{tabular}{|c|c|c|}
\hline
\textbf{Method}                                                                                    & \textbf{Recommends}                                                                           & \textbf{Meta-model type}                \\ \hline
\begin{tabular}[c]{@{}c@{}}ATOMIC\\ \citep{moniz2021automated}\end{tabular}       & \begin{tabular}[c]{@{}c@{}}Learning algorithm\\ and Resampling strategy\end{tabular}          & Single                                  \\ \hline
\begin{tabular}[c]{@{}c@{}}Micro-MetaStream\\ \citep{rossi2021micro}\end{tabular} & Learning algorithm                                                                            & One for each example                    \\ \hline
\begin{tabular}[c]{@{}c@{}}Meta-Scaler\\ \citep{de2024meta}\end{tabular}          & Scaling technique                                                                             & Multiple, one for each base model       \\ \hline
\begin{tabular}[c]{@{}c@{}}Multi-target\\ \citep{aguiar2019meta}\end{tabular}     & \begin{tabular}[c]{@{}c@{}}Multi-target method\\ and base-learner\end{tabular}                & Two models, one for each label          \\ \hline
\textbf{Meta-IR}                                                                                   & \textbf{\begin{tabular}[c]{@{}c@{}}Learning algorithm\\ and Resampling strategy\end{tabular}} & \textbf{Two models, one for each label} \\ \hline
\end{tabular}
\end{table}

In addition to the studies and methods previously presented, \cite{ghaderi2023threshold} focus on binary classification tasks, where they apply meta-learning to optimize decision thresholds to improve sensitivity for rare positive samples. Their approach aims to fine-tune classifier decision boundaries for enhanced detection. Conversely, our work differs in several aspects, addressing the challenges of imbalanced regression, which involves predicting continuous target variables with low-represented values. Our approach proposes a meta-learning framework that recommends the most effective predictive models and selects resampling strategies tailored to manage imbalance regression data.

In this way, while meta-learning techniques have been explored in various domains, our method -- Meta-learning for Imbalance Regression (Meta-IR) -- stands out as the only approach specifically tailored to address the challenge of imbalanced regression tasks by incorporating specialized mechanisms to handle imbalanced data distributions. Meta-IR employs meta-learning to build two models to select the best learning model and resampling strategy automatically in a zero-shot fashion.
Regarding the meta-learning level,~\citep{aguiar2022using} is similar to our proposal. In the same way, we induce two meta-models addressing two recommendation types that we named Independent and Chained.

\subsection{Automated Machine Learning}\label{AutoML}

Automated Machine Learning (AutoML) automates the process of constructing and selecting the best machine learning models for a given task, aiming to reduce the effort required for model building and parameter optimization~\citep{yao2018taking, zoller2021benchmark, he2021automl}. Several search strategies are employed in AutoML, such as Bayesian optimization~\citep{hutter2011sequential, snoek2012practical, garnett2023bayesian}, Evolutionary algorithms~\citep{back1997handbook, simon2013evolutionary}, Gradient-based optimization~\citep{bengio2000gradient}, Random search~\citep{bergstra2012random}, and Meta-Learning~\citep{brazdil2022metalearning}. Table~\ref{tab: RelatedWorks} describes some AutoML frameworks that use different forms of optimization. 

\begin{table}[]
\scalefont{0.9}
\caption{AutoML Related works.}
\label{tab: RelatedWorks}
\centering
\setlength{\tabcolsep}{3pt}
\begin{tabular}{|l|c|c|c|c|c|}
\hline
\textbf{Framework} & \textbf{Imb}          & \textbf{MtL}          & \textbf{Task}       & \textbf{Search Space} & \textbf{Regression Metrics}                                                                                                         \\ \hline
Auto-sklearn       & $\times$              & $\checkmark$          & Both                & DP, MS, FE, HO        & MedAE, MAE, MSE, R²                                                                                                                 \\ \hline
TPOT               & $\times$              & $\times$              & Both                & DP, MS, FE, HO        & MedAE, MAE, MSE, R²                                                                                                                 \\ \hline
LightAutoML        & $\times$              & $\times$              & Both                & DP, MS, HO            & MSE, RMSE, MAE, R²                                                                                                                  \\ \hline
H2O                & $\times$              & $\times$              & Both                & MS, HO                & \begin{tabular}[c]{@{}c@{}}MSE, RMSE, RMSLE,\\ MAE, R²\end{tabular}                                                                 \\ \hline
FLAML              & $\times$              & $\times$              & Both                & MS, HO                & MAE, MSE, R², MAPE                                                                                                                  \\ \hline
Naive AutoML       & $\times$              & $\times$              & Both                & DP, MS, FE, HO        & \begin{tabular}[c]{@{}c@{}}EV, ME, -MAE, -MSE,\\ -RMSE, -MSLE, -MedAE, R2,\\ -MPD, -MGD, -MAPE, D2-AE,\\  D2-PB, D2-TW\end{tabular} \\ \hline
\textbf{Meta-IR}   & \textbf{$\checkmark$} & \textbf{$\checkmark$} & \textbf{Regression} & \textbf{MS, RS}       & \textbf{SERA, F1-scoreR}                                                                                                            \\ \hline
\end{tabular}
\end{table}

Auto-sklearn\footnotemark[1]~\citep{feurer2015efficient} facilitates the exploration of potential solutions by leveraging meta-learning to warm start the Bayesian optimization. The Bayesian optimization determine the optimal pipeline configuration. TPOT (Tree-based Pipeline Optimization Tool)~\citep{OlsonGECCO2016} utilizes genetic programming to optimize machine learning pipelines. Each pipeline represents a sequence of preprocessing steps, feature transformations, and machine learning models. However, the method's primary limitation is the substantial computational resources needed for the optimization process, making it impractical for large datasets or limited computing environments. H2O AutoML~\citep{H2OAutoML20} use a blend of quick random search and stacked ensembles to produce highly competitive outcomes.

\footnotetext[1]{The version of Auto-sklearn used in this work is 1.0, as version 2.0 does not include regression models.}

The Fast and Lightweight AutoML Library (FLAML)~\citep{wang2021flaml} uses Estimated cost for improvement (ECI) to learner choices and adopted a randomized direct search method~\citep{wu2021frugal} to perform cost-effective optimization for cost-related hyperparameters. In LightAutoML framework~\citep{vakhrushev2021lightautoml} Tree-structured Parzen Estimators~\citep{bergstra2011implementations} are employed for model fine-tuning. Additionally, warm-starting and early stopping techniques are utilized for optimizing linear models through grid search.

Naive AutoML~\citep{mohr2023naive} aims to offer a basic method that acts as a starting point for comparing with more complicated ones. Unlike other approaches that view process steps as interconnected, Naive AutoML pretends independence in how other components are chosen. It then selects the best algorithm to construct the final process for each step. This method reduces the amount of searching needed, making it faster and more efficient. In terms of execution time, when compared to Meta-IR, Naive AutoML follows a more iterative approach, exploring different components and hyperparameters for each pipeline step. This process can be time-consuming, necessitating multiple iterations to optimize the model. For example, with 6 learning models, this method first searches for the best model among the 6 before optimizing. Conversely, Meta-IR takes a zero-shot method, recommending both a learning model and a resampling strategy based on the problem's characteristics.


Regarding the search space, Auto-sklearn, TPOT, Naive AutoML address the entire pipeline, recommending Data preprocessing (DP), Model Selection (MS), Feature Engineering (FE), and Hyperparameter Optimization (HO). The LightAutoML recommends nearly the entire pipeline except for Feature Engineering. In contrast, H2O and FLAML specifically concentrate on Model Selection and Hyperparameter Optimization.  It is worth noting that none of these methods include resampling strategies for regression in their search space. Conversely, Meta-IR recommends the learning model and resampling strategy for imbalanced regression problems.


Another critical aspect to consider is that the approaches described focus on defining pipelines without considering the imbalance problem. As a result, these frameworks do not incorporate specific metrics for imbalanced regression. They employ standard metrics like MedAE, MSE, MAE and R² as evaluation metrics. These metrics have limitations when used for evaluation in imbalanced regression problems, as they can deceive the user when the focus is on the accuracy of rare values~\citep{moniz2014resampling}, as they do not consider the relevance of each example. Thus, we hypothesize that employing specialized metrics and methods for imbalanced regression is crucial for improving the performance and reliability of machine learning models in these scenarios. By addressing the relevance of each example and focusing on the accuracy of predictions for rare values, it is possible to achieve a more accurate and fair assessment of model performance. 

In this way, our proposal -- Meta-IR -- differs from previous work in multiple perspectives: (1) As the first framework to deal with imbalanced regression, we recommend learning models and resampling strategies; (2) It is different in how to evaluate the pipelines, using specific metrics for imbalanced regression; (3) Our method is a zero-shot method based on meta-learning, enabling an efficient and adaptive recommendation phase; (4) We employ two different ways in training phase (Independent and Chained).

\section{Proposed Method}\label{Meta-IR}

This section presents a recommendation system that suggests a pipeline (learning models and resampling strategies) for imbalanced regression using meta-learning. Section~\ref{sec:problem-definition} introduces the problem and Section~\ref{sec:meta-IRb} describes all phases of the proposed method.

\subsection{Problem Definition}
\label{sec:problem-definition}

The performance of learning models in imbalanced problems can be improved by using strategies that balance the datasets. Therefore, the problem of defining a pipeline is addressed in this work through a meta-learning perspective. The meta-features of the data are used as a criterion for making recommendations. Given a set of datasets $\mathbb{D} = \{\mathbf{D}_1, \mathbf{D}_2, \mathbf{D}_3, \ldots, \mathbf{D}_m\}$, a set of learning algorithms $\mathbb{L} = \{l_1, l_2, l_3, \ldots, l_s\}$ and a set of resampling strategies $\mathbb{R} = \{r_1, r_2, r_3, \ldots, r_n\}$, the task is defined as follows:

\begin{itemize}[noitemsep]

    \item \textbf{Meta-problem}: the task is to predict the best learning algorithm ($l \in \mathbb{L}$) and the best resampling strategy ($r \in \mathbb{R}$) for an imbalanced regression dataset $\mathbf{D}_i$, $i=\{1,2,\ldots,m\}$, based on its meta-features ($\mathbf{f}_{\mathbf{D}_i}$).

    \item \textbf{Meta-feature}: each dataset ($\mathbf{D}_i$) is represented by a meta-feature vector $\mathbf{f}_{\mathbf{D}_i} = \{f_1^i, f_2^i, \ldots, f_k^i\}$ composed of $k$ features.

    \item \textbf{Meta-targets}: the meta-targets per dataset $\mathbf{D}_i$ are $[l_i,r_i]$, where $l_i \in \mathbb{L}$ is the learning algorithm and $r_i \in \mathbb{R}$ is the resampling algorithm that are the best fit for $\mathbf{D}_i$.  
    
    \item \textbf{Meta-dataset}: the meta-dataset $\mathbf{M}$ stores the meta-features and the meta-targets of each dataset $\mathbf{D}_i$. $\mathbf{M}$ has $m$ tuples $(\mathbf{f}_{\mathbf{D}_i}, l_i, r_i$), where $\mathbf{f}_{\mathbf{D}_i}$ is the meta-feature vector, and $l_i$ and $r_i$ are the meta-targets that represent the learning and the resampling algorithms respectively. Thus, $\mathbf{M}$ has $m$ rows and $(k+2)$ columns; $m$ datasets, $k$ features, and two meta-targets.
    
    \item \textbf{Meta-classifiers}: two meta-classifier ($\lambda_L$ and $\lambda_R$) are trained using the meta-dataset $\mathbf{M}$. The meta-classifier $\lambda_L$ predicts the best learning algorithm $l \in  \mathbb{L}$, while the meta-classifier $\lambda_R$ predicts the best resampling algorithm $r \in \mathbb{R}$ per dataset.
\end{itemize}

\subsection{Meta-learning for Imbalanced Regression (Meta-IR)}
\label{sec:meta-IRb}

The Meta-IR method is a meta-learning-based solution to deal with imbalanced regression problems. The main objective of Meta-IR is to facilitate the choice of the learning model and preprocessing strategy for the datasets, given that the conventional workflow for making this choice is extensive. 

\begin{figure}[htbp]
    \centering
    \includegraphics[width=1\textwidth]{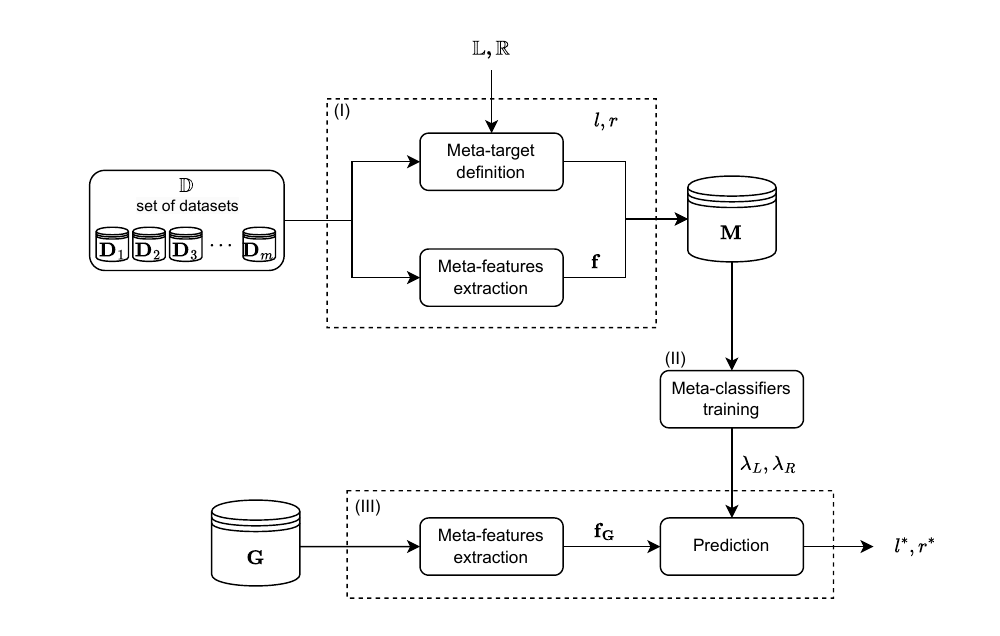}
    \caption{The proposed Meta-IR framework. (I) \textbf{Meta-dataset construction:} the meta-features ($\mathbf{f}$) are extracted, and the pipelines are evaluated to construct the meta-dataset $\mathbf{M}$. (II) \textbf{Meta-classifiers training:} the two meta-classifiers ($\lambda_L$ and $\lambda_R$) are trained. (III) \textbf{Recommendation:} Given the meta-features ($\mathbf{f_G}$) of an imbalanced regression dataset $\mathbf{G}$, this phase recommends a learning model ($l^* \in \mathbb{L}$) and a resampling strategy ($r^* \in \mathbb{R}$).}
    
    
    \label{meta}  
\end{figure}

Meta-IR is divided into three phases: (I) Meta-dataset construction, (II) Meta-classifiers training, and (III) Recommendation, as shown in Figure~\ref{meta}. In the first phase (Meta-dataset construction), the meta-dataset $\mathbf{M}$ is constructed by extracting meta-features and defining the meta-target through pipeline evaluation. Then, the meta-models $\lambda_L$ and $\lambda_R$ are trained in the Meta-classifiers training phase to recommend the learning model and resampling strategy, respectively. Finally, in the Recommendation Phase, given an imbalanced regression dataset ($\mathbf{G}$), its meta-features are extracted, and the meta-classifiers generated in the previous phase perform the prediction, recommending a learning model and resampling strategy that are supposedly the best for that dataset. The following sections detail each phase.

\subsubsection{Meta-dataset construction phase}

This phase, described in Algorithm \ref{alg:metadata}, has two modules: Meta-feature extraction and meta-target definition. In the Meta-features extraction, each $\mathbf{D}_i \in \mathbb{D}$ is represented by a vector ${\mathbf{f_D}}_i = \{f_1^i, f_2^i, \ldots, f_k^i\}$ with $k$ meta-features. These meta-features include both simple and complexity-based characteristics, aiming to capture diverse aspects of the datasets (refer to Appendix B, Table \ref{tab:mf}, for details). The extraction of meta-features is implemented in line 16 using the function \texttt{ExtractMetaFeatures}.

In parallel, the Meta-target definition module evaluates a total of $\lvert \mathbb{L}\rvert \times \lvert \mathbb{R}\rvert$ pairs of of learning algorithms ($l_i \in \mathbb{L}$) and resampling techniques ($r_i \in \mathbb{R}$) for each dataset $\mathbf{D}_i$, selecting the pair $(l_i,r_i)$ that optimizes a given performance metric. This process is detailed in lines 7–15 of Algorithm \ref{alg:metadata}, where nested loops iterate through all combinations of algorithms and strategies, applying the respective resampling technique (line 9), training the model (line 10), and evaluating its performance (line 11). The selected best pair is defined based on conditions specific to the metric (line 15). 

The outputs of these two modules are concatenated to compose the meta-dataset $\mathbf{M}$. As implemented in line 17, $\mathbf{M}$ consists of $m$ tuples ($\mathbf{f_D}_i, l_i, r_i$), one for each $\mathbf{D}_i$.

\begin{algorithm}[h]
\caption{Meta-Dataset Construction Phase}
\label{alg:metadata}
\begin{algorithmic}[1]
\State \textbf{Input:}

$\mathbb{D} = \{\mathbf{D}_1, \mathbf{D}_2, \mathbf{D}_3, \ldots, \mathbf{D}_m\}$ - Datasets

$\mathbb{L} = \{l_1, l_2, l_3, \ldots, l_s\}$ - Learning models

$\mathbb{R} = \{r_1, r_2, \dots, r_q\}$ - Resampling strategies

$\text{metric}$ - Evaluation metric

\State \textbf{Output:}

$\mathbf{M}$ - Meta-dataset

\State $\mathbf{M} \gets \emptyset$ \Comment{Initialize the meta-dataset}

\For{each dataset $\mathbf{D}_i \in \mathbb{D}$}
    \State $\mathcal{T}, \mathcal{S} \gets \text{Split}(\mathbf{D}_i)$ \Comment{Split Dataset $\mathbf{D}_i$ in Train $\mathcal{T}$ and Test $\mathcal{S}$ sets}
    
    \State $C \gets \emptyset$ \Comment{Initialize candidates set}

    \For{each model $l_j \in \mathbb{L}$}
        \For{each strategy $r_k \in \mathbb{R}$}
            \State $\mathcal{T}' \gets \text{Apply}(r_k, \mathcal{T})$ \Comment{Apply resampling strategy to the train set}
            \State $m \gets \text{Train}(l_j, \mathcal{T}')$ \Comment{Train the model}
            \State $\text{score} \gets \text{Evaluate}(m, \mathcal{S}, \text{metric})$ \Comment{Evaluate the model on the test set $\mathcal{S}$}
            
            \State  $C \gets C \cup (\text{score}, l_j, r_k)$
        \EndFor
    \EndFor

    \State $(\text{best\_score}, l_i, r_i) \gets \text{argmax}_{(score, l, r) \in C}
        \begin{cases} 
            score & \text{if metric = F1-ScoreR} \\
            -score & \text{if metric = SERA} 
        \end{cases}$

    \State $\mathbf{f_D}_i \gets \text{ExtractMetaFeatures}(\mathbf{D}_i)$
    \State $\mathbf{M} \gets \mathbf{M} \cup (\mathbf{f_D}_i, l_i, r_i)$
\EndFor

\State \Return $\mathbf{M}$

\end{algorithmic}
\end{algorithm}

\subsubsection{Meta-classifiers training phase}

In this phase, described in Algorithm \ref{alg:training}, we train two meta-classifiers using a learning algorithm \( \lambda \), such as Random Forest, Bagging, or XGBoost. This learning algorithm is responsible for capturing patterns in the meta-dataset and creating classifiers capable of providing effective recommendations. The trained meta-classifiers are \( \lambda_L \), which recommends a learning model (\( l_i \in \mathbb{L} \)), and \( \lambda_R \), which indicates the resampling strategy (\( r_i \in \mathbb{R} \)) for an imbalanced regression dataset (\( \mathbf{D}_i \)). We proposed two training strategies: Independent and Chained.

\begin{figure}[htp]
  \centering
  \subfigure[Independet]{\includegraphics[scale=0.8]{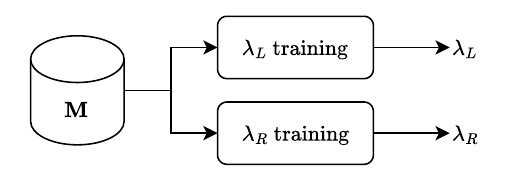}}
  \subfigure[Model first]{\includegraphics[scale=0.7]{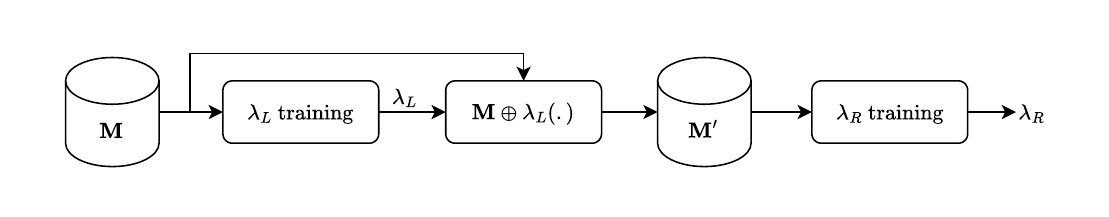}}
  \subfigure[Strategy first]{\includegraphics[scale=0.7]{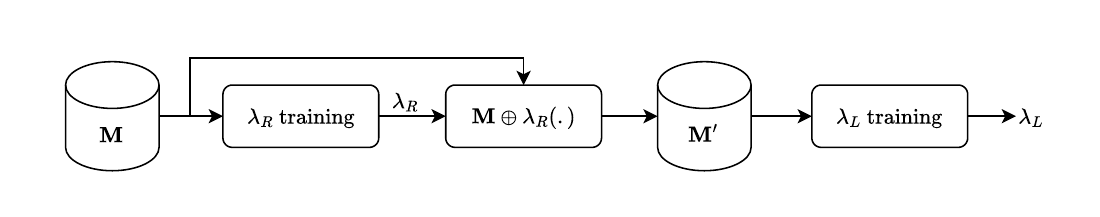}}
  \caption{Training of the meta-classifiers $\lambda_L$ and $\lambda_R$.}
  \label{fig:meta-classifier-training-mode}
\end{figure}

\textbf{Independent}. Let $\lambda_L$ and $\lambda_R$ be learning models that recommend which learning algorithm and resampling strategy should be used for $\mathbf{D}_i$. The $\lambda_L$ and $\lambda_R$ models are trained using $\mathbf{M}$ to predict $\mathbf{D}_i$'s best learning algorithm and best resampling strategy. Thus, the final recommendation combines the learning algorithm recommended by $\lambda_L$ and the resampling strategy recommended by $\lambda_R$. The training process for this strategy is implemented in lines 6–9 of Algorithm \ref{alg:training}, where both meta-classifiers are trained independently. The assumption is that the choice of the resampling strategy and learning model is independent and can be performed separately; the best resampling strategy can be chosen independently of the chosen learning model and vice versa. Figure~\ref{fig:meta-classifier-training-mode}(a) illustrates this training strategy.
    
\textbf{Chained}. There are two alternatives for the chaining order: Model first~(Figure~\ref{fig:meta-classifier-training-mode}(b)), and Strategy first~(Figure~\ref{fig:meta-classifier-training-mode}(c)). In the Model First approach, $\lambda_L$ is trained using the meta-features in $\mathbf{M}$, as implemented in line 11, and the model's recommendation $l \in \mathbb{L}$ is added to the meta-features (lines 12-14). After that, the model $\lambda_R$ is trained using the updated meta-features to recommend the resampling strategy (line 16). The assumption is that the choice of the learning model helps to select the resampling strategy. In the Strategy First approach, $\lambda_R$ is trained using $\mathbf{M}$, as shown in line 19, and the model's recommendation $r \in \mathbb{R}$ is added to the meta-features (lines 20-22). The $\lambda_L$ model is then trained to recommend the learning model using the updated meta-features (line 24). The assumption is that the choice of resampling strategy helps to select the learning model.

\begin{algorithm}[h]
\caption{Meta-Classifiers Training Phase}
\label{alg:training}
\begin{algorithmic}[1]
\State \textbf{Input:}

$\mathbf{M} = [(\mathbf{f_{D_1}}, l_1, r_1), (\mathbf{f_{D_2}}, l_2, r_2), \dots, (\mathbf{f_{D_m}}, l_m, r_m)]$, where ${\mathbf{f_D}} = \{f_1, f_2, \ldots, f_k\}$

$\lambda$ - Learning Algorithm

\textit{approach} - Training approach

\State \textbf{Output:}

$\lambda_L$ - Meta-classifier for learning models

$\lambda_R$ - Meta-classifier for resampling strategies

\State $x \gets [\mathbf{f_{D_1}}, \mathbf{f_{D_2}},\dots,\mathbf{f_{D_m}}]$    
\State $y_L \gets [l_1, l_2,\dots,l_m]$
\State $y_R \gets  [r_1, r_2,\dots,r_m]$

\If{\textit{approach} = ``Independent"} \Comment{Train $\lambda_L$ and $\lambda_R$ independently}

    \State $\lambda_L \gets \text{Train}(\lambda, x, y_L)$
    \State $\lambda_R \gets \text{Train}(\lambda, x, y_R)$

    \State \Return $\lambda_L$, $\lambda_R$

\ElsIf{\textit{approach} = ``Model First"} \Comment{Train $\lambda_L$ first and use its predictions as meta-features to train $\lambda_R$}
    \State $\lambda_L \gets \text{Train}(\lambda, x, y_L)$
    \For{each $\mathbf{f_{D_i}} \in x$} \Comment{Update meta-features with $\lambda_L$ recommendation}
        \State $\mathbf{f_{D_i}} \gets \mathbf{f_{D_i}} \cup \lambda_L(\mathbf{f_{D_i}})$
    \EndFor
    \State $x \gets [\mathbf{f_{D_1}}, \mathbf{f_{D_2}},\dots,\mathbf{f_{D_n}}]$
    \State $\lambda_R \gets \text{Train}(\lambda, x, y_R)$

    \State \Return $\lambda_L$, $\lambda_R$

\ElsIf{\textit{approach} = ``Strategy First"} \Comment{Train $\lambda_R$ first and use its predictions as meta-features to train $\lambda_L$}

    \State $\lambda_R \gets \text{Train}(\lambda, x, y_R)$
    \For{each $\mathbf{f_{D_i}} \in x$} \Comment{Update meta-features with the $\lambda_R$ recommendation}
        \State $\mathbf{f_{D_i}} \gets \mathbf{f_{D_i}} \cup \lambda_R(\mathbf{f_{D_i}})$
    \EndFor
    \State $x \gets [\mathbf{f_{D_1}}, \mathbf{f_{D_2}},\dots,\mathbf{f_{D_n}}]$
    \State $\lambda_L \gets \text{Train}(\lambda, x, y_L)$

    \State \Return $\lambda_L$, $\lambda_R$
\EndIf
\end{algorithmic}
\end{algorithm}

\subsubsection{Recommendation Phase}
\label{procedimentos}

Given an imbalanced regression dataset $\mathbf{G}$, this phase recommends a learning model ($l^* \in \mathbb{L}$) and a resampling strategy ($r^* \in \mathbb{R}$). The ``Meta-features extraction'' module (Phase III in Figure~\ref{meta}) represents $\mathbf{G}$ as a vector of meta-features $\mathbf{f_G}$. This process is detailed in line 3 of Algorithm \ref{alg:rec}, where the dataset is converted into its meta-representation. After, $\mathbf{f_G}$ is used as input to the meta-classifiers $\lambda_L$ and $\lambda_R$, which return the the recommended learning model ($l^*$) and the resampling strategy ($r^*$). These recommendations are computed in lines 4–5. The type of recommendation is the best algorithm, where only the model and strategy with the highest performance are recommended. This process ensures that the choices of $l^*$ and $r^*$ are driven by the specific characteristics of the dataset $\mathbf{G}$, captured through its meta-features.

\begin{algorithm}[h]
\caption{Recommendation Phase}
\label{alg:rec}
\begin{algorithmic}[1]
\State \textbf{Input:}

$G$ - New dataset

$\lambda_L$ - Trained meta-classifier for learning models

$\lambda_R$ - Trained meta-classifier for resampling strategies

\State \textbf{Output:} Recommended learning model $l^*$ and resampling strategy $r^*$

\State $\mathbf{f_G} \gets \text{ExtractMetaFeatures}(G)$

\State $l^* \gets \lambda_L(\mathbf{f_G})$
\State $r^* \gets \lambda_R(\mathbf{f_G})$

\State \Return $l^*$, $r^*$
\end{algorithmic}
\end{algorithm}

\section{Experimental Methodology}\label{Methodology}

\subsection{Datasets}
\label{sec:datasets}

The datasets were taken from the OpenML repository~\citep{vanschoren2014openml}. After searching only for regression problems on OpenML, a total of 660 datasets was returned. Subsequently, the datasets having less than 2.0\% of rare cases, as defined by the relevance function outlined in Section \ref{sec:relevancefunction}, were removed, resulting in 200 datasets. Additionally, we included 18 datasets from~\cite{moniz2017evaluation}.

Thus, the evaluation study is performed on $\vert\mathbb{D}\vert = 218$ datasets. An important aspect of these datasets is their diversity in terms of number of examples varying between 27 and 20.640, the number of features ranging from 5 to 1024, and percentage of rare examples varying from $2.0\%$ to $39.6\%$. This diversity is crucial in the meta-learning process, as it allows for the generalization of the model, making it more flexible and capable of dealing with a greater variety of datasets (refer to Appendix A, Table~\ref{tab:datasets}, for more details).

\subsection{Meta-features}
\label{sec:meta-features}

Each dataset $\mathbf{D}_i$ in the meta-dataset is represented by a vector of meta-features $\mathbf{f_D}_i$. Forty-three meta-features were considered, with measures from different categories: simple information about the dataset and complexity measures proposed in~\cite{lorena2018data}. The simple category includes the number of examples, the number of attributes, the number of rare cases, and the percentage of rare cases. Regarding complexity measures, data set distribution, correlation between attributes and targets, performance metrics related to linear regression, and data smoothness are considered (refer to Appendix B for details). The ECoL package~\citep{lorena2018data, lorena2019complex} was used to extract complexity measures.
\subsection{Learning Algorithms}
\label{sec:learningalg}

The learning algorithms set $\mathbb{L}$ is composed of six models from different families (single and ensemble): Bagging~(BG)\footnotemark[1], Decision Tree~(DT)\footnotemark[1], Multilayer Perceptron~(MLP)\footnotemark[1], Random Forest~(RF)\footnotemark[1], Support Vector Machine~(SVM)\footnotemark[1], and XGBoost~(XG)\footnotemark[2]. These models were previously employed in~\cite{avelino2024resampling}, where their sensitivity to the selection of resampling strategies was demonstrated. For all models, the default hyperparameters were used, as per the definitions provided by the respective packages and versions employed in this study.

\footnotetext[1]{\url{https://scikit-learn.org/} - version 1.0.2.}
\footnotetext[2]{\url{https://xgboost.readthedocs.io/} - version 0.0.9.}

\subsection{Resampling strategies}
\label{sec:learningalg}

The resample strategies set $\mathbb{R}$ comprises six strategies plus no resampling (NONE). The strategies, previously described in Section~\ref{sec:strategies}, are SmoteR~(SMT)\footnotemark[3], Random Over-sampling~(RO)\footnotemark[3], Random Under-sampling~(RU)\footnotemark[3], Introduction of Gaussian Noise~(GN)\footnotemark[3], SMOGN~(SG)\footnotemark[4], and WEighted Relevance-based Combination Strategy~(WERCS)\footnotemark[5]. For all resampling strategies, the default parameters were used, as defined by the respective packages and versions employed in this study.

\footnotetext[3]{\url{https://pypi.org/project/ImbalancedLearningRegression/} - version 0.0.1.}
\footnotetext[4]{\url{https://pypi.org/project/smogn/} - version 0.1.2}
\footnotetext[5]{\url{https://pypi.org/project/resreg/} - version 0.2}

\subsection{Meta-Models}
\label{sec:meta-classifier}

We select classification algorithms for the meta-models, focusing on models based on Decision Trees. These algorithms are renowned for their interpretability, widespread adoption in classification tasks, and consistently strong performance across diverse problem domains~\citep{hastie2009elements}. Besides, Decision Trees requires no scaling data transformation~\citep{amorim2023}. Our chosen models include Random Forest~(RF)~\citep{breiman2001random}, Extra trees~(ET)~\citep{geurts2006extremely}, Bagging~(BG)~\citep{breiman1996bagging}, k-Nearest Oracles-Eliminate (Knora-E)~\citep{ko2008dynamic}, k-Nearest Oracles-Union (Knora-U)~\citep{ko2008dynamic}, XGBoost~(XG)~\citep{chen2016xgboost} e DES-MI~\citep{garcia2018dynamic}.

These classification models have also been used in other meta-learning tasks~\citep{aguiar2019meta, de2024meta, sousa2016active}. We employed the default configurations for all models. After comparing these models, we found that Random Forest (RF) outperformed the others, achieving superior results (refer to Appendix C, Tables~\ref{tab:meta-modelsF1} and \ref{tab:meta-modelsSERA}, for detailed comparisons). As a result, we selected this model for the Meta-IR analyses.

\subsection{Evaluation Methodology}
\label{sec:evaluationmethodology}

The Meta-IR is evaluated using the leave-one-dataset-out method. This cross-validation technique involves using each dataset as the test set in an iteration, and the other datasets are used to compose the meta-dataset~\citep{vanschoren2019meta}. Since we have 218 datasets, for each iteration, the meta-dataset is constructed using $\vert\mathbb{D}\vert = 217$ datasets, while the test $\mathbf{G}$ has only one dataset.

The proposal is evaluated at the meta-level and base-level. The meta-level focuses on evaluating the meta-classifiers for recommending the resampling strategy and regressor, with the two proposed meta-classifier strategies, Independent and Chained, also being evaluated. The evaluation at this level utilizes the F1-macro (classification) metric. At the base-level, performance is evaluated on the recommendations of the meta-classifiers from the meta-level, using the F1-scoreR~\citep{torgo2009precision} and Squared error-relevance area (SERA) metric~\citep{ribeiro2020imbalanced}. These metrics were chosen because they aim to effectively evaluate model performance for extreme value predictions while being robust to model bias. Furthermore, we also analyze the proposed method's execution time and compare it to the state-of-the-art AutoML methods. 

The proposed method is compared at the base-level with two baseline methods: Random and Majority. The Random model randomly selects a class label from the available labels. Let $\mathbb{L}$ and $\mathbb{R}$, where $\mathbb{L}$ contains the learning models and $\mathbb{R}$ the resampling algorithms. For each instance of the meta-dataset, the random recommendation model randomly chooses one instance from $\mathbb{L}$ and one from $\mathbb{R}$. The Majority model selects the class that appears most frequently. In other words, this model chooses the class label most frequently in $\mathbb{L}$ and $\mathbb{R}$. For the F1-scoreR metric, the resampling strategy most represented is WERCS, and the learning model is DT. For SERA is WERCS and RF (refer to Appendix D, Figure~\ref{fig:frequency},  for details). These baselines are commonly employed to emphasize the necessity of a recommendation system~\citep{brazdil2008metalearning}.

At the meta-level, Meta-IR is evaluated from three perspectives: i)~Comparison with each pipeline; ii)~Comparison with AutoML frameworks such as Auto-sklearn, H2O, TPOT, FLAML, LightAutoML, NaiveAutoML and the baselines Random and Majority; iii)~Meta-IR as a preprocessing step for AutoML frameworks.

The comparison of each pipeline demonstrates that no single combination is capable of achieving more wins than Meta-IR. This indicates that there is no one-size-fits-all solution for all datasets. This highlights the importance of our method, which determines the pipeline according to the characteristics of the problem. On the other hand, the comparison with AutoML frameworks aims to demonstrate that Meta-IR achieves better results than these methods in imbalanced regression scenarios. Finally, by using Meta-IR output as a preprocessing step for AutoML frameworks, we highlight the importance of incorporating resampling strategy recommendation into AutoML methods to address imbalanced problems.

\section{Results and Discussion}\label{Results and Discussions}

The evaluation of Meta-IR is divided into three perspectives: Meta-Level Analysis, Base-level Analysis and Execution time analysis. In the Meta-Level Analysis (Section \ref{sec:meta-level}), the meta-models performance is evaluated and compared with two baseline approaches: Random and Majority. In Section \ref{sec:base-level}, we present the Base-level Analysis, where we evaluate the recommended learning models and resampling strategies. In this analysis, at the base-level, we compare the recommendation of Meta-IR with Random, Majority, and the AutoML frameworks Auto-sklearn, H2O, TPOT, FLAML, LightAutoML and NaiveAutoML. Furthermore, in this section, we evaluate the Meta-IR as a preprocessing step for AutoML frameworks. The third analysis presented in Section~\ref{sec:time-analysis}, where we consider the execution time of Meta-IR and each AutoML framework. Finally, in Section~\ref{sec:meta-featuresanalysis}, an analysis of the meta-features is conducted.

\subsection{Meta-level analysis}
\label{sec:meta-level}

\begin{figure}[!h]
  \centering
  \subfigure[F1-scoreR]{\includegraphics[scale=0.05]{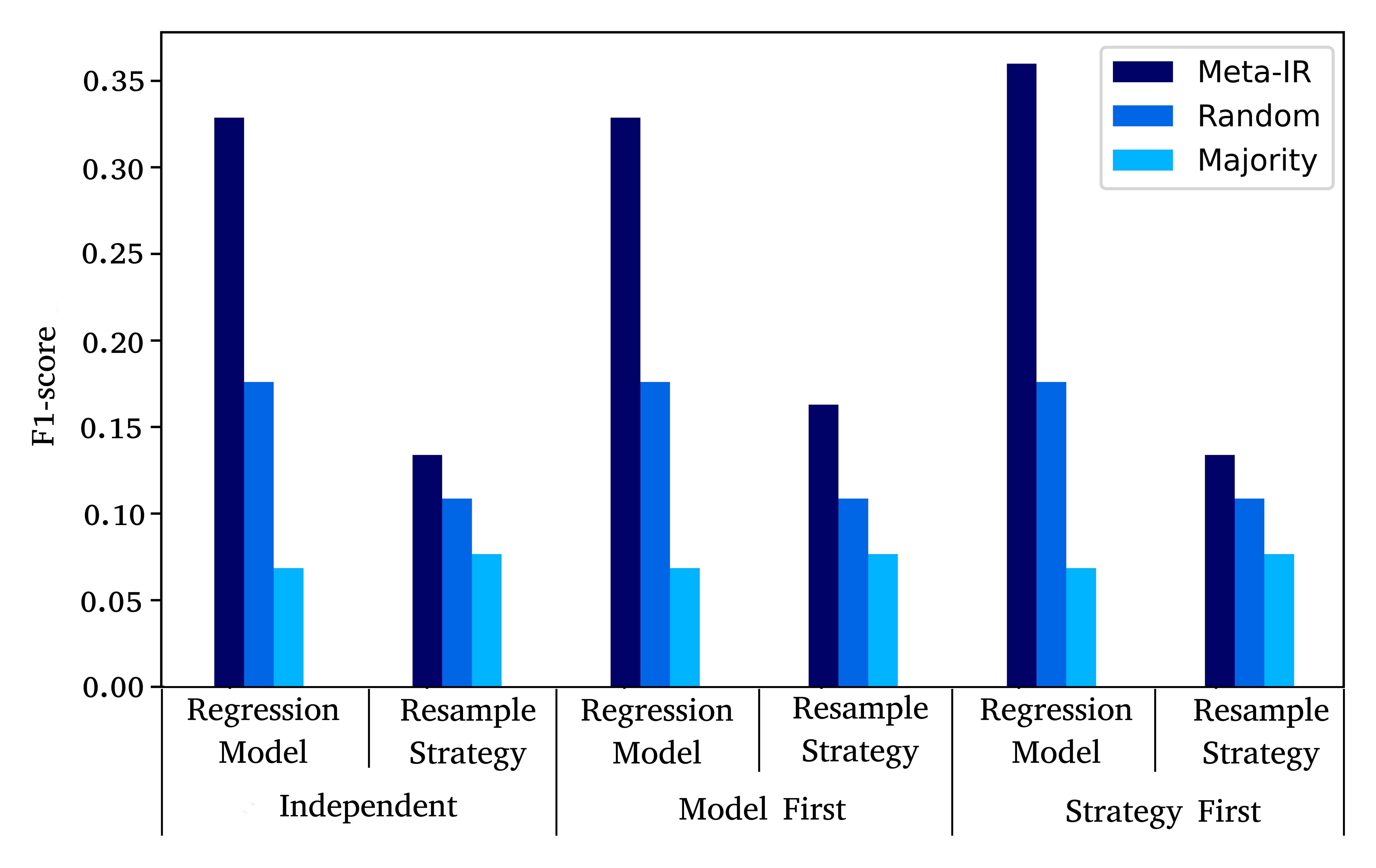}}
  \subfigure[SERA]{\includegraphics[scale=0.05]{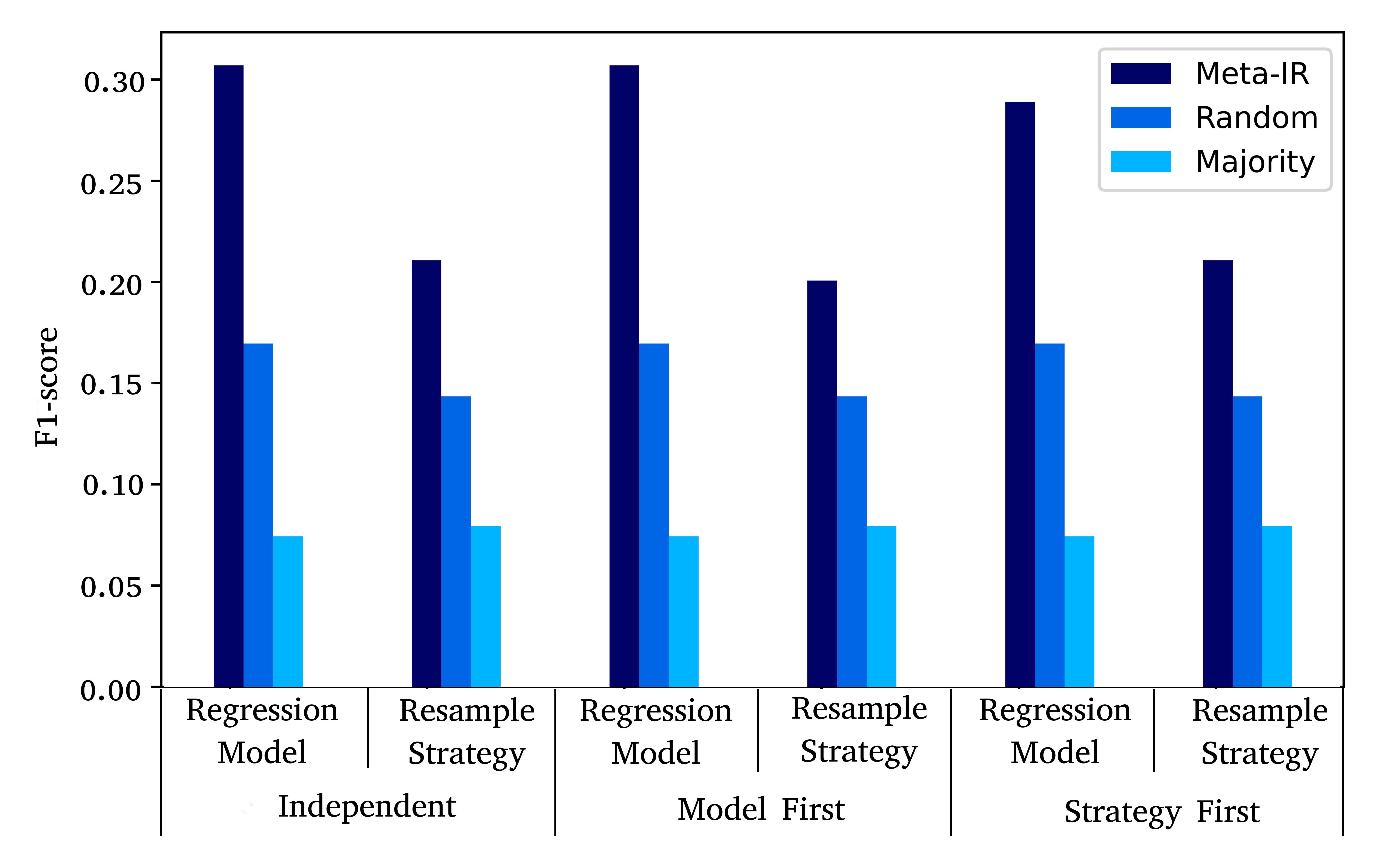}}
  \caption{Meta-IR performance for each type of training (Independent, Model First, and Strategy First) compared with the baselines Random and Majority.}
  \label{fig:performance}
\end{figure}

The meta-level refers to evaluating the meta-model performance, that is, the effectiveness exhibited by the meta-model in predicting the meta-target for the datasets. The Meta-IR performance is compared with the performance of Random and Majority models, considering the F1-macro metric. These results are illustrated in Figure~\ref{fig:performance}~(a-b), considering F1-scoreR and SERA metrics as optimization functions. The optimization function refers to a specific metric used to guide the optimization process. It defines the objective that the model aims to maximize (F1-scoreR) or minimize (SERA) during the learning process. The F1-scoreR and SERA metrics were chosen as optimization functions because they are well-suited for capturing the performance of models in the imbalanced regression problem.

The results are presented for each recommendation procedure. Meta-IR obtained better results recommending the learning model and resampling strategy than Random and Majority for all types of training in both optimization functions.

Another analysis we can extract from Figure~\ref{fig:performance} is about the training strategy. It is observed that the order in which the model and resampling strategy are recommended affects the recommendation. For the F1-scoreR, the Independent model achieved a lower F1-macro score compared to the Chained-trained model. Specifically, the Independent model attained an F1-macro score of 0.33 for recommending the learning model, while the Chained model (Strategy first) achieved 0.36, indicating that recommending the resampling strategy first led to an increase in the F1-macro score for recommending the learning model. Regarding the recommendation of the resampling strategy, the Independent model scored 0.14, while the Chained model (Model first) scored 0.16, showing that recommending the learning model first led to an increase in the F1-macro score for recommending the resampling strategy.

On the other hand, when we evaluate considering the SERA metric, the behavior is different. The use of the Chained model for training has a negative impact on the recommendation of the learning model and resampling strategy. The Independent model attained an F1-macro score of 0.31 for recommending the learning model, while the Chained model (Strategy first) achieved 0.21. Regarding the recommendation of the resampling strategy, the Independent model scored 0.21, while the Chained model (Model first) scored 0.20.

In conclusion, the analysis emphasizes the importance of the training strategy on the performance of the Meta-IR meta-model, especially in how the order of recommendations affects the outcomes. The Chained approach (Strategy first and Model first) improved the F1-macro score for recommending the learning model and the resampling strategy, but it was less effective when evaluated using the SERA metric. These findings highlight the importance of selecting an appropriate training strategy based on the evaluation metric used, as it can significantly impact the effectiveness of model and resampling strategy recommendations for addressing imbalanced regression problems.


\subsection{Base-level analysis}
\label{sec:base-level}

The Base-level Analysis assesses the learning models and resampling strategies recommended by Meta-IR. We conducted the following analysis: (1)~\mbox{Meta-IR} compared to each combination of sampling strategy and learning model (Section~\ref{sec:com-pipeline}); 2)~Meta-IR compared to Random, Majority and the AutoML frameworks: Auto-sklearn, H2O, TPOT, FLAML, LightAutoML and NaiveAutoML (Section~\ref{sec:comp-automl}).

\subsubsection{Pairwise comparisons with each pipeline}
\label{sec:com-pipeline}

\begin{figure}[!h]
  \centering
  \subfigure[F1-scoreR]{\includegraphics[scale=0.3]{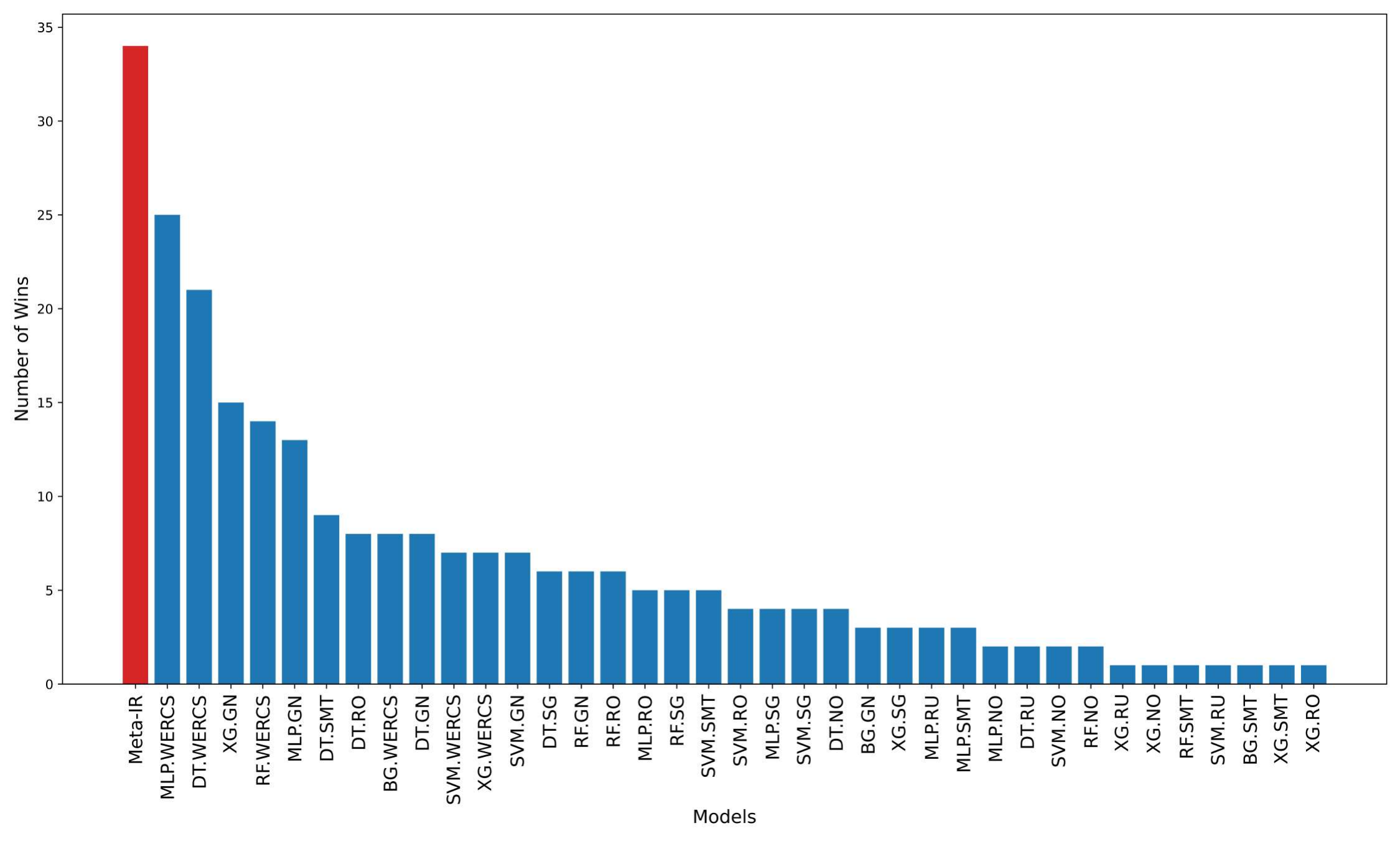}}

  \subfigure[SERA]{\includegraphics[scale=0.3]{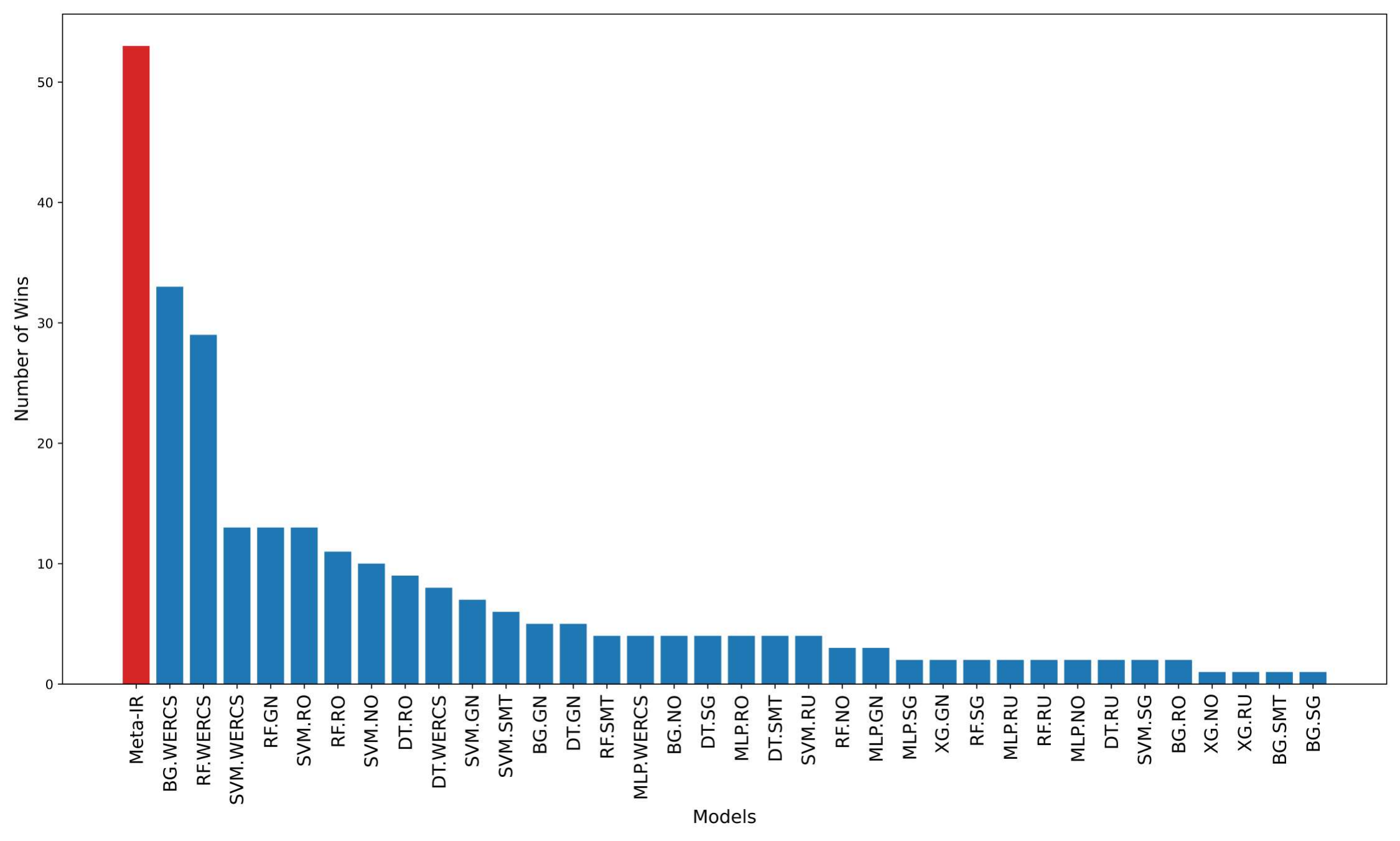}}
  \caption{Number of wins over all datasets: Meta-IR \textit{versus} each pipeline (resampling strategy and regressor).}
  \label{fig:winsMeta}
\end{figure}

Figure~\ref{fig:winsMeta} shows how many times each model achieved the best result among all evaluated combinations of resampling strategies and learning algorithms. These results represent the oracle, showcasing the best possible outcomes for each dataset. The number of wins summation per figure is equal to 218, which is the number of datasets in the used corpus. The results indicate that Meta-IR outperformed all other strategies for both metrics, indicating that no single combination consistently achieves a significant number of wins compared to Meta-IR. It is important to note that the version of Meta-IR applied in this analysis is Model First, chosen because it is the most effective approach overall.
Therefore, these outcomes highlight the potential of the Meta-IR model and the importance of integrating Meta-IR into the pipeline recommendation process for imbalanced regression problems.

Tables~\ref{tab:p-valueF1}~and~\ref{tab:p-valueSERA} show the p-values of the Wilcoxon signed rank test, illustrating the significance of performance discrepancies between Meta-IR and each specific combination. There are 42 combinations in total (6 learning algorithms $\times$ 7 resampling strategies).  For the F1-scoreR, Meta-IR achieved statistically significantly better results, except for DT.GN and DT.RO. It is crucial to highlight that the Decision Tree Regressor (DT) was the model that appears more frequently as the best one, and it was defined as the Majority model as described in Section~\ref{sec:evaluationmethodology}. Meta-IR consistently performed significantly better regarding the SERA metric, with p-values $< 0.001$.  These findings suggest that Meta-IR consistently outperforms all combinations.

\begin{table}[!h]
\caption{Pairwise comparison between the performances of the Meta-IR \textit{versus} each pipeline (resampling strategy and regressor) over all datasets. The values represent the p-value of the Wilcoxon signed-rank test using the F1-scoreR metric. All results with $\alpha < 0.05$ are in bold.}
\label{tab:p-valueF1}
\centering
\scalefont{0.9}
\setlength{\tabcolsep}{4pt}
\begin{tabular}{cccccccc}
\hline
Resampl Strat & NONE              & GN                & RO                & RU                & SG                & SMT               & WERCS             \\ \hline
\multicolumn{8}{c}{\textbf{p-value}}         \\ \hline
BG  & \textbf{5.65e-24} & \textbf{7.06e-13} & \textbf{3.79e-13} & \textbf{5.08e-26} & \textbf{1.73e-18} & \textbf{4.98e-16} & \textbf{2.95e-11} \\
DT  & \textbf{2.04e-5}  & 0.323             & 0.177             & \textbf{9.65e-8}  & \textbf{0.009}             & \textbf{0.012}             & \textbf{0.049}    \\
MLP & \textbf{4.13e-17} & \textbf{3.18e-8}  & \textbf{1.86e-11} & \textbf{5.73e-15} & \textbf{3.48e-10} & \textbf{9.44e-13} & \textbf{1.06e-8}  \\
RF  & \textbf{8.07e-18} & \textbf{9.26e-10} & \textbf{5.94e-14} & \textbf{3.54e-16} & \textbf{8.22e-16} & \textbf{8.18e-15} & \textbf{4.86e-8}  \\
SVR & \textbf{1.53e-16} & \textbf{1.78e-6}  & \textbf{3.32e-13} & \textbf{1.20e-17} & \textbf{9.85e-11} & \textbf{1.20e-7}  & \textbf{3.35e-10} \\
XG  & \textbf{7.99e-18} & \textbf{4.46e-11} & \textbf{2.34e-16} & \textbf{4.02e-16} & \textbf{2.91e-15} & \textbf{9.66e-17} & \textbf{5.15e-13} \\ \hline
\end{tabular}
\end{table}

\begin{table}[!h]
\caption{Pairwise comparison between the performances of the Meta-IR \textit{versus} each pipeline (resampling strategy and regressor) over all datasets. The values represent the p-value of the Wilcoxon signed-rank test using the SERA metric. All results with $\alpha < 0.05$ are in bold.}
\label{tab:p-valueSERA}
\centering
\scalefont{0.9}
\setlength{\tabcolsep}{4pt}
\begin{tabular}{cccccccc}
\hline
Resampl Strat & NONE              & GN                & RO                & RU                & SG                & SMT               & WERCS             \\ \hline
\multicolumn{8}{c}{\textbf{p-value}}                                                                                                                      \\ \hline
BG            & \textbf{3.81e-15} & \textbf{5.77e-15} & \textbf{6.50e-16} & \textbf{4.45e-19} & \textbf{1.52e-18} & \textbf{1.11e-16} & \textbf{2.24e-14} \\
DT            & \textbf{9.81e-21} & \textbf{1.75e-15} & \textbf{3.24e-18} & \textbf{7.57e-23} & \textbf{1.02e-21} & \textbf{1.03e-20} & \textbf{4.32e-23} \\
MLP           & \textbf{1.68e-25} & \textbf{1.01e-28} & \textbf{2.16e-25} & \textbf{1.16e-27} & \textbf{3.28e-29} & \textbf{1.35e-25} & \textbf{1.96e-23} \\
RF            & \textbf{3.19e-7}  & \textbf{4.70e-7}  & \textbf{1.17e-3}  & \textbf{2.81e-12} & \textbf{9.16e-10} & \textbf{4.84e-7}  & \textbf{2.88e-6}  \\
SVR           & \textbf{5.60e-14} & \textbf{8.58e-13} & \textbf{2.67e-14} & \textbf{4.21e-18} & \textbf{6.38e-16} & \textbf{2.22e-16} & \textbf{6.14e-15} \\
XG            & \textbf{1.11e-29} & \textbf{1.42e-26} & \textbf{1.07e-29} & \textbf{1.30e-28} & \textbf{1.02e-28} & \textbf{7.87e-30} & \textbf{4.44e-29} \\ \hline
\end{tabular}
\end{table}

\subsubsection{Multiple comparisons with AutoML frameworks}
\label{sec:comp-automl}

\begin{figure}[!h]
\centering
\subfigure[F1-scoreR]{
\includegraphics[scale=0.52]{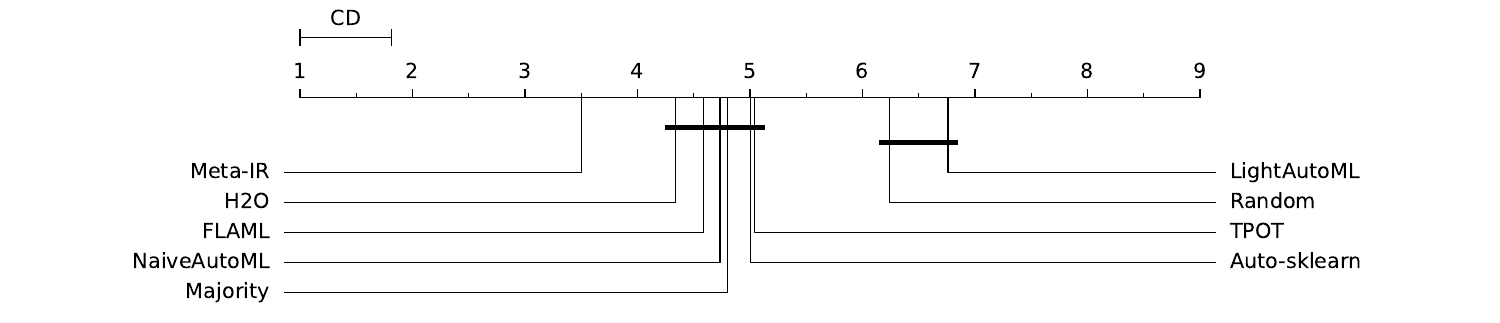}}
\subfigure[SERA]{
\includegraphics[scale=0.52]{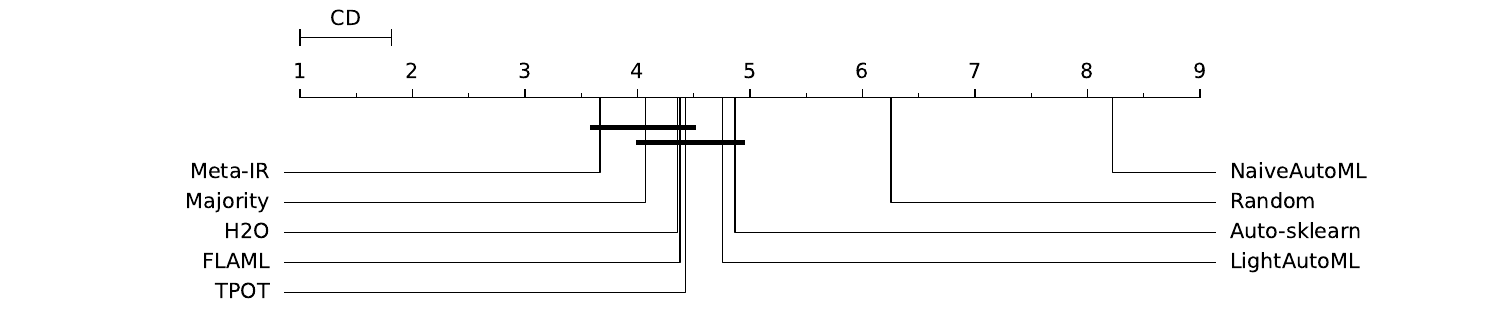}}

\caption{Critical Difference diagrams obtained from the recommendations given by the models.}
\label{CD}
    
\end{figure}

The comparison of Meta-IR and various AutoML frameworks such as Auto-sklearn, H2O, TPOT, FLAML, LightAutoML and NaiveAutoML, as well as the baselines Random and Majority based on two metrics: F1-scoreR and SERA, is presented in Figure~\ref{CD}(a-b). To verify which results are statistically different, the post-hoc Nemenyi test was applied. The presented Critical Difference diagrams~\citep{demvsar2006statistical} for each model consider both metrics. The horizontal line demonstrates the significance of the difference between the models. Models that are not connected present a significant difference (p-value $<$ 0.05) compared to the other methods.

For both the F1-ScoreR and SERA metrics, Meta-IR demonstrates the best results. However, for the SERA metric, there is no significant difference when compared to the Majority, H2O, FLAML and TPOT methods. Despite this, Meta-IR stands out when compared to other methods. Therefore, we can conclude that there are benefits to using an MtL-based model for recommending learning models and resampling strategies.


To better understand the behavior of the proposed model compared to the AutoML frameworks, we evaluated it across different percentages of rare cases. By utilizing percentiles, we segmented the data into three categories of rare cases: low, medium, and high. The low level consists of 45 datasets with rare case percentages below 8.2, while the medium level consists of 78 datasets with percentages above 8.3 and below 15.2. Lastly, the high level consists of 95 datasets with percentages exceeding 15.2. We performed a pairwise comparison of the performance of the Meta-IR model and AutoML frameworks for each category (Tables~\ref{tab:p.rarelevelf1} and \ref{tab:p.rarelevel}). At the high level of rare cases, Meta-IR outperformed all other AutoML frameworks for both metrics, with p-values below the significance threshold ($\alpha < 0.05$). 
However, for the SERA metric, at the medium and low rare case levels, there were no statistically significant differences between the performance of Meta-IR and the AutoML frameworks, as indicated by p-values above 0.05. For the F1-scoreR, although Meta-IR performance remains superior for datasets with medium and low percentages of rare cases, in some cases, the significance decreases. This suggests that as the proportion of rare cases decreases, the advantage of Meta-IR reduces.

These findings suggest that the Meta-IR's ability to weigh the importance of rare cases effectively makes it advantageous in situations where rare instances significantly impact the overall performance. Therefore, Meta-IR is an effective solution for tasks where addressing rare cases is essential. The strong performance, especially on datasets with a high percentage of rare cases, can be attributed to the fact that Meta-IR uses specific resampling strategies to balance data, focusing on metrics that are crucial for imbalanced regression tasks. Unlike traditional methods that use standard measures, Meta-IR selects models based on their ability to improve the F1-scoreR and SERA measures, which highlight how well the model handles rare cases. By focusing on these measures and using resampling strategies, Meta-IR is better at tackling the challenges of imbalanced regression. This leads to better performance on datasets with more rare instances.

\begin{table}[!h]
\caption{Pairwise comparison between the performances of the Meta-IR \textit{versus} AutoML frameworks at each level of percentage of rare cases (High, Medium and Low). The values represent the p-value of the Wilcoxon signed-rank test using the F1-scoreR metric. All results with $\alpha < 0.05$ are in bold.}
\label{tab:p.rarelevelf1}
\centering
\scalefont{0.9}
\begin{tabular}{lcccccc}
\hline
       & Auto-sklearn      & H2O               & TPOT              & FLAML             & LightAutoML        & NaiveAutoML    \\ \hline
High   & \textbf{4.462e-7} & \textbf{1.298e-4} & \textbf{6.373e-6} & \textbf{2.612e-6} & \textbf{4.196e-10} & \textbf{0.016} \\
Medium & 0.178             & \textbf{0.022}    & \textbf{5.865e-4} & \textbf{0.017}    & \textbf{7.975e-5}  & 0.289          \\
Low    & \textbf{0.033}    & \textbf{0.034}    & 0.112             & \textbf{0.045}    & \textbf{0.007}     & 0.484          \\ \hline
\end{tabular}
\end{table}

\begin{table}[!h]
\caption{Pairwise comparison between the performances of the Meta-IR \textit{versus} AutoML frameworks at each level of percentage of rare cases (High, Medium and Low). The values represent the p-value of the Wilcoxon signed-rank test using the SERA metric. All results with $\alpha < 0.05$ are in bold.}
\label{tab:p.rarelevel}
\centering
\scalefont{0.9}
\begin{tabular}{@{}lcccccc@{}}
\toprule
       & Auto-sklearn      & H2O               & TPOT              & FLAML          & LightAutoML       & NaiveAutoML        \\ \midrule
High   & \textbf{5.070e-6} & \textbf{1.829e-4} & \textbf{7.714e-5} & \textbf{0.001} & \textbf{1.502e-7} & \textbf{3.185e-16} \\
Medium & 0.387             & 0.558             & 0.638             & 0.167          & 0.775             & \textbf{3.879e-8}  \\
Low    & 0.376             & 0.474             & 0.947             & 0.991          & 0.308             & \textbf{6.655e-8}  \\ \bottomrule
\end{tabular}

\end{table}


Another comparison between Meta-IR and AutoML frameworks is made in terms of wins, ties, and losses. The results, which are presented in Table~\ref{tab:mean_win}, show that Meta-IR consistently outperformed the AutoML frameworks across various datasets, achieving a minimum victory percentage of 57.80\% and a maximum of 75.69\%. These percentages underscore Meta-IR's superiority over AutoML approaches for handling imbalanced regression, highlighting its effectiveness across diverse datasets and reinforcing its potential as a robust solution for imbalanced regression tasks.

\begin{table}[!h]
\scalefont{0.8}
\caption{Win/tie/loss of the Meta-IR versus AutoML frameworks.}
\label{tab:mean_win}
\begin{tabular}{@{}lcccccc@{}}
\toprule
                   & \textbf{Auto-sklearn} & \textbf{H2O} & \textbf{TPOT} & \textbf{FLAML} & \textbf{LightAutoML} & \textbf{NaiveAutoML} \\ \midrule
\textbf{F1-scoreR} & 143/0/75              & 143/0/75     & 147/2/69      & 145/0/73       & 165/4/49             & 132/4/82             \\
\textbf{SERA}      & 135/0/83              & 128/0/90     & 129/0/89      & 126/0/92       & 142/0/76             & 199/0/19             \\ \bottomrule
\end{tabular}
\end{table}

Those findings are promising and showcase the potential of Meta-IR as a competitive alternative in automated machine learning, specifically in imbalanced regression problems. A notable limitation of existing AutoML frameworks is their lack of specific metrics for imbalanced regression as an optimization function. Traditional regression metrics may not adequately capture the performance of models on rare or extreme values, which are critical in imbalanced regression tasks. Unlike Meta-IR, which explicitly incorporates resampling strategies and specialized evaluation measures, most AutoML frameworks neglect these aspects, leaving a significant gap in addressing imbalanced data effectively.

It is important to note that despite differences among AutoML frameworks—such as preprocessing steps, predefined search spaces, search heuristics, and performance metrics—these distinctions do not invalidate comparisons. Instead, they highlight the diversity of approaches and the potential for improvement. This comparison highlights the strengths of Meta-IR and emphasizes the importance of expanding current AutoML frameworks to include relevant measures and resampling strategies. These advancements ensure better alignment between model evaluation and the goals of imbalanced regression tasks. By explicitly addressing these limitations, Meta-IR fills this gap, offering a solution for imbalanced regression problems.

\subsubsection{Meta-IR as a preprocessing step for AutoML frameworks}
\label{sec:auto+meta}

This section explores the advantages of using the Meta-IR method in conjunction with AutoML frameworks for imbalanced regression tasks. While previous sections have demonstrated the effectiveness of Meta-IR, here we focus on its role as a preprocessing step to enhance the performance of AutoML frameworks. We used the recommendations from the resampling strategies provided by the Independent Meta-IR model and applied the recommendation as preprocessing before the AutoML models.

Table \ref{tab:Meta-IRstep} compares different AutoML frameworks based on their F1-scoreR and SERA mean performance before and after the Meta-IR preprocessing step. The column $\Delta$ shows the percentage improvement from using Meta-IR. Regarding the F1-scoreR, all AutoML frameworks improve with Meta-IR. LightAutoML shows the highest improvement of 16.50\%, followed by FLAML with 8.64\%. However, the SERA metric results vary. Auto-sklearn and TPOT have significantly increased SERA values (1008.41\% and 42.84\%, respectively), while H2O, NaiveAutoML, and LightAutoML have substantially decreased SERA values, indicating improved performance, with reductions of -82.87\%, -28.37\% and -10.00\%, respectively.


\begin{table}[!h]
\caption{Meta-IR as a preprocessing step and Comparison with Baseline}
\label{tab:Meta-IRstep}
\begin{tabular}{@{}ccccccc@{}}
\toprule
                      & Baseline & Meta-IR & $\Delta$ (\%)                  & Baseline  & Meta-IR   & $\Delta$ (\%)                       \\ \midrule
                      & \multicolumn{3}{c}{$\uparrow$ \textbf{F1-scoreR} }             & \multicolumn{3}{c}{$\downarrow$ \textbf{SERA}}                          \\ \midrule
\textbf{Auto-sklearn} & 0.246    & 0.258   & \cellcolor[HTML]{9AFF99}4.88   & 1.485e+10 & 1.645e+11 & \cellcolor[HTML]{FD6864}1008.41     \\
\textbf{H2O}          & 0.248    & 0.259   & \cellcolor[HTML]{9AFF99}4.44   & 2.219e+11 & 3.801e+10 & \cellcolor[HTML]{9AFF99}-82.87      \\
\textbf{TPOT}         & 0.244    & 0.258   & \cellcolor[HTML]{9AFF99}5.74   & 2.116e+10 & 3.022e+10 & \cellcolor[HTML]{FD6864}42.84       \\
\textbf{FLAML}        & 0.243    & 0.264   & \cellcolor[HTML]{9AFF99}8.64   & 1.875e+10 & 2.355e+10 & \cellcolor[HTML]{FD6864}25.60       \\
\textbf{LightAutoML}  & 0.206    & 0.240   & \cellcolor[HTML]{9AFF99}16.50  & 3.132e+10 & 2.819e+10 & \cellcolor[HTML]{9AFF99}-10.00      \\
\textbf{Naive AutoML} & 0.264    & 0.277   & \cellcolor[HTML]{9AFF99}4.92   &   6.388e+10
        &  4.575e+10   &   \cellcolor[HTML]{9AFF99}-28.37                                \\ \bottomrule
\end{tabular}
\end{table}

\begin{figure}[!h]
\centering
\subfigure[F1-scoreR]{
\includegraphics[scale=0.41]{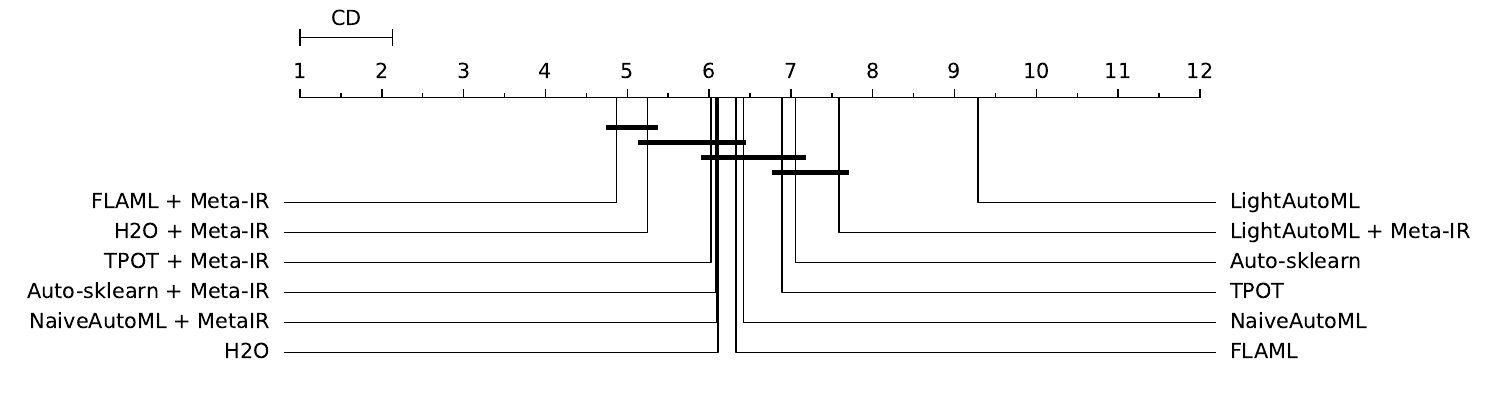}}
\subfigure[SERA]{
\includegraphics[scale=0.41]{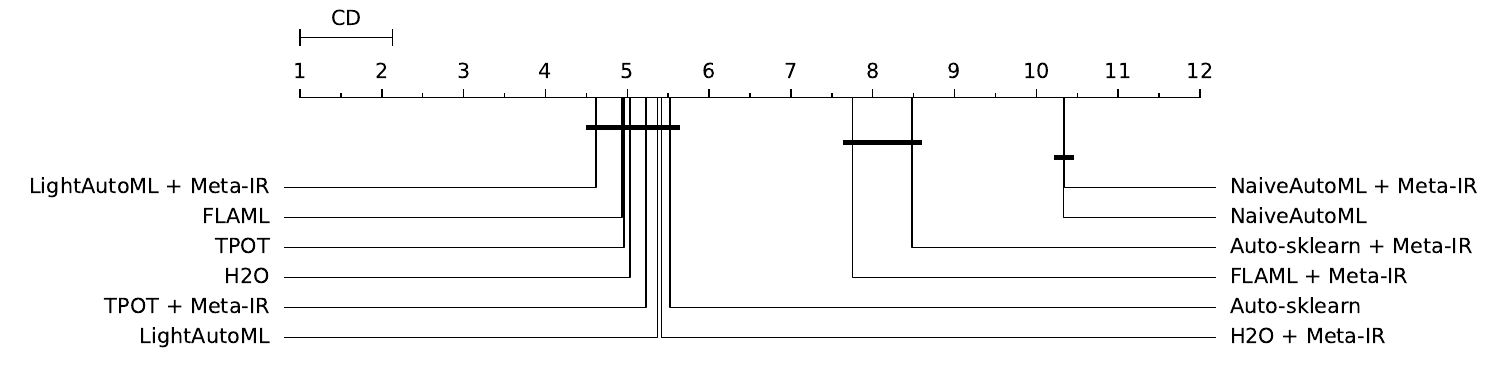}}

\caption{Critical Difference diagrams obtained from the recommendations given by the models.}
\label{fig:CD2}
    
\end{figure}

In Figure~\ref{fig:CD2}, critical difference diagrams illustrate the performance of the F1-scoreR and SERA metrics. Regarding F1-scoreR, the results reveal significant insights into the performance improvements the Meta-IR approach brings when combined with various AutoML methods. Notably, combinations like FLAML + Meta-IR, H2O + Meta-IR, and Auto-sklearn + Meta-IR consistently rank higher than their baseline. This suggests that the Meta-IR technique effectively enhances the predictive capabilities and overall F1-scoreR  performance of these AutoML frameworks. On the other hand, considering the SERA metric, the combination of LightAutoML with Meta-IR achieves the highest performance, significantly outperforming other methods, including its baseline.


In summary, the analysis of Meta-IR as a preprocessing step shows its potential to improve the performance of various AutoML frameworks significantly. There were noticeable enhancements in F1-scoreR for all tested methods, especially for LightAutoML and FLAML. These results suggest that Meta-IR effectively enhances the predictive capabilities of these frameworks. However, evaluating the SERA metric provides a different perspective, revealing substantial improvements with Meta-IR for LightAutoML, H2O and NaiveAutoML, while other methods yield mixed results. This indicates that the impact of Meta-IR on SERA varies by method, necessitating further exploration to understand the relationship. Overall, the positive improvements in F1-scoreR underscore the value of Meta-IR in enhancing model performance. Nonetheless, carefully considering its effects on different metrics is crucial to maximize its benefits in practical applications.

\subsection{Execution time analysis}
\label{sec:time-analysis}




This section evaluates the execution time of Meta-IR against AutoML frameworks. The Meta-IR process involves extracting meta-features and recommending a pipeline and for all AutoML frameworks was given a maximum budget of 1 hour. However, it is important to emphasize that AutoML models can produce output before the 1-hour mark. A comparison was made using ten datasets (see Table \ref{tab:dadosTempo}) to determine whether Meta-IR offers a time-efficient option compared to AutoML frameworks. This assessment aims to highlight the scalability and potential time-saving benefits of the proposed approach.

\begin{table}[!h]
\caption{Data sets used to calculate time.}
\label{tab:dadosTempo}
\centering
\scalefont{0.8}
\begin{tabular}{@{}lcccc@{}}
\toprule
\textbf{Dataset} & \textbf{n.samples} & \textbf{n.attributes} & \textbf{n.rare} & \textbf{p.rare} \\ \midrule
cpu-act          & 209                & 36                    & 33              & 15.8            \\
a1               & 198                & 11                    & 28              & 14.1            \\
auto93           & 93                 & 57                    & 11              & 11.8            \\
QSAR-TID-10929   & 154                & 1024                  & 18              & 11.7            \\
heat             & 7400               & 11                    & 664             & 9.0             \\
fri-c4-1000-10   & 1000               & 10                    & 89              & 8.9             \\
compactiv        & 8192               & 21                    & 713             & 8.7             \\
sleuth-case2002  & 147                & 6                     & 12              & 8.2             \\
space-ga         & 3107               & 6                     & 173             & 5.6             \\
wind             & 6574               & 14                    & 283             & 4.3             \\ \bottomrule
\end{tabular}

\end{table}

\begin{figure}[!h]
  \centering
  \subfigure[F1-scoreR]{\includegraphics[scale=0.5]{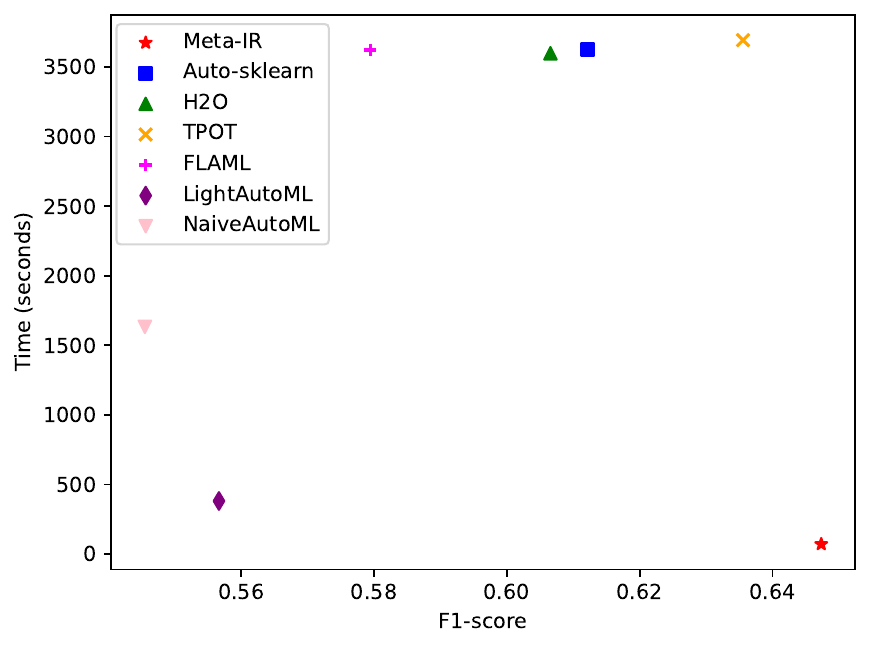}}

  \subfigure[SERA]{\includegraphics[scale=0.5]{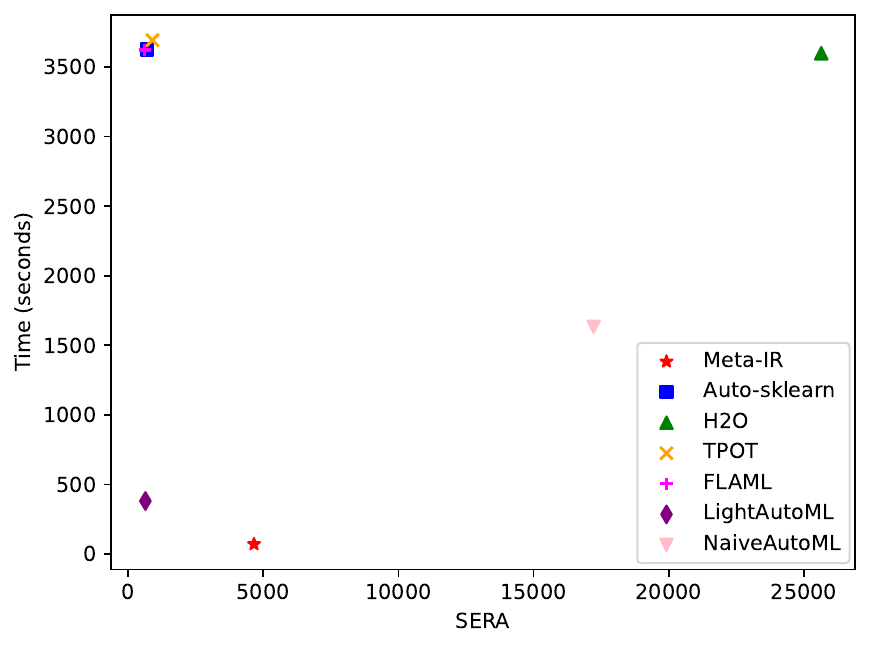}}
  \caption{Time and performance of each model}
  \label{fig:time}
\end{figure}

Figure~\ref{fig:time} illustrates the relationship between time and performance for each of the models considered, showing that Meta-IR outperforms all AutoML frameworks in terms of processing speed, highlighting its efficiency and scalability. While execution times can vary depending on factors like input data size and the number of parameters evaluated, computation time is not a limitation for our framework. The most time-consuming phase occurs offline, where configurations are trained and tested for each dataset to build the meta-dataset, followed by training the meta-model. During the Recommendation Phase, when a new dataset is presented, the framework extracts meta-features and uses the pre-trained meta-model to recommend the learning model and resampling strategy, avoiding the need to test and train all 42 possible configurations. This zero-shot recommendation process significantly reduces time, making Meta-IR approximately 50 times faster than traditional AutoML approaches.






\subsection{Meta-features analysis}
\label{sec:meta-featuresanalysis}


This analysis explored the importance of meta-features for the Meta-IR meta-model. We determined their importance using the \texttt{feature\_importance} attribute of the Random Forest model, which measures the relative influence of each meta-feature on the model's predictive ability. The feature importance is determined based on Gini impurity~\citep{breiman2017classification}, providing a measure of how much each meta-feature contributes to the overall performance of the model.

\begin{figure}[H]
    \centering
    {\includegraphics[scale=0.25]{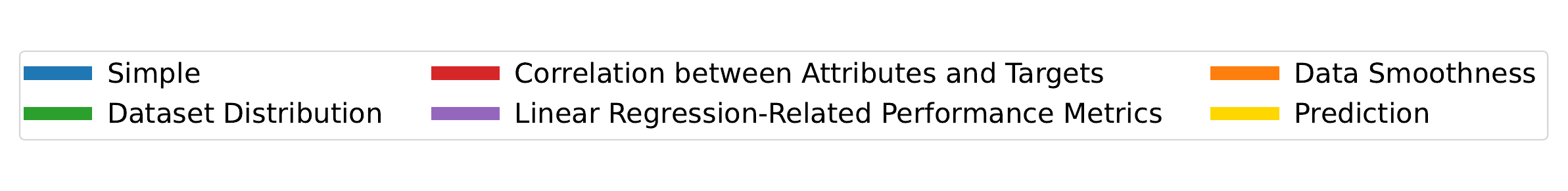}}
    \subfigure[Independent]{\includegraphics[scale=0.29, angle=-90]{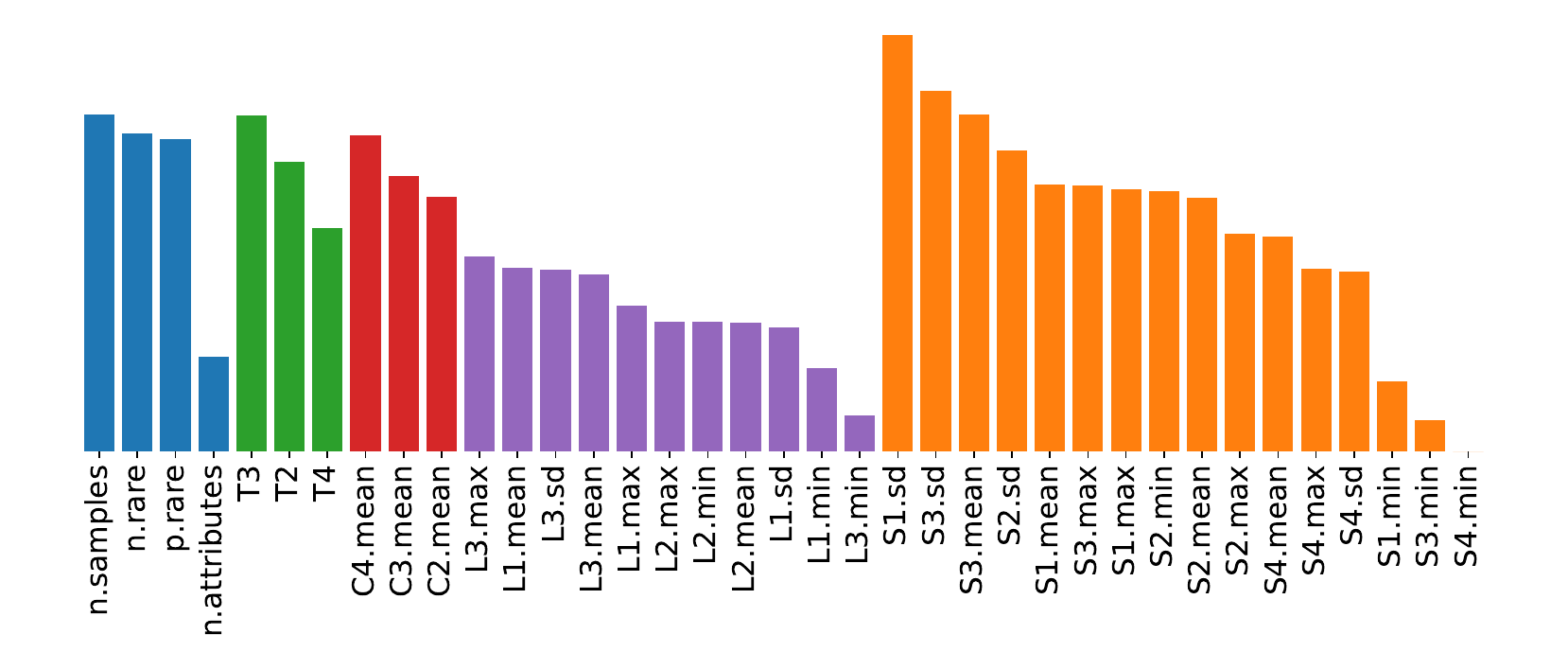}}
    \subfigure[ModelFirst]{\includegraphics[scale=0.29, angle=-90]{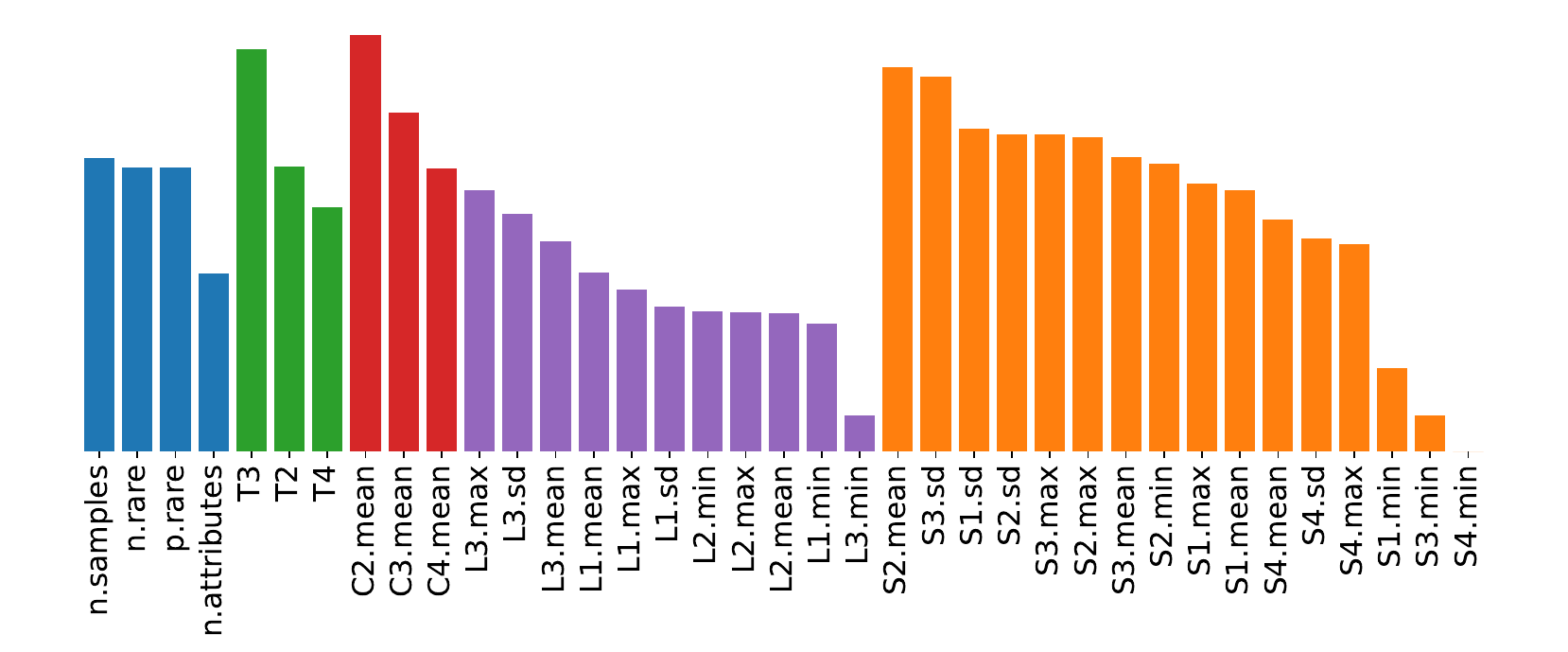}}
    \subfigure[StrategyFirst]{\includegraphics[scale=0.29, angle=-90]{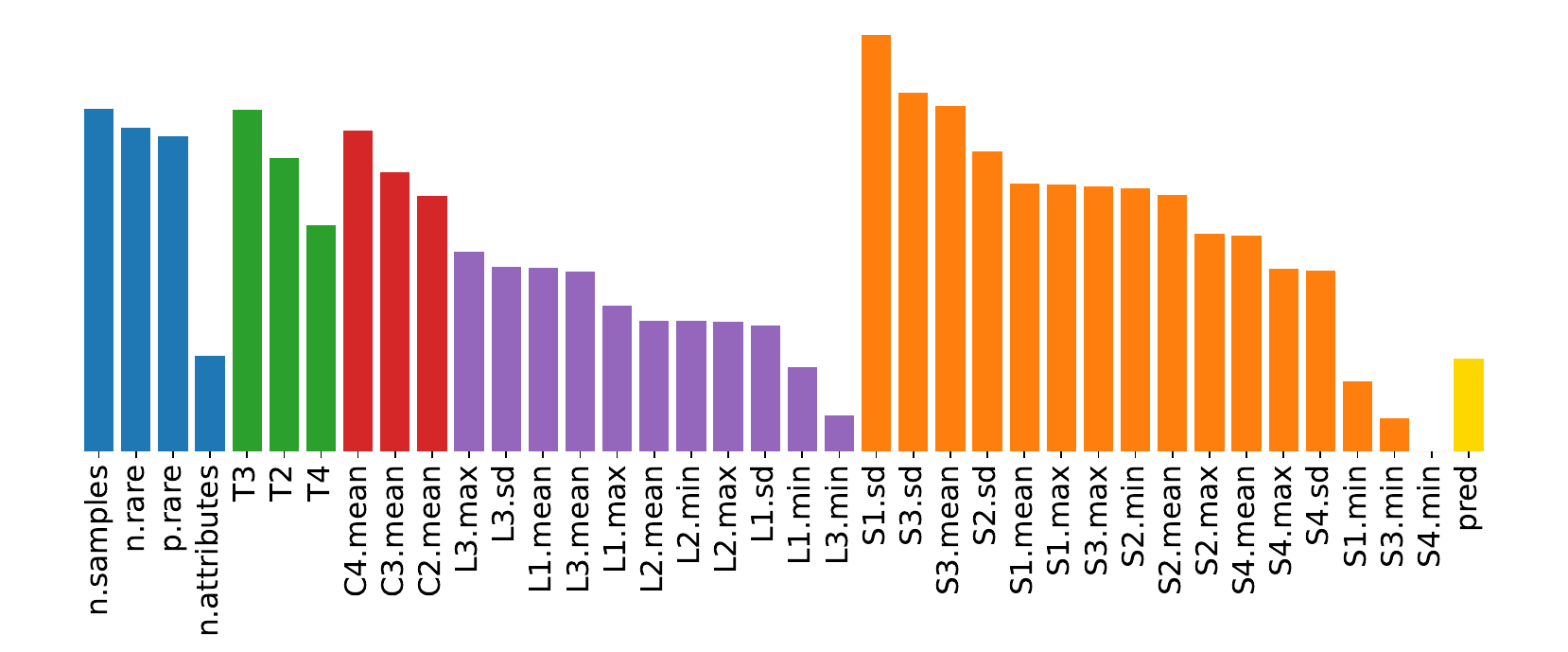}}

    \subfigure[Independent]{\includegraphics[scale=0.29, angle=-90]{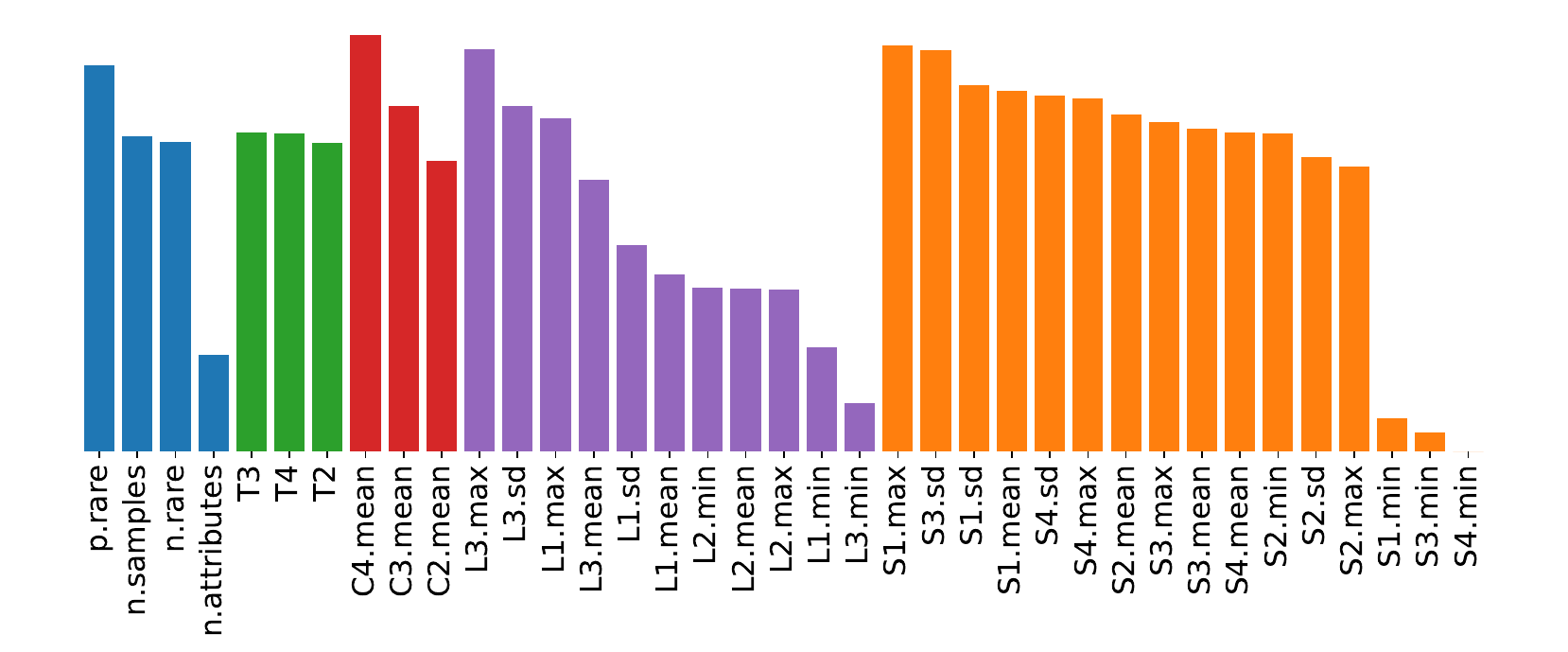}}
    \subfigure[StrategyFirst]{\includegraphics[scale=0.29, angle=-90]{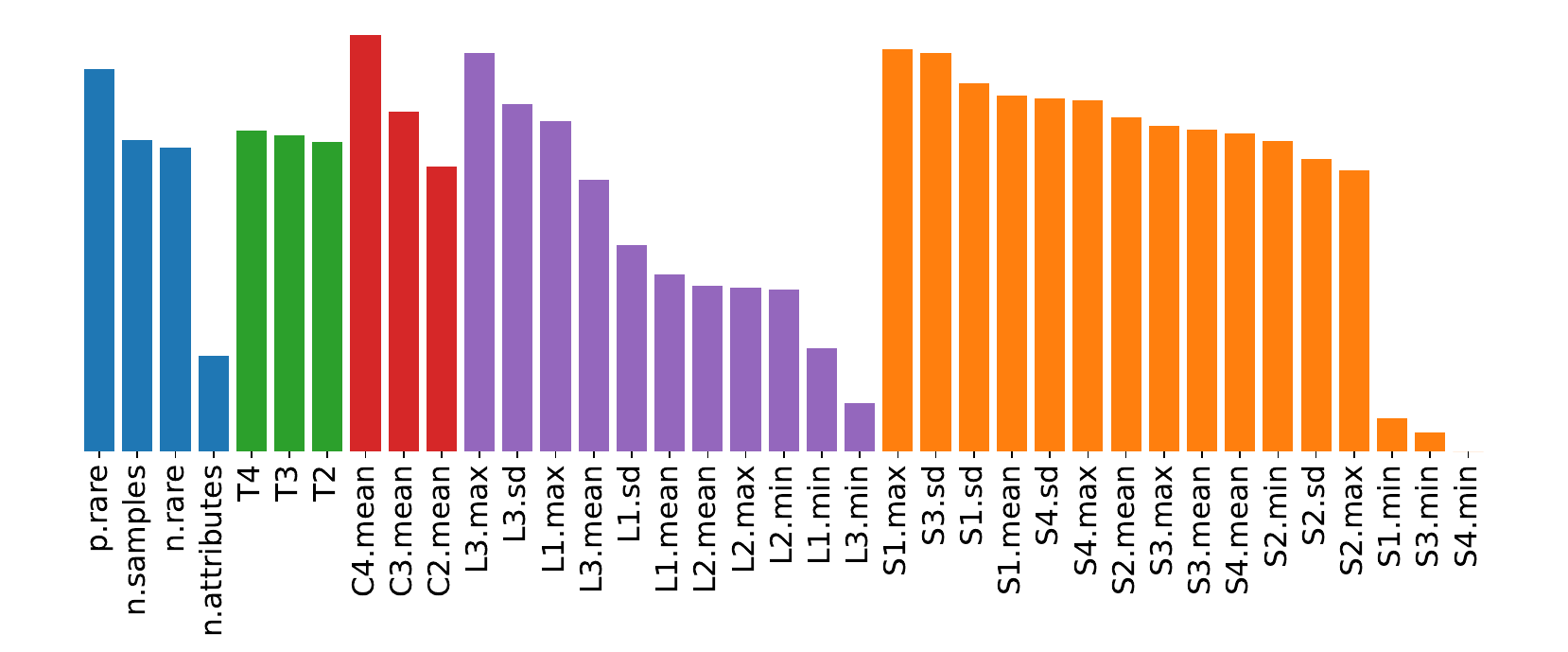}}
    \subfigure[ModelFirst]{\includegraphics[scale=0.29, angle=-90]{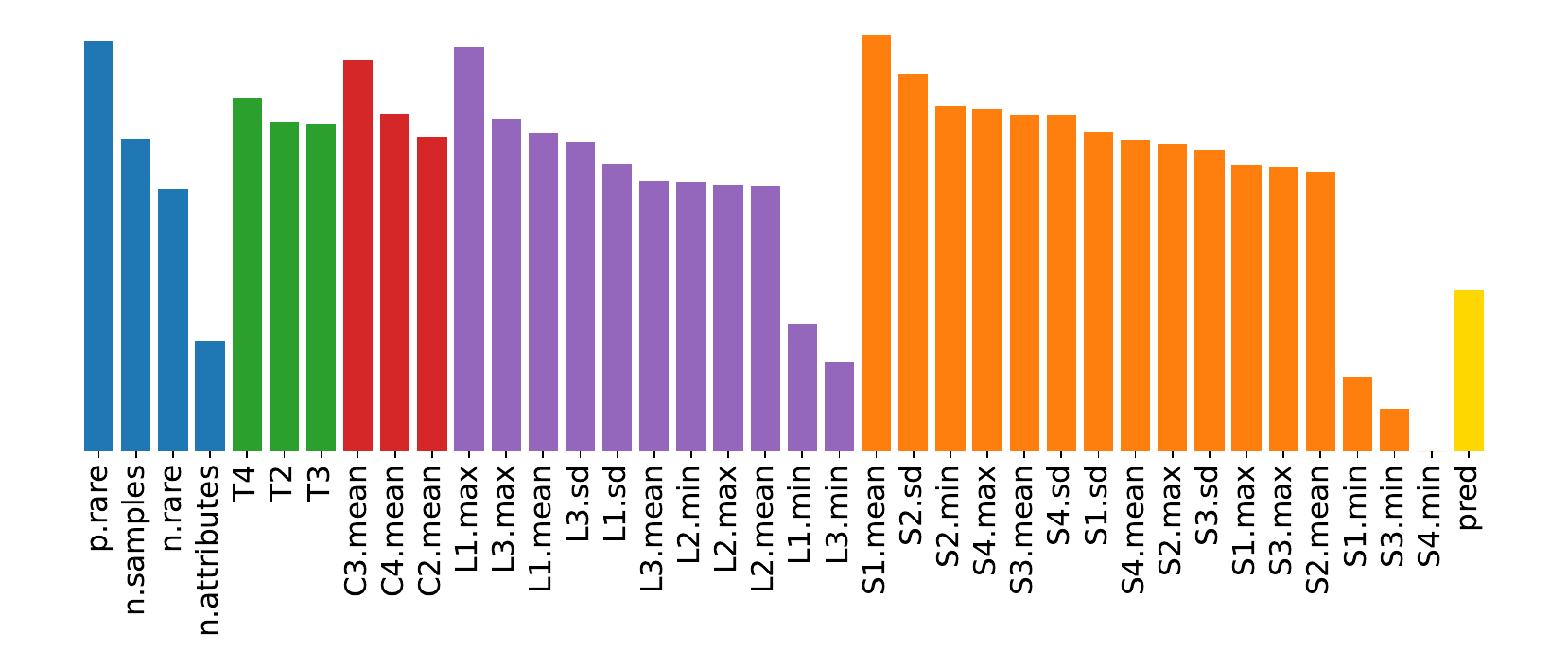}}

  \caption{Feature Importance for recommending the Learning model (a-c) and the Resampling Strategy (d-f), considering F1-scoreR metric as optimization function.}
  \label{fig:featureimportanceF1}
\end{figure}

\begin{figure}[H]
    \centering
    {\includegraphics[scale=0.25]{Figs/legend_only.pdf}}
    \subfigure[Independent]        {\includegraphics[scale=0.29, angle=-90]{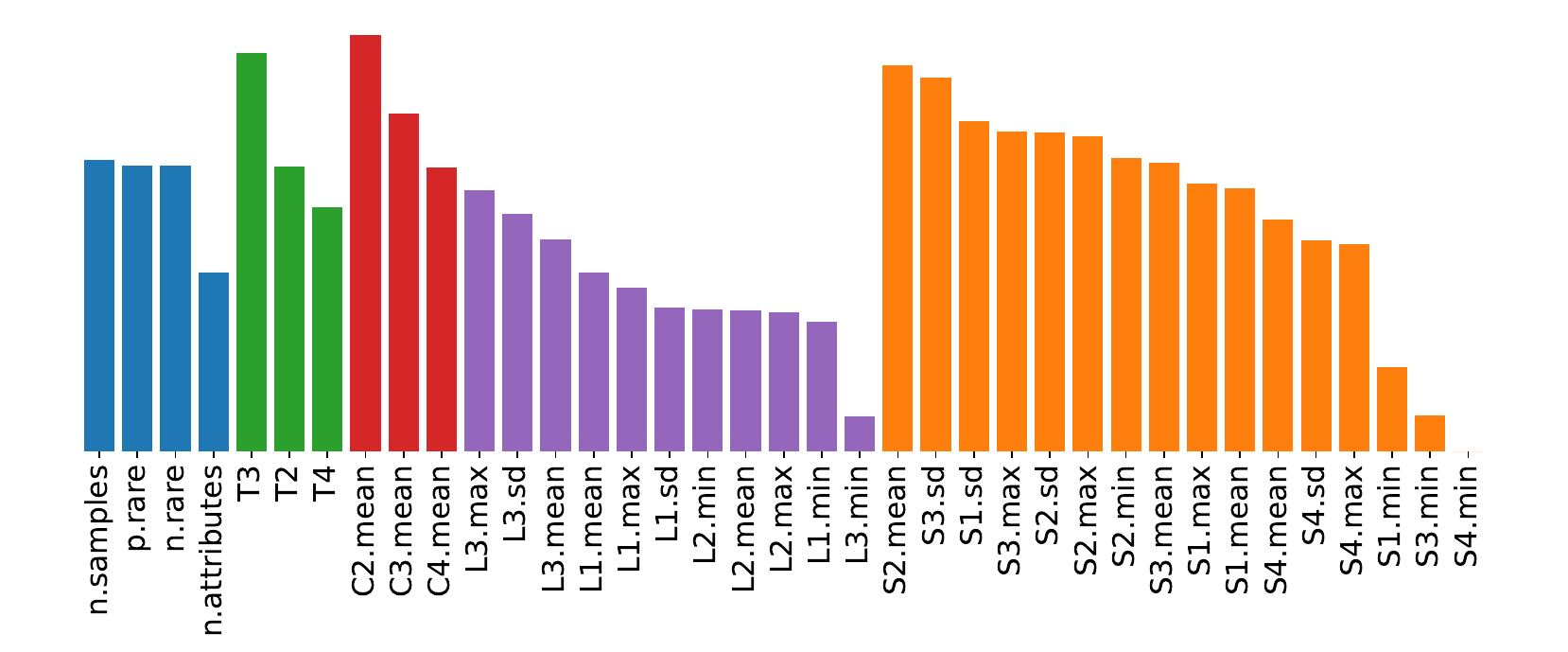}}
    \subfigure[ModelFirst]{\includegraphics[scale=0.29, angle=-90]{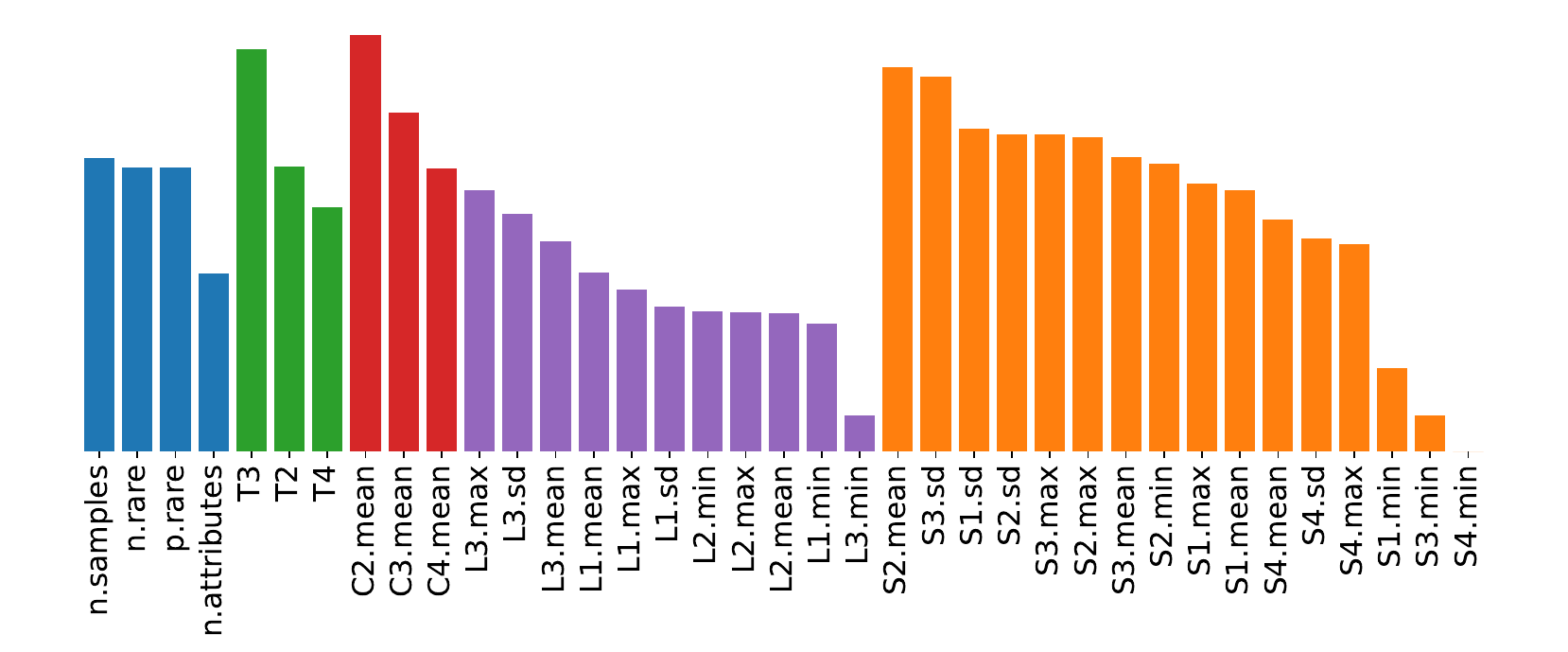}}
    \subfigure[StrategyFirst]{\includegraphics[scale=0.29, angle=-90]{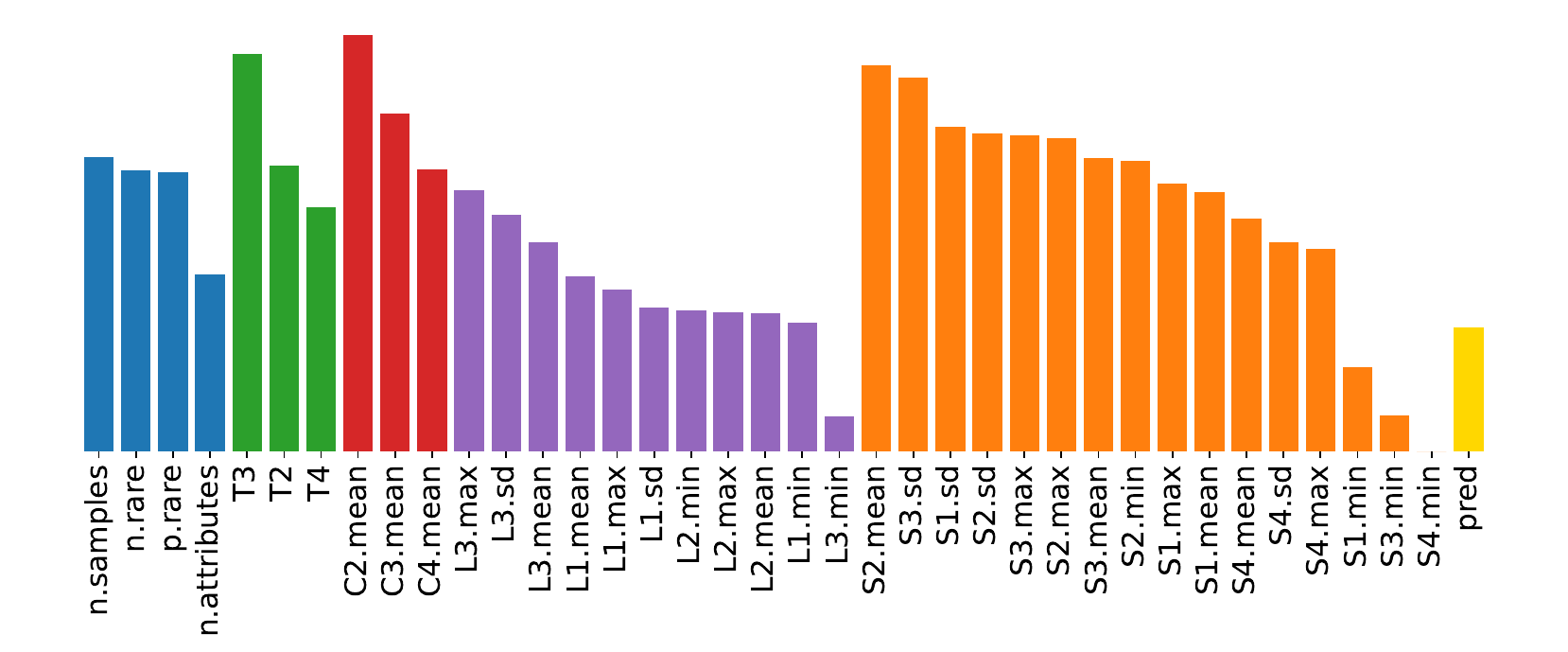}}

    \subfigure[Independent]{\includegraphics[scale=0.29, angle=-90]{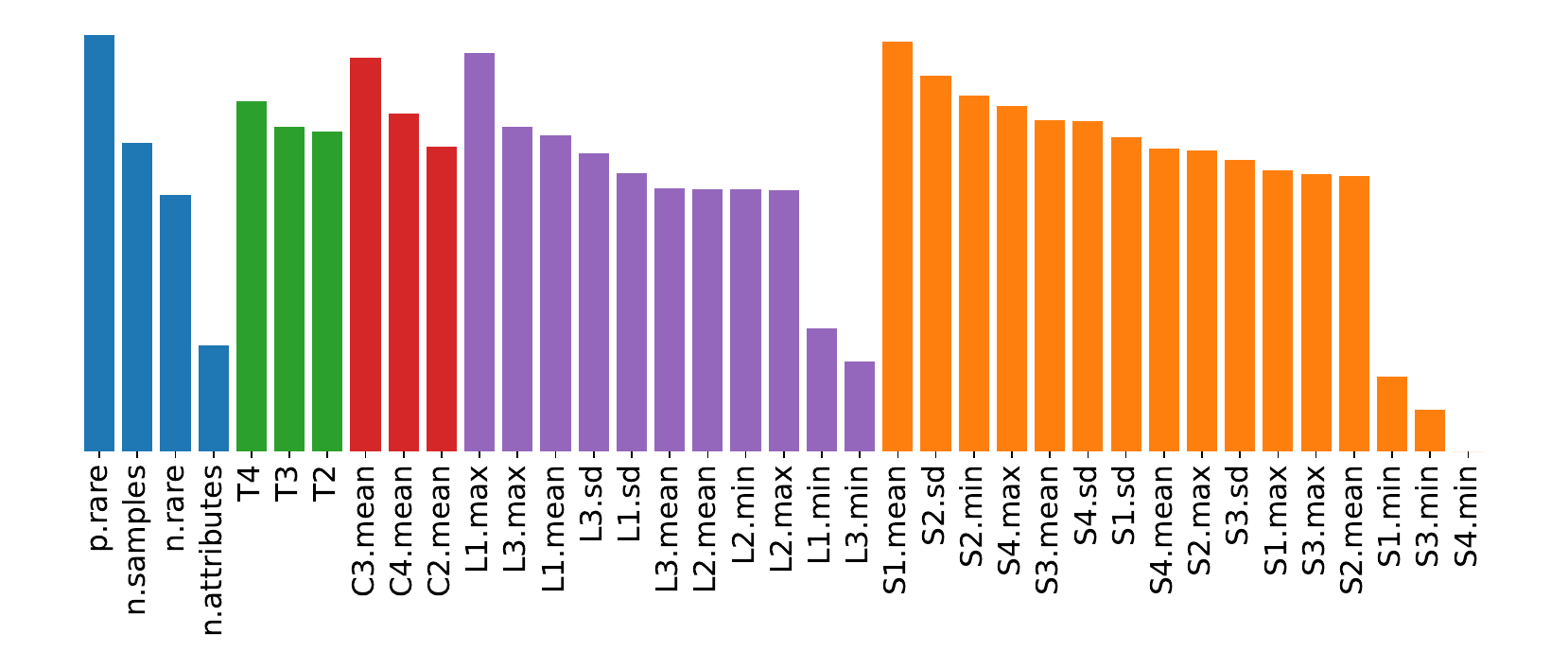}}
    \subfigure[StrategyFirst]{\includegraphics[scale=0.29, angle=-90]{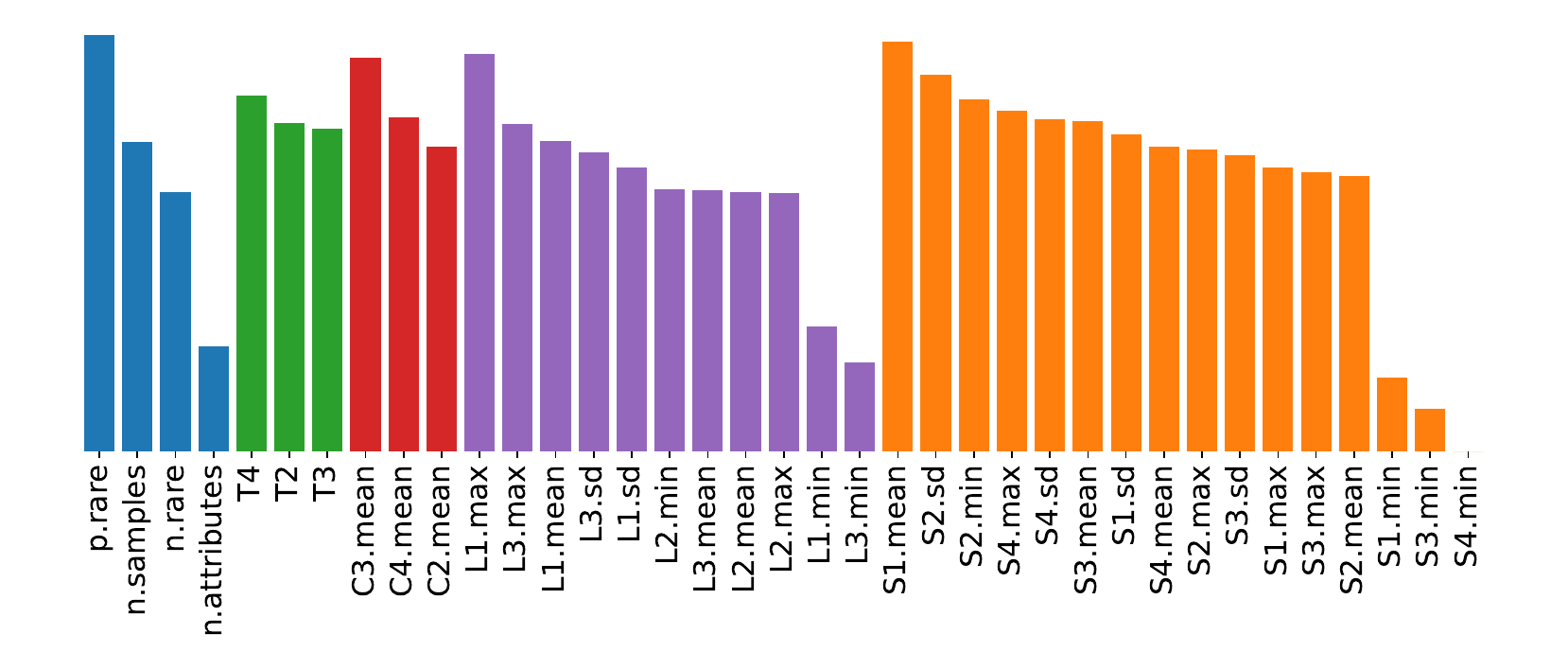}}
    \subfigure[ModelFirst]{\includegraphics[scale=0.29, angle=-90]{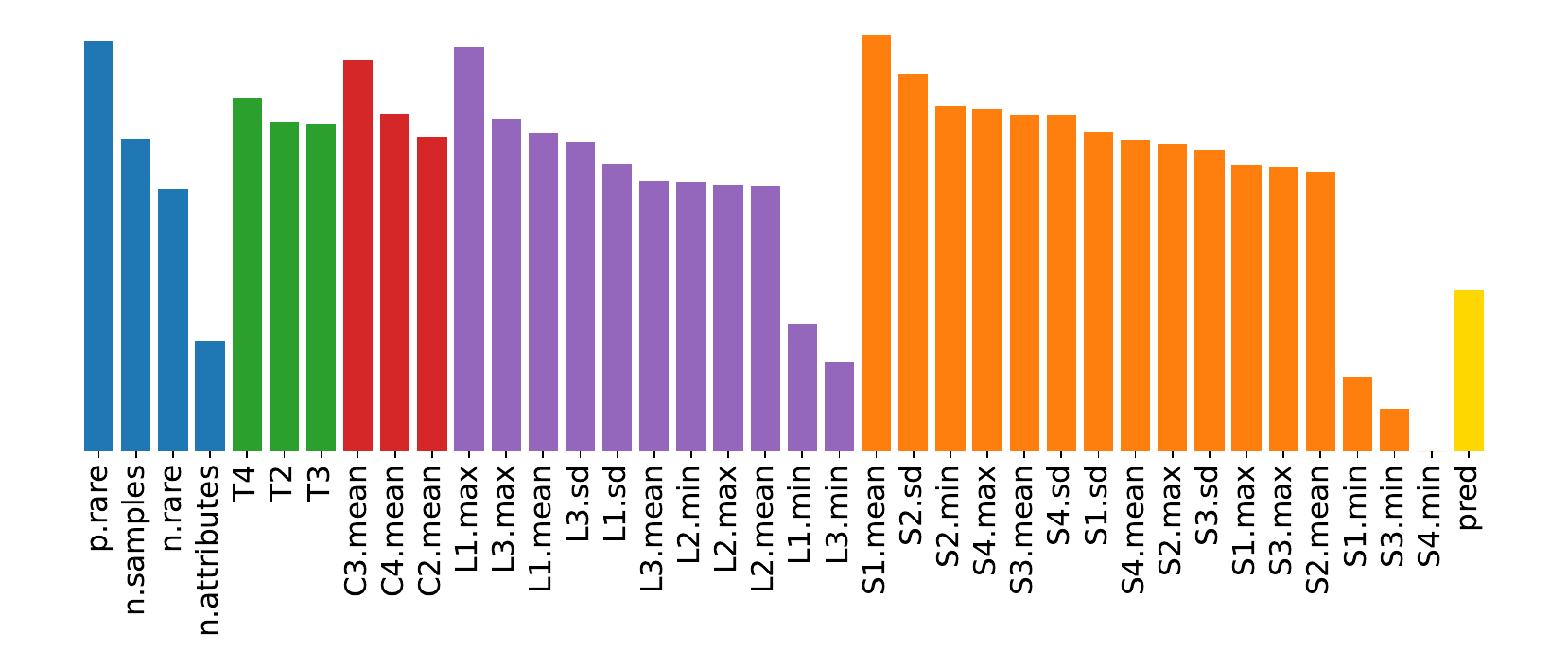}}

  \caption{Feature Importance for recommending the Learning model (a-c) and the Resampling Strategy (d-f), considering SERA metric as optimization function.}
  \label{fig:featureimportanceSERA}
\end{figure}

In Figures~\ref{fig:featureimportanceF1} and~\ref{fig:featureimportanceSERA}, we plot the feature importance (x-axis) of all meta-features (y-axis) for the meta-models trained for both metrics (F1-scoreR and SERA) as optimization function. The plot displays feature importance, organized into groups of meta-features, with each group represented by a distinct color. Within each group, the features are ordered to highlight the most important ones, providing a clear visualization of the relevance of each feature within its respective group.

In analyzing both metrics, it became evident that the percentage of rare cases (p.rare) is a crucial factor in determining the recommended resampling strategy, due to its direct correlation with data imbalance. Additionally, the measures C3.min, C4.mean, and S1.mean also play a significant role in this determination. When considering model recommendation, S1.sd emerges as a key factor for the F1-scoreR, while T3 and C2.mean are notable for the SERA metric. Generally, the minimum values of the measures (.min) have the least influence, indicating that, in the context of an imbalanced regression problem, these minimum values might not be significantly important in recommending learning models and resampling strategies. This suggests that focusing on other aspects or statistical properties, such as maximum values, average values, or variance, as well as simple statistics including the number of examples and the number of rare cases can be more beneficial in making informed decisions.

Upon analyzing the most important meta-features, we have observed that three (C2.mean, C3.min, and C4.mean) measure the relationship between features and the output. This indicates that the relationship between features and the target is crucial in representing imbalanced datasets. The other two meta-features (S1.mean and S1.sd) are related to the output distribution, a fundamental characteristic for addressing the imbalance. The target distribution may be directly linked to selecting the best resampling strategies and learning models. Understanding the target distribution is essential for choosing the pipeline.




It is essential to recognize a limitation related to the meta-features used in this study. Integrating additional high-quality features relevant to the problem could significantly enhance the model's effectiveness. This improvement would positively impact the outcomes of the meta-models and refine the base-level analysis. This recognition highlights the potential for future enhancements that could lead to more robust and accurate results, extending the applicability and effectiveness of the models proposed in this work.

\section{Conclusions}\label{Conclusions}

This work introduced Meta-IR, a meta-learning framework for recommending pipelines for imbalanced regression. Recommendations are made in two ways: Independent and Chained. We conducted an extensive experimental study consisting of 218 imbalanced regression datasets and taking into account 6 resampling techniques and 6 regression models. The meta-level analysis showed Meta-IR outperforms Majority and Random baselines in F1-scoreR and SERA metrics, highlighting the advantage of a meta-learning approach. The base-level assessment revealed that Meta-IR recommendations significantly surpassed Majority and Random. Furthermore, the results showed that Meta-IR consistently outperforms all AutoML frameworks when using the F1-scoreR metric, and when considering the SERA metric, Meta-IR has advantages, particularly in scenarios with a high percentage of rare cases. In terms of time-efficiency, Meta-IR is superior to AutoML frameworks.

The analysis also considered the impact of Meta-IR as a preprocessing step on the performance of various AutoML frameworks. It found that using Meta-IR significantly improved the F1-scoreR for all AutoML frameworks. However, its impact on the SERA metric varied depending on the AutoML method used. While it improved SERA for LightAutoML, H2O and NaiveAutoML, other methods showed deteriorated results. This indicates that the effect of Meta-IR on SERA depends on the AutoML method and needs more investigation.

In summary, the key findings of this work are as follows:

\begin{itemize}
    \item Meta-IR framework outperforms Majority and Random baselines on F1-scoreR and SERA metrics, demonstrating the advantage of using a meta-learning approach for imbalanced regression.
    \item Meta-IR recommendations significantly surpass Majority and Random baselines in base-level evaluations.
    \item Outperforms AutoML frameworks in terms of F1-scoreR and provides advantages in scenarios with a high percentage of rare cases when using the SERA metric.
    \item The Meta-IR approach showcased a significant advantage in time efficiency, being approximately 50 times faster than existing AutoML frameworks.
    \item The Meta-IR approach, when used as a preprocessing step, consistently enhances the F1-scoreR across all evaluated AutoML frameworks. However, its impact on the SERA metric is method-dependent: while improvements were observed with H2O, LightAutoML, and NaiveAutoML, further investigation is needed to fully understand the variability and optimize its effectiveness for this metric.
\end{itemize}

One limitation of this work is the lack of hyperparameter optimization. We chose not to perform hyperparameter optimization in this study in order to manage computational complexity and to emphasize the primary contribution of the research. We have shown that the recommended models perform consistently well across various problems without requiring fine-tuning, thus underscoring the generality and applicability of our approach. While including hyperparameter optimization in future work could potentially further improve results, its absence does not compromise the validity of the main conclusions of this study. Our future work will also enhance the chained algorithm recommendations by assessing multilabel strategies and investigating another meta-feature set. Additionally, we plan to model our approach to provide a ranked list of Top-N pipelines, enabling users to select options that are best aligned with their computational and performance constraints.

\bmhead{Acknowledgments}

This work was partially supported by Brazilian agencies: {\it Fundação de Amparo à Ciência e Tecnologia do Estado de Pernambuco} (FACEPE) and {\it Conselho Nacional de Desenvolvimento Científico e Tecnológico} (CNPq).

\section*{Statements and Declarations}

\subsection*{Competing Interests and Funding}

The authors have no competing interests to declare that are relevant to the content of this article.

\bibliography{sn-bibliography}

\begin{thebibliography}{85}
\providecommand{\natexlab}[1]{#1}
\providecommand{\url}[1]{{#1}}
\providecommand{\urlprefix}{URL }
\providecommand{\doi}[1]{\url{https://doi.org/#1}}
\providecommand{\eprint}[2][]{\url{#2}}
 \bibcommenthead

\bibitem[{Aguiar et~al(2019)Aguiar, Mantovani, Mastelini, de~Carvalho, Campos,
  and Junior}]{aguiar2019meta}
Aguiar GJ, Mantovani RG, Mastelini SM, et~al (2019) A meta-learning approach
  for selecting image segmentation algorithm. \textit{Pattern Recognition
  Letters} 128:480--487

\bibitem[{Aguiar et~al(2022)Aguiar, Santana, de~Carvalho, and
  Junior}]{aguiar2022using}
Aguiar GJ, Santana EJ, de~Carvalho AC, et~al (2022) Using meta-learning for
  multi-target regression. \textit{Information Sciences} 584:665--684

\bibitem[{Aminian et~al(2021)Aminian, Ribeiro, and Gama}]{aminian2021chebyshev}
Aminian E, Ribeiro RP, Gama J (2021) Chebyshev approaches for imbalanced data
  streams regression models. \textit{Data Mining and Knowledge Discovery}
  35:2389--2466

\bibitem[{de~Amorim et~al(2024)de~Amorim, Cavalcanti, and Cruz}]{de2024meta}
de~Amorim LB, Cavalcanti GD, Cruz RM (2024) Meta-scaler: A meta-learning
  framework for the selection of scaling techniques. \textit{IEEE Transactions
  on Neural Networks and Learning Systems} 36:4805--4819

\bibitem[{Avelino et~al(2024)Avelino, Cavalcanti, and
  Cruz}]{avelino2024resampling}
Avelino JG, Cavalcanti GD, Cruz RM (2024) Resampling strategies for imbalanced
  regression: a survey and empirical analysis. \textit{Artificial Intelligence
  Review} 57(4):82

\bibitem[{B{\"a}ck et~al(1997)B{\"a}ck, Fogel, and
  Michalewicz}]{back1997handbook}
B{\"a}ck T, Fogel DB, Michalewicz Z (1997) Handbook of evolutionary
  computation. \textit{Release} 97(1):B1

\bibitem[{Barella et~al(2018)Barella, Garcia, de~Souto, Lorena, and
  de~Carvalho}]{barella2018data}
Barella VH, Garcia LP, de~Souto MP, et~al (2018) Data complexity measures for
  imbalanced classification tasks. In: \textit{2018 International Joint
  Conference on Neural Networks}, IEEE, pp 1--8

\bibitem[{Barella et~al(2020)Barella, Garcia, and
  de~Carvalho}]{barella2020simulating}
Barella VH, Garcia LP, de~Carvalho AC (2020) Simulating complexity measures on
  imbalanced datasets. In: \textit{Brazilian Conference on Intelligent
  Systems}, Springer, pp 498--512

\bibitem[{Barella et~al(2021)Barella, Garcia, de~Souto, Lorena, and
  de~Carvalho}]{barella2021assessing}
Barella VH, Garcia LP, de~Souto MC, et~al (2021) Assessing the data complexity
  of imbalanced datasets. \textit{Information Sciences} 553:83--109

\bibitem[{Batista et~al(2004)Batista, Prati, and Monard}]{batista2004study}
Batista GE, Prati RC, Monard MC (2004) A study of the behavior of several
  methods for balancing machine learning training data. \textit{ACM SIGKDD
  Explorations Newsletter} 6(1):20--29

\bibitem[{Bengio(2000)}]{bengio2000gradient}
Bengio Y (2000) Gradient-based optimization of hyperparameters. \textit{Neural
  Computation} 12(8):1889--1900

\bibitem[{Bergstra and Bengio(2012)}]{bergstra2012random}
Bergstra J, Bengio Y (2012) Random search for hyper-parameter optimization.
  \textit{Journal of Machine Learning Research} 13(1):281--305

\bibitem[{Bergstra et~al(2011)Bergstra, Bardenet, K{\'e}gl, and
  Bengio}]{bergstra2011implementations}
Bergstra J, Bardenet R, K{\'e}gl B, et~al (2011) Implementations of algorithms
  for hyper-parameter optimization. In: \textit{NIPS Workshop on Bayesian
  Optimization}

\bibitem[{Branco et~al(2016)Branco, Ribeiro, and Torgo}]{branco2016ubl}
Branco P, Ribeiro RP, Torgo L (2016) Ubl: an r package for utility-based
  learning. arXiv preprint ArXiv:1604.08079

\bibitem[{Branco et~al(2017)Branco, Torgo, and Ribeiro}]{branco2017smogn}
Branco P, Torgo L, Ribeiro RP (2017) Smogn: a pre-processing approach for
  imbalanced regression. In: \textit{First International Workshop on Learning
  with Imbalanced Domains: Theory and Applications}, pp 36--50

\bibitem[{Branco et~al(2019)Branco, Torgo, and Ribeiro}]{branco2019pre}
Branco P, Torgo L, Ribeiro RP (2019) Pre-processing approaches for imbalanced
  distributions in regression. \textit{Neurocomputing} 343:76--99

\bibitem[{Brazdil et~al(2008)Brazdil, Carrier, Soares, and
  Vilalta}]{brazdil2008metalearning}
Brazdil P, Carrier CG, Soares C, et~al (2008) \textit{Metalearning:
  Applications to data mining}. Springer Science \& Business Media, Heidelberg

\bibitem[{Brazdil et~al(2022)Brazdil, van Rijn, Soares, and
  Vanschoren}]{brazdil2022metalearning}
Brazdil P, van Rijn JN, Soares C, et~al (2022) \textit{Metalearning:
  Applications to automated machine learning and data mining}. Springer Nature,
  Cham

\bibitem[{Brazdil and Soares(2000)}]{brazdil2000comparison}
Brazdil PB, Soares C (2000) A comparison of ranking methods for classification
  algorithm selection. In: \textit{European Conference on Machine Learning}.
  Springer, pp 63--75

\bibitem[{Brazdil et~al(2003)Brazdil, Soares, and
  Da~Costa}]{brazdil2003ranking}
Brazdil PB, Soares C, Da~Costa JP (2003) Ranking learning algorithms: Using ibl
  and meta-learning on accuracy and time results. \textit{Machine Learning}
  50:251--277

\bibitem[{Breiman(1996)}]{breiman1996bagging}
Breiman L (1996) Bagging predictors. \textit{Machine Learning} 24(2):123--140

\bibitem[{Breiman(2001)}]{breiman2001random}
Breiman L (2001) Random forests. \textit{Machine Learning} 45(1):5--32

\bibitem[{Breiman(2017)}]{breiman2017classification}
Breiman L (2017) \textit{Classification and regression trees}. Routledge

\bibitem[{Camacho et~al(2022)Camacho, Douzas, and Bacao}]{camacho2022geometric}
Camacho L, Douzas G, Bacao F (2022) Geometric smote for regression.
  \textit{Expert Systems with Applications} 193:116387

\bibitem[{Cavalcanti et~al(2012)Cavalcanti, Ren, and Vale}]{cavalcanti2012data}
Cavalcanti GD, Ren TI, Vale BA (2012) Data complexity measures and nearest
  neighbor classifiers: a practical analysis for meta-learning. In:
  \textit{2012 IEEE 24th International Conference on Tools with Artificial
  Intelligence}, IEEE, pp 1065--1069

\bibitem[{Chawla et~al(2002)Chawla, Bowyer, Hall, and
  Kegelmeyer}]{chawla2002smote}
Chawla NV, Bowyer KW, Hall LO, et~al (2002) Smote: synthetic minority
  over-sampling technique. \textit{Journal of Artificial Intelligence Research}
  16:321--357

\bibitem[{Chen and Guestrin(2016)}]{chen2016xgboost}
Chen T, Guestrin C (2016) Xgboost: A scalable tree boosting system. In:
  \textit{Proceedings of the 22nd ACM SIGKDD International Conference on
  Knowledge Discovery and Data Mining}. Association for Computing Machinery, pp
  785--794

\bibitem[{{de Amorim} et~al(2023){de Amorim}, Cavalcanti, and
  Cruz}]{amorim2023}
{de Amorim} LB, Cavalcanti GD, Cruz RM (2023) The choice of scaling technique
  matters for classification performance. \textit{Applied Soft Computing}
  133:109924

\bibitem[{Dem{\v{s}}ar(2006)}]{demvsar2006statistical}
Dem{\v{s}}ar J (2006) Statistical comparisons of classifiers over multiple data
  sets. \textit{Journal of Machine Learning Research} 7(1):1--30

\bibitem[{Dougherty et~al(1989)Dougherty, Edelman, and
  Hyman}]{dougherty1989nonnegativity}
Dougherty RL, Edelman AS, Hyman JM (1989) Nonnegativity-, monotonicity-, or
  convexity-preserving cubic and quintic hermite interpolation.
  \textit{Mathematics of Computation} 52(186):471--494

\bibitem[{Escalante et~al(2009)Escalante, Montes, and
  Sucar}]{escalante2009particle}
Escalante HJ, Montes M, Sucar LE (2009) Particle swarm model selection.
  \textit{Journal of Machine Learning Research} 10(15):405--440

\bibitem[{Feurer et~al(2015)Feurer, Klein, Eggensperger, Springenberg, Blum,
  and Hutter}]{feurer2015efficient}
Feurer M, Klein A, Eggensperger K, et~al (2015) Efficient and robust automated
  machine learning. In: \textit{Advances in Neural Information Processing
  Systems}, vol~28. Curran Associates, Inc.

\bibitem[{Ganganwar(2012)}]{ganganwar2012overview}
Ganganwar V (2012) An overview of classification algorithms for imbalanced
  datasets. \textit{International Journal of Emerging Technology and Advanced
  Engineering} 2(4):42--47

\bibitem[{Garcia et~al(2016)Garcia, de~Carvalho, and Lorena}]{garcia2016noise}
Garcia LP, de~Carvalho AC, Lorena AC (2016) Noise detection in the
  meta-learning level. \textit{Neurocomputing} 176:14--25

\bibitem[{Garcia et~al(2018)Garcia, Lorena, de~Souto, and
  Ho}]{garcia2018classifier}
Garcia LP, Lorena AC, de~Souto MC, et~al (2018) Classifier recommendation using
  data complexity measures. In: \textit{2018 24th International Conference on
  Pattern Recognition}, IEEE, pp 874--879

\bibitem[{Garc{\'\i}a et~al(2018)Garc{\'\i}a, Zhang, Altalhi, Alshomrani, and
  Herrera}]{garcia2018dynamic}
Garc{\'\i}a S, Zhang ZL, Altalhi A, et~al (2018) Dynamic ensemble selection for
  multi-class imbalanced datasets. \textit{Information Sciences} 445:22--37

\bibitem[{Garnett(2023)}]{garnett2023bayesian}
Garnett R (2023) \textit{Bayesian Optimization}. Cambridge University Press,
  Cambridge

\bibitem[{Garouani et~al(2023)Garouani, Ahmad, Bouneffa, and
  Hamlich}]{garouani2023autoencoder}
Garouani M, Ahmad A, Bouneffa M, et~al (2023) Autoencoder-knn meta-model based
  data characterization approach for an automated selection of ai algorithms.
  \textit{Journal of Big Data} 10(1):14

\bibitem[{Geurts et~al(2006)Geurts, Ernst, and Wehenkel}]{geurts2006extremely}
Geurts P, Ernst D, Wehenkel L (2006) Extremely randomized trees.
  \textit{Machine Learning} 63:3--42

\bibitem[{Ghaderi~Zefrehi et~al(2023)Ghaderi~Zefrehi, Sheikhi, and
  Alt{\i}n{\c{c}}ay}]{ghaderi2023threshold}
Ghaderi~Zefrehi H, Sheikhi G, Alt{\i}n{\c{c}}ay H (2023) Threshold prediction
  for detecting rare positive samples using a meta-learner. \textit{Pattern
  Analysis and Applications} 26(1):289--306

\bibitem[{Giraud-Carrier et~al(2004)Giraud-Carrier, Vilalta, and
  Brazdil}]{giraud2004introduction}
Giraud-Carrier C, Vilalta R, Brazdil P (2004) Introduction to the special issue
  on meta-learning. \textit{Machine Learning} 54(3):187--193

\bibitem[{Haixiang et~al(2017)Haixiang, Yijing, Shang, Mingyun, Yuanyue, and
  Bing}]{haixiang2017learning}
Haixiang G, Yijing L, Shang J, et~al (2017) Learning from class-imbalanced
  data: Review of methods and applications. \textit{Expert Systems with
  Applications} 73:220--239

\bibitem[{Hastie et~al(2009)Hastie, Tibshirani, Friedman, and
  Friedman}]{hastie2009elements}
Hastie T, Tibshirani R, Friedman JH, et~al (2009) \textit{The elements of
  statistical learning: data mining, inference, and prediction}, vol~2.
  Springer

\bibitem[{He et~al(2021)He, Zhao, and Chu}]{he2021automl}
He X, Zhao K, Chu X (2021) Automl: A survey of the state-of-the-art.
  \textit{Knowledge-Based Systems} 212:106622

\bibitem[{Hutter et~al(2011)Hutter, Hoos, and
  Leyton-Brown}]{hutter2011sequential}
Hutter F, Hoos HH, Leyton-Brown K (2011) Sequential model-based optimization
  for general algorithm configuration. In: \textit{Learning and Intelligent
  Optimization}. Springer Berlin Heidelberg, pp 507--523

\bibitem[{Khan et~al(2020)Khan, Zhang, Rehman, and Ali}]{khan2020literature}
Khan I, Zhang X, Rehman M, et~al (2020) A literature survey and empirical study
  of meta-learning for classifier selection. \textit{IEEE Access}
  8:10262--10281

\bibitem[{Ko et~al(2008)Ko, Sabourin, and Britto~Jr}]{ko2008dynamic}
Ko AH, Sabourin R, Britto~Jr AS (2008) From dynamic classifier selection to
  dynamic ensemble selection. \textit{Pattern Recognition} 41(5):1718--1731

\bibitem[{Kubat et~al(1997)Kubat, Matwin et~al}]{kubat1997addressing}
Kubat M, Matwin S, et~al (1997) Addressing the curse of imbalanced training
  sets: one-sided selection. In: \textit{Proceedings of the 14th International
  Conference on Machine Learning}, vol~97. Morgan Kaufmann Publishers Inc., pp
  179--186

\bibitem[{LeDell and Poirier(2020)}]{H2OAutoML20}
LeDell E, Poirier S (2020) {H2O} {A}uto{ML}: Scalable automatic machine
  learning. In: \textit{Proceedings of the 7th ICML Workshop on Automated
  Machine Learning (AutoML)}, p~24

\bibitem[{Lee(1999)}]{lee1999regularization}
Lee SS (1999) Regularization in skewed binary classification.
  \textit{Computational Statistics} 14(2):277--292

\bibitem[{Lee(2000)}]{lee2000noisy}
Lee SS (2000) Noisy replication in skewed binary classification.
  \textit{Computational Statistics \& Data Analysis} 34(2):165--191

\bibitem[{Leyva et~al(2014)Leyva, Gonz{\'a}lez, and Perez}]{leyva2014set}
Leyva E, Gonz{\'a}lez A, Perez R (2014) A set of complexity measures designed
  for applying meta-learning to instance selection. \textit{IEEE Transactions
  on Knowledge and Data Engineering} 27(2):354--367

\bibitem[{Lorena et~al(2018)Lorena, Maciel, de~Miranda, Costa, and
  Prud{\^e}ncio}]{lorena2018data}
Lorena AC, Maciel AI, de~Miranda PB, et~al (2018) Data complexity meta-features
  for regression problems. \textit{Machine Learning} 107(1):209--246

\bibitem[{Lorena et~al(2019)Lorena, Garcia, Lehmann, Souto, and
  Ho}]{lorena2019complex}
Lorena AC, Garcia LP, Lehmann J, et~al (2019) How complex is your
  classification problem? a survey on measuring classification complexity.
  \textit{ACM Computing Surveys} 52(5):1--34

\bibitem[{Mohr and Wever(2023)}]{mohr2023naive}
Mohr F, Wever M (2023) Naive automated machine learning. \textit{Machine
  Learning} 112(4):1131--1170

\bibitem[{Moniz and Cerqueira(2021)}]{moniz2021automated}
Moniz N, Cerqueira V (2021) Automated imbalanced classification via
  meta-learning. \textit{Expert Systems with Applications} 178:115011

\bibitem[{Moniz et~al(2014)Moniz, Torgo, and Rodrigues}]{moniz2014resampling}
Moniz N, Torgo L, Rodrigues F (2014) Resampling approaches to improve news
  importance prediction. In: \textit{International Symposium on Intelligent
  Data Analysis}, Springer, pp 215--226

\bibitem[{Moniz et~al(2017{\natexlab{a}})Moniz, Branco, and
  Torgo}]{moniz2017resampling}
Moniz N, Branco P, Torgo L (2017{\natexlab{a}}) Resampling strategies for
  imbalanced time series forecasting. \textit{International Journal of Data
  Science and Analytics} 3(3):161--181

\bibitem[{Moniz et~al(2017{\natexlab{b}})Moniz, Branco, and
  Torgo}]{moniz2017evaluation}
Moniz NM, Branco PO, Torgo L (2017{\natexlab{b}}) Evaluation of ensemble
  methods in imbalanced regression tasks. In: \textit{Proceedings of the First
  International Workshop on Learning with Imbalanced Domains: Theory and
  Applications}, pp 129--140

\bibitem[{Mor{\'a}n-Fern{\'a}ndez et~al(2017)Mor{\'a}n-Fern{\'a}ndez,
  Bol{\'o}n-Canedo, and Alonso-Betanzos}]{moran2017can}
Mor{\'a}n-Fern{\'a}ndez L, Bol{\'o}n-Canedo V, Alonso-Betanzos A (2017) Can
  classification performance be predicted by complexity measures? a study using
  microarray data. \textit{Knowledge and Information Systems} 51(3):1067--1090

\bibitem[{Olson et~al(2016)Olson, Bartley, Urbanowicz, and
  Moore}]{OlsonGECCO2016}
Olson RS, Bartley N, Urbanowicz RJ, et~al (2016) Evaluation of a tree-based
  pipeline optimization tool for automating data science. In:
  \textit{Proceedings of the Genetic and Evolutionary Computation Conference
  2016}. ACM, pp 485--492

\bibitem[{Pfahringer et~al(2000)Pfahringer, Bensusan, and
  Giraud-Carrier}]{pfahringer2000meta}
Pfahringer B, Bensusan H, Giraud-Carrier CG (2000) Meta-learning by landmarking
  various learning algorithms. In: \textit{Proceedings of the 17th
  International Conference on Machine Learning}. Morgan Kaufmann Publishers
  Inc., pp 743--750

\bibitem[{Rathore and Kumar(2017{\natexlab{a}})}]{rathore2017linear}
Rathore SS, Kumar S (2017{\natexlab{a}}) Linear and non-linear heterogeneous
  ensemble methods to predict the number of faults in software systems.
  \textit{Knowledge-Based Systems} 119:232--256

\bibitem[{Rathore and Kumar(2017{\natexlab{b}})}]{rathore2017towards}
Rathore SS, Kumar S (2017{\natexlab{b}}) Towards an ensemble based system for
  predicting the number of software faults. \textit{Expert Systems with
  Applications} 82:357--382

\bibitem[{Reif et~al(2014)Reif, Shafait, Goldstein, Breuel, and
  Dengel}]{reif2014automatic}
Reif M, Shafait F, Goldstein M, et~al (2014) Automatic classifier selection for
  non-experts. \textit{Pattern Analysis and Applications} 17:83--96

\bibitem[{Ribeiro(2011)}]{ribeiro2011utility}
Ribeiro R (2011) Utility-based regression. PhD thesis, Dep. Computer Science,
  Faculty of Sciences-University of Porto

\bibitem[{Ribeiro and Moniz(2020)}]{ribeiro2020imbalanced}
Ribeiro RP, Moniz N (2020) Imbalanced regression and extreme value prediction.
  \textit{Machine Learning} 109(9):1803--1835

\bibitem[{Rivolli et~al(2022)Rivolli, Garcia, Soares, Vanschoren, and
  de~Carvalho}]{rivolli2022meta}
Rivolli A, Garcia LP, Soares C, et~al (2022) Meta-features for meta-learning.
  \textit{Knowledge-Based Systems} 240:108101

\bibitem[{Rossi et~al(2021)Rossi, Soares, de~Souza, de~Leon~Ferreira
  et~al}]{rossi2021micro}
Rossi ALD, Soares C, de~Souza BF, et~al (2021) Micro-metastream: algorithm
  selection for time-changing data. \textit{Information Sciences} 565:262--277

\bibitem[{Simon(2013)}]{simon2013evolutionary}
Simon D (2013) \textit{Evolutionary optimization algorithms}. Wiley Online
  Library, Hoboken

\bibitem[{Snoek et~al(2012)Snoek, Larochelle, and Adams}]{snoek2012practical}
Snoek J, Larochelle H, Adams RP (2012) Practical bayesian optimization of
  machine learning algorithms. In: \textit{Proceedings of the 26th
  International Conference on Neural Information Processing Systems}. Curran
  Associates Inc., pp 2951--2959

\bibitem[{Song et~al(2022)Song, Dao, and Branco}]{song2022distsmogn}
Song XY, Dao N, Branco P (2022) Distsmogn: Distributed smogn for imbalanced
  regression problems. In: \textit{Fourth International Workshop on Learning
  with Imbalanced Domains: Theory and Applications}, PMLR, pp 38--52

\bibitem[{Sousa et~al(2016)Sousa, Prud{\^e}ncio, Ludermir, and
  Soares}]{sousa2016active}
Sousa AF, Prud{\^e}ncio RB, Ludermir TB, et~al (2016) Active learning and data
  manipulation techniques for generating training examples in meta-learning.
  \textit{Neurocomputing} 194:45--55

\bibitem[{Thornton et~al(2013)Thornton, Hutter, Hoos, and
  Leyton-Brown}]{thornton2013auto}
Thornton C, Hutter F, Hoos HH, et~al (2013) Auto-weka: Combined selection and
  hyperparameter optimization of classification algorithms. In:
  \textit{Proceedings of the 19th ACM SIGKDD International Conference on
  Knowledge Discovery and Data Mining}, pp 847--855

\bibitem[{Torgo and Ribeiro(2009)}]{torgo2009precision}
Torgo L, Ribeiro R (2009) Precision and recall for regression. In:
  \textit{Discovery Science}. Springer Berlin Heidelberg, pp 332--346

\bibitem[{Torgo et~al(2013)Torgo, Ribeiro, Pfahringer, and
  Branco}]{torgo2013smote}
Torgo L, Ribeiro RP, Pfahringer B, et~al (2013) Smote for regression. In:
  \textit{Portuguese Conference on Artificial Intelligence}, Springer, pp
  378--389

\bibitem[{Tukey(1970)}]{tukey1970exploratory}
Tukey J (1970) \textit{Exploratory Data Analysis}. Addison-Wesley, Reading,
  Mass

\bibitem[{Vakhrushev et~al(2021)Vakhrushev, Ryzhkov, Savchenko, Simakov,
  Damdinov, and Tuzhilin}]{vakhrushev2021lightautoml}
Vakhrushev A, Ryzhkov A, Savchenko M, et~al (2021) Lightautoml: Automl solution
  for a large financial services ecosystem. arXiv preprint ArXiv:2109.01528

\bibitem[{Vanschoren(2019)}]{vanschoren2019meta}
Vanschoren J (2019) Meta-learning. \textit{Automated machine learning: methods,
  systems, challenges} pp 35--61

\bibitem[{Vanschoren et~al(2014)Vanschoren, Van~Rijn, Bischl, and
  Torgo}]{vanschoren2014openml}
Vanschoren J, Van~Rijn JN, Bischl B, et~al (2014) Openml: networked science in
  machine learning. \textit{ACM SIGKDD Explorations Newsletter} 15(2):49--60

\bibitem[{Wang et~al(2021)Wang, Wu, Weimer, and Zhu}]{wang2021flaml}
Wang C, Wu Q, Weimer M, et~al (2021) Flaml: A fast and lightweight automl
  library. In: \textit{Proceedings of Machine Learning and Systems}, pp
  434--447

\bibitem[{Wu et~al(2021)Wu, Wang, and Huang}]{wu2021frugal}
Wu Q, Wang C, Huang S (2021) Frugal optimization for cost-related
  hyperparameters. In: \textit{Proceedings of the AAAI Conference on Artificial
  Intelligence}, pp 10347--10354

\bibitem[{Wu et~al(2022)Wu, Kunz, and
  Branco}]{wu2022imbalancedlearningregression}
Wu W, Kunz N, Branco P (2022) Imbalancedlearningregression-a python package to
  tackle the imbalanced regression problem. In: \textit{Joint European
  Conference on Machine Learning and Knowledge Discovery in Databases},
  Springer, pp 645--648

\bibitem[{Yao et~al(2018)Yao, Wang, Chen, Dai, Li, Tu, Yang, and
  Yu}]{yao2018taking}
Yao Q, Wang M, Chen Y, et~al (2018) Taking human out of learning applications:
  A survey on automated machine learning. arXiv preprint ArXiv:1810.13306v2

\bibitem[{Z{\"o}ller and Huber(2021)}]{zoller2021benchmark}
Z{\"o}ller MA, Huber MF (2021) Benchmark and survey of automated machine
  learning frameworks. \textit{Journal of Artificial Intelligence Research}
  70:409--472

\end{thebibliography}


\newpage

\appendix
\section*{Appendix A - Dataset description}
\label{appendixA}

\begin{center}
\centering
\begin{longtable}{@{}lcccc@{}}
\caption{Datasets ordered by the percentage of rare cases. (n.examples: Number of examples; n.attributes: Number of attributes; n.rare: Number of rare cases; p.rare: $100 \times n.raro/n.exemplos$}\\
\toprule
\multicolumn{1}{c}{\textbf{Dataset}} & \textbf{n.samples} & \textbf{n.attributes} & \textbf{n.rare} & \textbf{p.rare} \\ \midrule
\endfirsthead
\toprule
\multicolumn{1}{c}{\textbf{Dataset}} & \textbf{n.samples} & \textbf{n.attributes} & \textbf{n.rare} & \textbf{p.rare} \\ \midrule
\endhead
QSAR-TID-20137                       & 101                & 1024                  & 40              & 39,6            \\
QSAR-TID-11569                       & 120                & 1024                  & 44              & 36,7            \\
QSAR-TID-11637                       & 96                 & 1024                  & 34              & 35,4            \\
QSAR-TID-10426                       & 27                 & 1024                  & 9               & 33,3            \\
QSAR-TID-12847                       & 215                & 1024                  & 68              & 31,6            \\
QSAR-TID-101612                      & 91                 & 1024                  & 28              & 30,8            \\
QSAR-TID-101332                      & 106                & 1024                  & 32              & 30,2            \\
QSAR-TID-12263                       & 47                 & 1024                  & 13              & 27,7            \\
QSAR-TID-101602                      & 79                 & 1024                  & 21              & 26,6            \\
QSAR-TID-12959                       & 72                 & 1024                  & 19              & 26,4            \\
QSAR-TID-10778                       & 82                 & 1024                  & 21              & 25,6            \\
rabe-265                             & 51                 & 6                     & 13              & 25,5            \\
QSAR-TID-30050                       & 80                 & 1024                  & 20              & 25,0            \\
QSAR-TID-30005                       & 81                 & 1024                  & 20              & 24,7            \\
QSAR-TID-103453                      & 73                 & 1024                  & 18              & 24,7            \\
QSAR-TID-101333                      & 168                & 1024                  & 41              & 24,4            \\
QSAR-TID-101226                      & 79                 & 1024                  & 19              & 24,1            \\
QSAR-TID-11843                       & 84                 & 1024                  & 20              & 23,8            \\
QSAR-TID-101553                      & 80                 & 1024                  & 19              & 23,8            \\
QSAR-TID-10939                       & 81                 & 1024                  & 19              & 23,5            \\
wine-quality                         & 6497               & 11                    & 1523            & 23,4            \\
analcat-apnea3                   & 450                & 11                    & 103             & 22,9            \\
QSAR-TID-101231                      & 79                 & 1024                  & 18              & 22,8            \\
analcat-ncaa                     & 120                & 19                    & 27              & 22,5            \\
QSAR-TID-19607                       & 143                & 1024                  & 32              & 22,4            \\
QSAR-TID-101317                      & 99                 & 1024                  & 22              & 22,2            \\
QSAR-TID-10478                       & 86                 & 1024                  & 19              & 22,1            \\
analcat-apnea1                   & 475                & 11                    & 104             & 21,9            \\
QSAR-TID-100833                      & 83                 & 1024                  & 18              & 21,7            \\
QSAR-TID-30024                       & 84                 & 1024                  & 18              & 21,4            \\
QSAR-TID-100975                      & 75                 & 1024                  & 16              & 21,3            \\
analcat-apnea2                   & 475                & 11                    & 101             & 21,3            \\
QSAR-TID-101448                      & 80                 & 1024                  & 17              & 21,3            \\
QSAR-TID-101340                      & 86                 & 1024                  & 18              & 20,9            \\
analcat-supreme                  & 4052               & 7                     & 835             & 20,6            \\
meta                                 & 528                & 65                    & 108             & 20,5            \\
QSAR-TID-100918                      & 88                 & 1024                  & 18              & 20,5            \\
QSAR-TID-101048                      & 74                 & 1024                  & 15              & 20,3            \\
QSAR-TID-101191                      & 74                 & 1024                  & 15              & 20,3            \\
QSAR-TID-100925                      & 79                 & 1024                  & 16              & 20,3            \\
QSAR-TID-10621                       & 99                 & 1024                  & 20              & 20,2            \\
QSAR-TID-10618                       & 134                & 1024                  & 27              & 20,1            \\
kc1-numeric                          & 145                & 94                    & 29              & 20,0            \\
QSAR-TID-101399                      & 75                 & 1024                  & 15              & 20,0            \\
QSAR-TID-101433                      & 81                 & 1024                  & 16              & 19,8            \\
QSAR-TID-30009                       & 102                & 1024                  & 20              & 19,6            \\
QSAR-TID-100063                      & 149                & 1024                  & 29              & 19,5            \\
QSAR-TID-30027                       & 83                 & 1024                  & 16              & 19,3            \\
QSAR-TID-20162                       & 156                & 1024                  & 30              & 19,2            \\
QSAR-TID-101278                      & 79                 & 1024                  & 15              & 19,0            \\
QSAR-TID-101584                      & 74                 & 1024                  & 14              & 18,9            \\
QSAR-TID-11281                       & 43                 & 1024                  & 8               & 18,6            \\
QSAR-TID-100951                      & 81                 & 1024                  & 15              & 18,5            \\
QSAR-TID-11784                       & 65                 & 1024                  & 12              & 18,5            \\
QSAR-TID-11873                       & 115                & 1024                  & 21              & 18,3            \\
analcat-chlamydia                & 100                & 17                    & 18              & 18,0            \\
QSAR-TID-30034                       & 162                & 1024                  & 29              & 17,9            \\
socmob                               & 1156               & 39                    & 206             & 17,8            \\
QSAR-TID-101130                      & 79                 & 1024                  & 14              & 17,7            \\
QSAR-TID-101239                      & 80                 & 1024                  & 14              & 17,5            \\
QSAR-TID-12780                       & 121                & 1024                  & 21              & 17,4            \\
QSAR-TID-101055                      & 81                 & 1024                  & 14              & 17,3            \\
QSAR-TID-30022                       & 81                 & 1024                  & 14              & 17,3            \\
QSAR-TID-30041                       & 81                 & 1024                  & 14              & 17,3            \\
QSAR-TID-30015                       & 116                & 1024                  & 20              & 17,2            \\
analcat-wildcat                  & 163                & 5                     & 28              & 17,2            \\
QSAR-TID-11056                       & 41                 & 1024                  & 7               & 17,1            \\
QSAR-TID-30010                       & 82                 & 1024                  & 14              & 17,1            \\
QSAR-TID-10844                       & 166                & 1024                  & 28              & 16,9            \\
QSAR-TID-30020                       & 89                 & 1024                  & 15              & 16,9            \\
a6                                   & 198                & 11                    & 33              & 16,7            \\
cocomo-numeric                       & 60                 & 56                    & 10              & 16,7            \\
QSAR-TID-30004                       & 84                 & 1024                  & 14              & 16,7            \\
kdd-coil-6                           & 316                & 18                    & 52              & 16,5            \\
QSAR-TID-100906                      & 79                 & 1024                  & 13              & 16,5            \\
QSAR-TID-101324                      & 79                 & 1024                  & 13              & 16,5            \\
QSAR-TID-101309                      & 73                 & 1024                  & 12              & 16,4            \\
QSAR-TID-12718                       & 134                & 1024                  & 22              & 16,4            \\
QSAR-TID-100416                      & 122                & 1024                  & 20              & 16,4            \\
machineCPU                           & 209                & 6                     & 34              & 16,3            \\
abalone                              & 4177               & 8                     & 679             & 16,3            \\
a3                                   & 198                & 11                    & 32              & 16,2            \\
nasa-numeric                         & 93                 & 90                    & 15              & 16,1            \\
QSAR-TID-102669                      & 180                & 1024                  & 29              & 16,1            \\
QSAR-TID-30042                       & 81                 & 1024                  & 13              & 16,0            \\
QSAR-TID-101204                      & 75                 & 1024                  & 12              & 16,0            \\
QSAR-TID-10983                       & 119                & 1024                  & 19              & 16,0            \\
QSAR-TID-30032                       & 107                & 1024                  & 17              & 15,9            \\
cpu-act                              & 209                & 36                    & 33              & 15,8            \\
QSAR-TID-30000                       & 83                 & 1024                  & 13              & 15,7            \\
a4                                   & 198                & 11                    & 31              & 15,7            \\
QSAR-TID-30016                       & 97                 & 1024                  & 15              & 15,5            \\
QSAR-TID-17121                       & 182                & 1024                  & 28              & 15,4            \\
QSAR-TID-11692                       & 98                 & 1024                  & 15              & 15,3            \\
forestFires                          & 517                & 12                    & 79              & 15,3            \\
kdd-coil-3                           & 316                & 18                    & 48              & 15,2            \\
QSAR-TID-101503                      & 79                 & 1024                  & 12              & 15,2            \\
QSAR-TID-11361                       & 79                 & 1024                  & 12              & 15,2            \\
QSAR-TID-101610                      & 145                & 1024                  & 22              & 15,2            \\
analcat-olympic2000              & 66                 & 11                    & 10              & 15,2            \\
QSAR-TID-12407                       & 66                 & 1024                  & 10              & 15,2            \\
sleuth-case1202                      & 93                 & 6                     & 14              & 15,1            \\
QSAR-TID-101312                      & 80                 & 1024                  & 12              & 15,0            \\
QSAR-TID-10782                       & 116                & 1024                  & 17              & 14,7            \\
QSAR-TID-10143                       & 62                 & 1024                  & 9               & 14,5            \\
QSAR-TID-11639                       & 201                & 1024                  & 29              & 14,4            \\
a1                                   & 198                & 11                    & 28              & 14,1            \\
QSAR-TID-11107                       & 85                 & 1024                  & 12              & 14,1            \\
QSAR-TID-10466                       & 78                 & 1024                  & 11              & 14,1            \\
fri-c2-100-5                         & 100                & 5                     & 14              & 14,0            \\
fri-c4-250-10                        & 250                & 10                    & 35              & 14,0            \\
kdd-coil-4                           & 316                & 18                    & 44              & 13,9            \\
QSAR-TID-101506                      & 79                 & 1024                  & 11              & 13,9            \\
a7                                   & 198                & 11                    & 27              & 13,6            \\
QSAR-TID-100931                      & 110                & 1024                  & 15              & 13,6            \\
kdd-coil-2                           & 316                & 18                    & 43              & 13,6            \\
fri-c3-250-5                         & 250                & 5                     & 34              & 13,6            \\
QSAR-TID-100835                      & 125                & 1024                  & 17              & 13,6            \\
QSAR-TID-20025                       & 89                 & 1024                  & 12              & 13,5            \\
analcat-election2000             & 67                 & 14                    & 9               & 13,4            \\
humans-numeric                       & 75                 & 14                    & 10              & 13,3            \\
QSAR-TID-103800                      & 91                 & 1024                  & 12              & 13,2            \\
kidney                               & 76                 & 10                    & 10              & 13,2            \\
veteran                              & 137                & 13                    & 18              & 13,1            \\
kdd-coil-7                           & 316                & 18                    & 41              & 13,0            \\
analcat-seropositive             & 132                & 5                     & 17              & 12,9            \\
boston                               & 506                & 13                    & 65              & 12,8            \\
QSAR-TID-11113                       & 94                 & 1024                  & 12              & 12,8            \\
pdgfr                                & 79                 & 320                   & 10              & 12,7            \\
fri-c3-500-5                         & 500                & 5                     & 63              & 12,6            \\
QSAR-TID-10187                       & 64                 & 1024                  & 8               & 12,5            \\
QSAR-TID-11094                       & 192                & 1024                  & 24              & 12,5            \\
QSAR-TID-100424                      & 97                 & 1024                  & 12              & 12,4            \\
QSAR-TID-102667                      & 106                & 1024                  & 13              & 12,3            \\
kdd-coil-5                           & 316                & 18                    & 38              & 12,0            \\
sensory                              & 576                & 11                    & 69              & 12,0            \\
auto93                               & 93                 & 57                    & 11              & 11,8            \\
QSAR-TID-11299                       & 68                 & 1024                  & 8               & 11,8            \\
QSAR-TID-10929                       & 154                & 1024                  & 18              & 11,7            \\
delta-elevators                      & 9517               & 6                     & 1109            & 11,7            \\
QSAR-TID-10012                       & 224                & 1024                  & 26              & 11,6            \\
a2                                   & 198                & 11                    & 22              & 11,1            \\
QSAR-TID-101504                      & 99                 & 1024                  & 11              & 11,1            \\
fri-c3-100-50                        & 100                & 50                    & 11              & 11,0            \\
QSAR-TID-20128                       & 100                & 1024                  & 11              & 11,0            \\
QSAR-TID-17086                       & 74                 & 1024                  & 8               & 10,8            \\
kdd-coil-1                           & 316                & 18                    & 34              & 10,8            \\
triazines                            & 186                & 60                    & 20              & 10,8            \\
airfoild                             & 1503               & 5                     & 161             & 10,7            \\
a5                                   & 198                & 11                    & 21              & 10,6            \\
QSAR-TID-101301                      & 151                & 1024                  & 16              & 10,6            \\
treasury                             & 1049               & 15                    & 109             & 10,4            \\
pharynx                              & 195                & 10                    & 20              & 10,3            \\
mortgage                             & 1049               & 15                    & 106             & 10,1            \\
debutanizer                          & 2394               & 7                     & 240             & 10,0            \\
QSAR-TID-100790                      & 170                & 1024                  & 17              & 10,0            \\
QSAR-TID-103071                      & 150                & 1024                  & 15              & 10,0            \\
fri-c4-1000-25                       & 1000               & 25                    & 99              & 9,9             \\
QSAR-TID-30046                       & 81                 & 1024                  & 8               & 9,9             \\
QSAR-TID-10407                       & 176                & 1024                  & 17              & 9,7             \\
QSAR-TID-11209                       & 95                 & 1024                  & 9               & 9,5             \\
QSAR-TID-101033                      & 75                 & 1024                  & 7               & 9,3             \\
fuel-consumption-country             & 1764               & 37                    & 164             & 9,3             \\
QSAR-TID-12131                       & 111                & 1024                  & 10              & 9,0             \\
fri-c2-500-25                        & 500                & 25                    & 45              & 9,0             \\
heat                                 & 7400               & 11                    & 664             & 9,0             \\
QSAR-TID-100127                      & 101                & 1024                  & 9               & 8,9             \\
QSAR-TID-10967                       & 101                & 1024                  & 9               & 8,9             \\
fri-c4-1000-10                       & 1000               & 10                    & 89              & 8,9             \\
california                           & 20640              & 8                     & 1821            & 8,8             \\
available-power                      & 1802               & 15                    & 157             & 8,7             \\
compactiv                            & 8192               & 21                    & 713             & 8,7             \\
cpu                                  & 8192               & 21                    & 713             & 8,7             \\
fishcatch                            & 158                & 7                     & 13              & 8,2             \\
fri-c3-1000-10                       & 1000               & 10                    & 82              & 8,2             \\
fri-c3-1000-5                        & 1000               & 5                     & 82              & 8,2             \\
autoPrice                            & 159                & 15                    & 13              & 8,2             \\
sleuth-case2002                      & 147                & 6                     & 12              & 8,2             \\
QSAR-TID-103062                      & 74                 & 1024                  & 6               & 8,1             \\
cps-85-wages                         & 534                & 23                    & 43              & 8,1             \\
chscase-census2                      & 400                & 7                     & 32              & 8,0             \\
QSAR-TID-11567                       & 76                 & 1024                  & 6               & 7,9             \\
chatfield-4                          & 235                & 12                    & 18              & 7,7             \\
places                               & 329                & 8                     & 25              & 7,6             \\
QSAR-TID-12536                       & 202                & 1024                  & 15              & 7,4             \\
plasma-retinol                       & 315                & 18                    & 23              & 7,3             \\
QSAR-TID-100155                      & 138                & 1024                  & 10              & 7,2             \\
maximal                              & 1802               & 32                    & 129             & 7,2             \\
QSAR-TID-12635                       & 85                 & 1024                  & 6               & 7,1             \\
QSAR-TID-10113                       & 131                & 1024                  & 9               & 6,9             \\
QSAR-TID-12887                       & 117                & 1024                  & 8               & 6,8             \\
QSAR-TID-102988                      & 88                 & 1024                  & 6               & 6,8             \\
chscase-census3                      & 400                & 7                     & 27              & 6,8             \\
QSAR-TID-10856                       & 121                & 1024                  & 8               & 6,6             \\
cloud                                & 108                & 9                     & 7               & 6,5             \\
Moneyball                            & 1232               & 53                    & 79              & 6,4             \\
fri-c0-250-10                        & 250                & 10                    & 16              & 6,4             \\
bank8FM                              & 8192               & 8                     & 524             & 6,4             \\
QSAR-TID-10728                       & 172                & 1024                  & 11              & 6,4             \\
QSAR-TID-19689                       & 157                & 1024                  & 10              & 6,4             \\
no2                                  & 500                & 7                     & 31              & 6,2             \\
QSAR-TID-12173                       & 114                & 1024                  & 7               & 6,1             \\
QSAR-TID-12013                       & 200                & 1024                  & 12              & 6,0             \\
QSAR-TID-12587                       & 157                & 1024                  & 9               & 5,7             \\
lungcancer-shedden                   & 442                & 24                    & 25              & 5,7             \\
cholesterol                          & 303                & 13                    & 17              & 5,6             \\
space-ga                             & 3107               & 6                     & 173             & 5,6             \\
liver-disorders                      & 345                & 5                     & 19              & 5,5             \\
chscase-census6                      & 400                & 6                     & 22              & 5,5             \\
concreteStrength                     & 1030               & 8                     & 55              & 5,3             \\
acceleration                         & 1732               & 14                    & 89              & 5,1             \\
pwLinear                             & 200                & 10                    & 10              & 5,0             \\
wind                                 & 6574               & 14                    & 283             & 4,3             \\
pm10                                 & 500                & 7                     & 21              & 4,2             \\
chscase-census5                      & 400                & 7                     & 13              & 3,3             \\
fri-c0-1000-50                       & 1000               & 50                    & 27              & 2,7             \\
fri-c3-1000-50                       & 1000               & 50                    & 25              & 2,5             \\
fri-c4-1000-50                       & 1000               & 50                    & 20              & 2,0             \\ \bottomrule

\end{longtable}
\label{tab:datasets}
\end{center}

\section*{Appendix B - Meta-features description}
\label{AppendixB}

\subsection*{Maximum feature correlation to the output ($C_1$):}

The absolute value of Spearman's correlation ($\rho$) is calculated between each attribute ($x$) and the outputs ($y$). $C_1$ is the maximum value obtained among all attributes, as represented in Equation \ref{c1}. A higher value of this measure indicates a simpler problem.

\begin{equation}
\label{c1}
     C_1 = \max\limits_{j=1,...,d}|\rho(x^j, y)|
\end{equation}

\subsection*{Average feature correlation to the output ($C_2$):}

In this measure, the average of the correlations of all features with the output is calculated, as shown in Equation \ref{c2}. Similarly to $C_1$, higher values indicate simpler problems.

\begin{equation}
\label{c2}
     C_2 = \sum\limits_{j=1}^{d}\frac{|\rho(x^j, y)|}{d}
\end{equation}

\subsection*{Individual feature efficiency ($C_3$):}

For each attribute, the number of examples that need to be removed from the dataset to achieve a high correlation with the output is calculated. Then, the number of removed examples ($n^j$) is divided by the total number of examples ($n$) for each attribute, and the minimum of these values is returned, as shown in Equation \ref{c3}.

\begin{equation}
\label{c3}
     C_3 = \min\limits_{j=1}^{d}\frac{n^j}{n}
\end{equation}

\subsection*{Collective feature efficiency ($C_4$):}

Initially, the attribute with the highest correlation with the output is identified. Then, all examples with $|\epsilon_i| < 0.1$ are excluded. Subsequently, the most correlated attribute with the remaining points is found, and the process is repeated until all attributes have been analyzed. Finally, the proportion of examples where $|\epsilon_i| > 0.1$ is returned, as described in Equation \ref{c4}.

\begin{equation}
\label{c4}
     C_4 = \frac{\#\{X_i||\epsilon_i|>0.1\}T_l}{n}
\end{equation}

Where $n$ is the number of examples remaining in the dataset. $T_l$ is the dataset from which this number is calculated, and $l$ is the number of iterations performed by the algorithm.

\subsection*{Mean absolute error ($L_1$):}

In $L_1$ (Equation \ref{l1}), the average of the absolute values of the residuals from a multiple linear regressor is calculated. Lower values indicate simpler problems.

\begin{equation}
\label{l1}
     L_1 = \sum\limits_{i=1}^{n}\frac{|\varepsilon_i|}{n}
\end{equation}

\subsection*{Residual variance ($L_2$):}

In $L_2$, the average of the squared residuals from a multiple linear regressor is calculated. Lower values indicate simpler problems, as described in Equation \ref{l2}.

\begin{equation}
\label{l2}
     L_2 = \sum\limits_{i=1}^{n}\frac{\varepsilon_i^2}{n}
\end{equation}

\subsection*{Output distribution ($S_1$):}

Initially, a Minimum Spanning Tree (MST) is generated from the input data. Each data item corresponds to a vertex in the graph, while edges are weighted according to the Euclidean distance between examples in the input space. The MST connects the closest examples to each other. Finally, $S_1$ monitors whether the examples joined in the MST have similar output values. Lower values indicate simpler problems. This measure is expressed in Equation \ref{s1}.

\begin{equation}
\label{s1}
     S_1 = \frac{1}{n}\sum\limits_{i:j\in MST}|y_i-y_j|
\end{equation}

Where the sum is taken over all vertices $i$ and $j$ that are adjacent in the MST. $S_1$ calculates the average of the outputs of the points connected in the MST.

\subsection*{Input distribution ($S_2$):}

In this measure, the Euclidean distance between pairs of neighboring examples is calculated. To achieve this, the data points are initially sorted according to their output values. $S_2$ is presented in Equation \ref{s2}.

\begin{equation}
\label{s2}
     S_2 = \frac{1}{n}\sum\limits_{i=1}^{n}||x_i-x_{i-1}||_2
\end{equation}

\subsection*{Error of a nearest neighbor regressor  ($S_3$):}

In $S_3$, the closeness of examples is measured. For this purpose, a 1-NN regressor looks for the training example ($x_i$) most similar to the new example and assigns it the same output ($y_i$), as shown in Equation \ref{s3}.

\begin{equation}
\label{s3}
     S_3 = \frac{1}{n}\sum\limits_{i=1}^{n}(NN(x_i)-y_i)^2
\end{equation}

\subsection*{Non-linearity of a linear regressor ($L_3$):}

Given a dataset, pairs of examples with similar outputs are initially selected, and a new test point is created by performing random interpolation. The original data is then used to train a linear regressor, and the new points are used to measure the mean squared error (MSE).

\begin{equation}
\label{l3}
     L_3 = \frac{1}{l}\sum\limits_{i=1}^{l}(f(x'_i)-y'_i)^2
\end{equation}

Where $l$ is the number of interpolated examples, $x'_i$ are the generated points, and $y'_i$ are their labels. Lower values indicate simpler problems.

\subsection*{Non-linearity of nearest neighbor regressor ($S_4$):}

This measure employs the same steps described for $L_3$, but uses a nearest neighbor regressor for predictions. $S_4$ is defined in Equation \ref{s4}.

\begin{equation}
\label{s4}
     S_4 = \frac{1}{l}\sum\limits_{i=1}^{l}(NN(x'_i)-y'_i)^2
\end{equation}

\subsection*{Average number of examples per dimension ($T_2$):}

It is defined as the average number of examples ($n$) per dimension ($d$), as presented in Equation \ref{t2}. Lower values indicate more complex datasets.

\begin{equation}
\label{t2}
     T_2 = \frac{n}{d}
\end{equation}

\subsection*{Average number of PCA dimensions per data point ($T_3$):}
The metric $T_3$ uses Principal Component Analysis (PCA) to assess dataset characteristics. Unlike $T_2$, which relies on the raw dimensionality of the feature vector, $T_3$ employs the number of PCA components required to capture 95\% of data variability (denoted as $m'$) as the foundation for evaluating data sparsity.

\begin{equation}
\label{t2}
     T_3 = \frac{m'}{n}
\end{equation}

\subsection*{Ratio of PCA dimensions to the original dimensions ($T_4$):}

This measure estimates the proportion of relevant dimensions within the dataset. The concept of relevance is evaluated based on the PCA criterion, which aims to transform features into uncorrelated linear functions that effectively describe the majority of data variability.

\begin{equation}
\label{t2}
     T_4 = \frac{m'}{m}
\end{equation}

\begin{table}[!h]
\caption{Characteristic, acronym, aggregation functions and description of meta-features.}
\scalefont{0.6}
\setlength{\tabcolsep}{2pt}
\begin{tabular}{@{}lccc@{}}
\toprule
\textbf{Characteristic} & \textbf{Acronym} & \textbf{Aggregation Functions} & \textbf{Description} \\
\midrule
\multirow{4}{*}{Simple} & n.examples & - & Number of examples \\
& n.attributes & - & Number of attributes \\
& n.rare & - & Number of rare cases ($\phi>0.8$) \\
& p.rare & - & Percentage of rare cases \\
\midrule
\multirow{3}{*}{\begin{tabular}[c]{@{}c@{}}Dataset\\ Distribution\end{tabular}} & T2 & - & Average number of examples per dimension \\
& T3 & - & Average intrinsic dimension per number of examples \\
& T4 & - & Proportion of intrinsic dimensionality \\
\midrule
\multirow{4}{*}{\begin{tabular}[c]{@{}c@{}}Correlation between\\ Attributes and Targets\end{tabular}} 
& C2 & \{avg, max, min, sd\} & Average feature correlation to the output \\
& C3 & \{avg, max, min, sd\} & Individual feature efficiency \\
& C4 & \{avg\} & Collective feature efficiency \\
\midrule
\multirow{3}{*}{\begin{tabular}[c]{@{}c@{}}Linear Regression-\\ Related Performance Metrics\end{tabular}} & L1 & \{avg, max, min, sd\} & Mean absolute error \\
& L2 & \{avg, max, min\} & Residual variance  \\
& L3 & \{avg, max, min, sd\} & Non-linearity of a linear regressor \\
\midrule
\multirow{4}{*}{Data Smoothness} & S1 & \{avg, max, min, sd\} & Output distribution \\
& S2 & \{avg, max, min, sd\} & Input distribution \\
& S3 & \{avg, max, min, sd\} & Error of a nearest neighbor regressor \\
& S4 & \{avg, max, min, sd\} & Non-linearity of nearest neighbor regressor \\
\midrule
\multirow{2}{*}{Prediction **} & Model & - & Predicted learning model \\
& Strategy & - & Predicted resampling strategy \\
\bottomrule
\end{tabular}
** Used only in Model First and Strategy First recommendation procedures.
\label{tab:mf}
\end{table}

\newpage

\section*{Appendix C - Meta-models evaluation}

\begin{table}[!h]
\caption{Performances (Accuracy) achieved by the meta-models for each type of training. Best results are in bold.}
\scalefont{0.8}
\begin{tabular}{@{}lcccccccc@{}}
\toprule
\textbf{Approach}        & \textbf{Label} & \textbf{DESMI} & \textbf{KNORAU} & \textbf{KNORAE} & \textbf{ET}    & \textbf{BG} & \textbf{RF}    & \textbf{XGB}   \\ \midrule
\textbf{Independent}     & \textbf{$r$}   & 0.376          & 0.385           & 0.376           & 0.417          & 0.408       & 0.417          & \textbf{0.431} \\
\textbf{Independent}     & \textbf{$l$}   & 0.376          & 0.394           & 0.417           & 0.477          & 0.440       & \textbf{0.500} & 0.431          \\
\textbf{Model\_first}    & \textbf{$r$}   & 0.353          & 0.390           & 0.289           & 0.417          & 0.408       & \textbf{0.454} & 0.436          \\
\textbf{Model\_first}    & \textbf{$l$}   & 0.372          & 0.385           & 0.367           & 0.482          & 0.454       & \textbf{0.491} & 0.431          \\
\textbf{Strategy\_first} & \textbf{$r$}   & 0.339          & 0.427           & 0.335           & 0.390          & 0.394       & 0.417          & \textbf{0.431} \\
\textbf{Strategy\_first} & \textbf{$l$}   & 0.408          & 0.422           & 0.390           & \textbf{0.468} & 0.450       & 0.459          & 0.431          \\ \bottomrule
\end{tabular}
\label{tab:meta-modelsF1}
\end{table}

\begin{table}[!h]
\caption{Performances (Accuracy) achieved by the meta-models for each type of training. Best results are in bold.}
\scalefont{0.8}
\begin{tabular}{@{}lcccccccc@{}}
\toprule
\textbf{Approach}        & \textbf{Label} & \textbf{DESMI} & \textbf{KNORAU} & \textbf{KNORAE} & \textbf{ET}    & \textbf{BG} & \textbf{RF}    & \textbf{XGB}   \\ \midrule
\textbf{Independent}     & \textbf{$r$}   & 0.330          & 0.339           & 0.303           & 0.353               & 0.317            & \textbf{0.362}        & 0.335            \\
\textbf{Independent}     & \textbf{$l$}   & 0.367          & 0.404           & 0.427           & 0.440               & 0.440            & \textbf{0.472}        & 0.394            \\
\textbf{Model\_first}    & \textbf{$r$}   & 0.257          & 0.321           & 0.307           & \textbf{0.376}      & 0.326            & \textbf{0.376}        & 0.339            \\
\textbf{Model\_first}    & \textbf{$l$}   & 0.404          & 0.381           & 0.394           & \textbf{0.459}      & 0.417            & 0.454                 & 0.394            \\
\textbf{Strategy\_first} & \textbf{$r$}   & 0.330          & 0.330           & 0.289           & \textbf{0.362}      & 0.358            & \textbf{0.362}        & 0.335            \\
\textbf{Strategy\_first} & \textbf{$l$}   & 0.376          & 0.413           & 0.344           & 0.436               & 0.427            & \textbf{0.468}        & 0.408            \\ \hline
\
\end{tabular}
\label{tab:meta-modelsSERA}
\end{table}

\newpage

\section*{Appendix D - Meta-target distribution}
\label{AppendixC}

\begin{figure}[!h]
  \centering

\subfigure[F1-scoreR]{\includegraphics[scale=0.4]{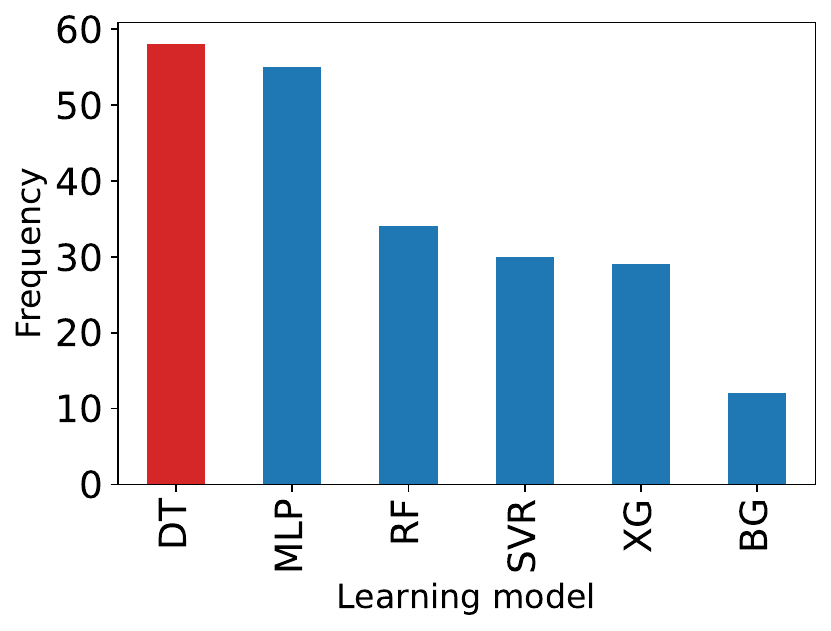}}
\subfigure[F1-scoreR]{\includegraphics[scale=0.4]{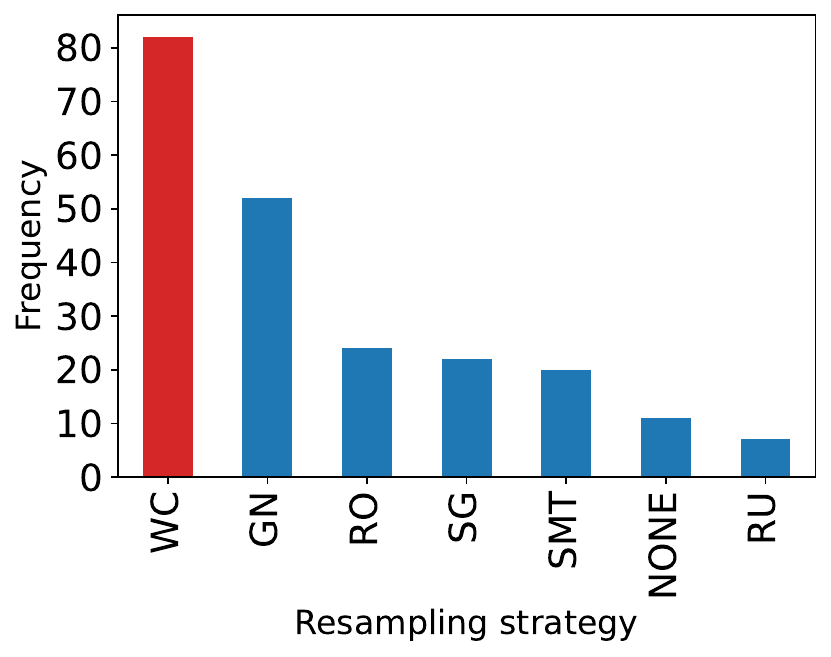}}

\subfigure[SERA]{\includegraphics[scale=0.4]{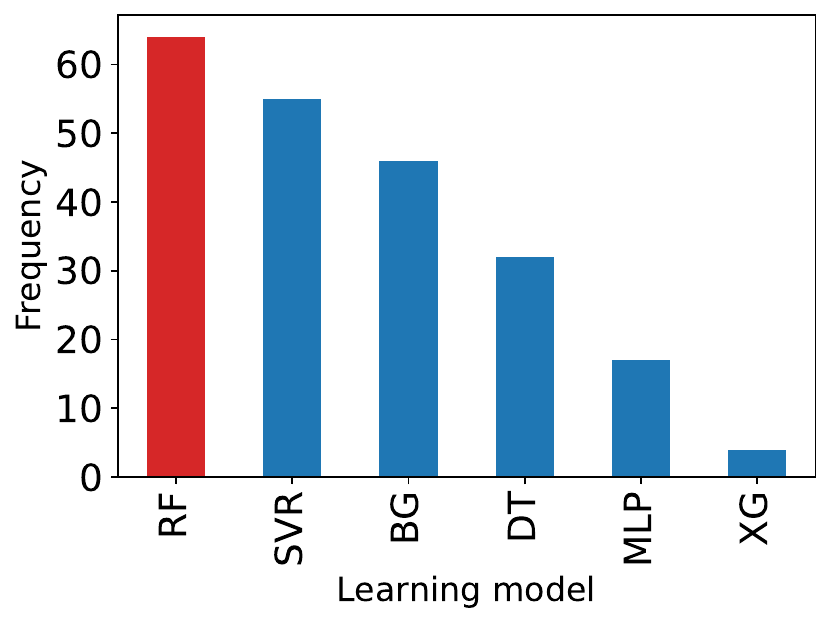}}
\subfigure[SERA]{\includegraphics[scale=0.4]{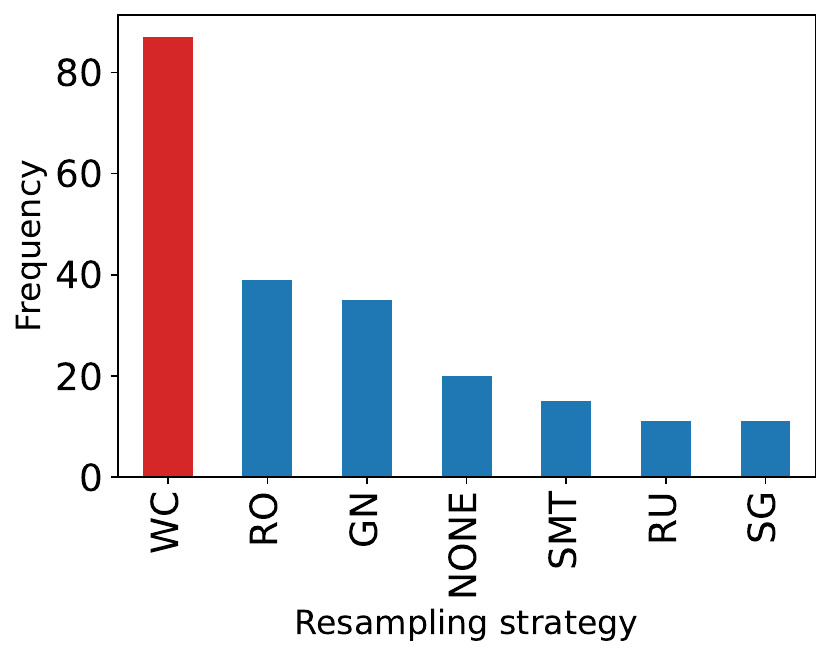}}

\caption{Frequency of resampling strategies and learning models used as meta-target.}
\label{fig:frequency}
\end{figure}

\end{document}